\documentclass[a4paper]{article}
\pdfoutput=1 

\usepackage{arxiv}

\usepackage[utf8]{inputenc} 
\usepackage[T1]{fontenc}    
\usepackage{hyperref}       
\usepackage{url}            
\usepackage{booktabs}       
\usepackage{amsfonts}       
\usepackage{nicefrac}       
\usepackage{microtype}      

\usepackage{lmodern}
\usepackage{todonotes}
\usepackage{cancel}
\usepackage{amsthm,amssymb, amsmath}
\usepackage{xcolor}
\usepackage{graphicx}
\usepackage{float}
\usepackage{subfig}
\usepackage{cleveref}

\newtheoremstyle{custom}
                {\topsep}
                {\topsep}
                {\itshape}
                {}
                {\bfseries}
                {.}
                {\newline}
                {\thmname{#1}\thmnumber{ #2}\thmnote{ (#3)}}%
\theoremstyle{custom}

\newtheorem{proposition}{Proposition}

\usepackage{caption}
\DeclareCaptionStyle{italic}[justification=centering]
 {labelfont={bf},textfont={it},labelsep=colon}
\title{Principles and Algorithms for Forecasting Groups of Time Series:
Locality and Globality
}

\author{
  Pablo Montero-Manso
    \\
  Discipline of Business Analytics\\
  University of Sydney\\
  Darlington, NSW 2006, Australia \\
  \texttt{pablo.monteromanso@sydney.edu.au} \\
   \And
 Rob J Hyndman\\
  Department of Econometrics and Business Statistics\\
  Monash University\\
  Clayton, VIC 3800, Australia \\
  \texttt{rob.hyndman@monash.edu} \\
}

\renewcommand{\headeright}{}
\renewcommand{\undertitle}{}
\begin{document}
\maketitle

\begin{abstract}

Forecasting of groups of time series (e.g. demand for multiple products offered by a retailer, server loads within a data center or the number of completed ride shares in zones within a city) can be approached locally, by considering each time series as a separate regression task and fitting a function to each, or globally,
by fitting a single function to all time series in the set.
While global methods can outperform local for groups composed of similar time series, recent empirical
evidence shows surprisingly good performance on heterogeneous groups.
This suggests a more general applicability of global methods, potentially leading to more accurate tools and new scenarios to study. However, the evidence has been of empirical nature and a more fundamental study is required.
Formalizing the setting of forecasting a set of time series with local and global methods, we provide the following contributions:

\begin{itemize}
\item We show that global methods are not more restrictive than local methods for time series forecasting, a result which does not apply to sets of regression problems in general.
Global and local methods can produce the same forecasts without any assumptions about similarity of the series in the set, therefore global models can succeed in a wider range of problems than previously thought.
\item We derive basic generalization bounds for local and global algorithms, linking global models to pre-existing results in multi-task learning: We find that the complexity of local methods grows with the size of the set while it remains constant for global methods.
Global algorithms can afford to be quite complex and still benefit from better generalization error than local methods for large datasets.
These bounds serve to clarify and support recent experimental results in the area of time series forecasting, and guide the design of new algorithms.
For the specific class of limited-memory autoregressive models, this bound leads to the design of global models with much larger memory than what is effective for local methods.
\item The findings are supported by an extensive empirical study. We show that purposely naïve algorithms derived from these principles, such as global linear models fit by least squares, deep networks or even high order polynomials, result in superior accuracy in benchmark datasets.
In particular, global linear models show an unreasonable effectiveness, providing competitive forecasting accuracy with far fewer parameters than the simplest of local methods.
Empirical evidence points towards global models being able to automatically learn long memory patterns and related effects that are only available to local models if introduced manually.
\end{itemize}

\end{abstract}


\section{Introduction}
\label{sec:intro}

``\textit{Unus pro omnibus, omnes pro uno.}'' --- D'Artagnan

Consider the problem of having to forecast many time series as a group.
We might need to forecast tourist arrivals at all our resorts for next season, the demand for all products we offer at our retailer every week, server loads within a data center, the number of completed ride shares for all zones within a city on a hourly basis, and so on.
The safest way (making fewest assumptions) to approach this problem is to assume each of the time series in the set potentially comes from a different data generating process so it should be modeled individually, as a separate regression problem of finding a function to predict future values from the observed part of the series. This is the standard univariate time series forecasting problem; we call it the \textit{local} approach to the problem.

From a statistical/machine learning perspective, the local approach suffers from one shortcoming: sample size.
Temporal dependence and the short length of the series makes time series forecasting a notoriously difficult problem. Individual time series cannot be modeled in a data-driven way because even basic models (e.g., linear) will suffer from over-fitting.
To overcome over-fitting, an excessive burden is placed on manually expressing prior knowledge of the phenomena at hand (assuming seasonal patterns, looking for external explanatory variables, calendar effects, separate long/short term effects, etc.) and restricting the fitting.
This limits accuracy, since finding appropriate simple models is not always possible; it also limits scalability, since modeling each time series requires human supervision.
As a result, ensembling simple forecast models has been the most widespread approach to automatic time series forecasting (i.e.\ without prior information) for the last four decades.

A univariate alternative to the local approach, called the \textit{global} \cite{salinas2019DeepAR} or cross-learning \cite{smyl2020hybrid} approach, has been introduced to exploit the natural scenario where all series in the set are ``similar'', ``analogous'' or ``related'' (e.g.\ the demand for fiction books follows a similar pattern for all subgenres, stores or cover designs).
The idea behind the global approach is to introduce the strong assumption of all time series in the set coming from the same process, because even when not true, it will pay off in terms of forecasting accuracy.
Global methods pool the data of all series together and fit a single univariate forecasting function \cite{rabanser2020effectiveness}.
The global approach overcomes the main limitation of the local approach: it prevents over-fitting because a larger sample size is used for learning compared to a local counterpart.
On the other hand, the same function must be used to forecast all time series in the set, even when the series are different, which seems restrictive (and less general than local).
The wider adoption of global methods has been prevented by the underlying assumption that they offer benefits over local methods \textit{only} when time series in the set come from intuitively similar or related data generating processes.

However, recent empirical results show puzzlingly good performance of global methods applied over sets of time series that cannot be considered related \cite{laptev2017extreme}.
Notably, top results in the M4 Competition \cite{makridakis2020m4} use some form of globality, mostly in combination with local parts \cite{smyl2020hybrid,montero2020fforma}.
Good performance with purely global methods \cite{gasthaus2019probabilistic, oreshkin2019nbeats} has soon followed these initial results.
Because heterogeneous datasets are general (we can always add series from other sources to get a heterogeneous set), the implications of global methods working well for heterogeneous datasets can have a profound impact in forecasting practice. This can set a new state-of-the-art for automatic methods and open the possibilities of adopting data-driven techniques that are inapplicable for local methods, thus keeping up with the advances in machine learning.

Nevertheless, all cited pieces of empirical evidence are closely tied to the specific models used to achieve them, which differ among themselves, sometimes making contradictory claims about the causes of this performance.
There is a lack of understanding of the underlying general principles behind this good performance (even if they exist).
These principles can help solidify existing knowledge, accelerate technology transfer to real-world applications and guide future improvements in the field.

Motivated by these possibilities and needs, in this paper we compare global and local forecasting algorithms from the perspective of statistics and machine learning applied to a finite number of series, each of finite length, and for a finite horizon of interest. This leads to the following contributions.

\begin{itemize}
\item In a formal setting of forecasting a set of time series, we show that local and global algorithms can produce the same forecasts, so they can approximate the series equally well (Section~\ref{sec:equivalence}).
A local algorithm might learn a different function to forecast each series in the set, but its forecasts can \textit{always} be replicated by a single function learnt by a global algorithm.
Global algorithms in the general setting of supervised learning (i.e.\ pooling together regression problems with arbitrary input-output pairs) do not exhibit this property.
This result can be understood as an existence result that motivates the use of global algorithms for a wider range of problems than previously thought, since we require no assumptions about similarity of the series in the set.

\item We derive generalization bounds of local and global algorithms for forecasting sets of time series (Section~\ref{sec:genbounds}).
These bounds relate model complexity (measured as the number of functions that can be fit to data) and sample size to the difference between training (in-sample) error and testing (out-of-sample) error.
The bounds are agnostic of specific models and the distribution of the data, allowing us to reason without knowing the specifics.
We show that the complexity of local algorithms grows with the size of the set, so it can easily surpass the constant complexity of global algorithms. For example, for sets of moderate size, a local algorithm such as fitting exponential smoothing models to each series in the set is (surprisingly) more complex than a single global deep network with thousands of parameters, therefore the individually simple local method might have poorer generalization.
 These results evidence that global models are a version of muti-task learning \cite{zhang2015multi} and have antecedents in sequence-to-sequence models for forecasting \cite{mariet2019foundations}.
\item Derived from the complexity analysis, we provide principles for designing forecasting algorithms for sets of time series that lead to improved forecasting accuracy.
These principles can serve as a ``toolbox'' which can be adapted to specific forecasting problems.
Furthermore, these principles unify local and global approaches under the common framework of controlling model complexity.
\item We motivate the use of global algorithms with large complexity (Section~\ref{sec:altcomplex}). We highlight three principles:
\begin{description}
  \item[Adding features] For example, adding manual features such as high degree polynomials, or considering alternate model classes such as nonparametric/kernel methods, deep networks or regression trees. This in turn is a strong theoretical support for the current empirical evidence based on deep learning.
  \item[Increasing Memory] Restricting the class of models to finite memory autoregressions, larger complexity of global models can be reached by increasing the memory/order of autoregression (Section \ref{sec:largemem}). In addition to benefits of better fitting, this shows that long memory patterns might be easier to pick with global models.
  \item[Partitioning] We cast the problem of forecasting a set of time series as a spectrum between two extremes based on how the set is partitioned. Partitioning the set into single series leads to the local approach, while the global approach works on the trivial partition that keeps the set intact. We show that the complexity of a learning algorithm can be controlled by the granularity of the partitioning (Section~\ref{sec:parti}). This result highlights the trade-off in sophisticated ways of partitioning, such as data-driven partitioning, leading to clustering and its fuzzy weighted alternatives.
\end{description}

\item We showcase the strength of these theoretically derived principles in a large empirical study on benchmark datasets (Sections~\ref{sec:experisetup}--\ref{sec:experires}). We incorporate these principles in global forecasting methods based on existing classes such as linear models, high order polynomials and deep networks (MLP ReLu). To isolate the contribution of the principles, these methods are \textit{naïve} in the sense that they do not rely on feature engineering, seasonality, advanced deep learning architectures, hyperparameter tuning, preprocessing steps or model selection/combination. These methods achieve good accuracy through a wide range of benchmark datasets. We find compelling empirical evidence towards global methods in the performance of linear models: while theoretically the complexity of a global algorithm must be larger than the one a local algorithm assigns to each individual series, we find that in practice it does not need to be much larger.
This means that global methods can be both simpler (3 orders of magnitude fewer parameters) and more accurate.
\item We provide a theoretical explanation for recent empirical results on the performance of global methods (Section~\ref{sec:related}). The main cause we propose --- global methods can have higher complexity and still generalize better --- differs from the explanations given in related literature, which we show are too narrow. We continue with a critical analysis of related literature claims about clustering, memory and preprocessing from the theoretical point of view of the elements involved in the bounds.
\end{itemize}

\section{Equivalence of global and local approaches for forecasting a set of time series}
\label{sec:equivalence}

``\textit{Each forecast that can be expressed by a local algorithm can also be expressed by a global algorithm.}''

\subsection{Local and global learning algorithms}

We compare two approaches for forecasting a set of time series. The \textit{local approach} fits a function to each time series in the set. The \textit{global approach} fits the same function to all time series in the set. Both approaches are learning algorithms, functions that take data as input and produce forecasting functions as output. These forecasting functions, in turn, take the observed part of a time series as input and produce the future part (forecast) as output.
Because the global approach is restricted to producing the same function for all time series in a set while the local can use a different function for each series, we might think the global is more limited than the local approach. It turns out that under realistic assumptions this is not true: local and global approaches do not differ in the forecasts they are able to produce.

Let $\mathbb{S}$ be the set of univariate time series of size $K$, $\mathbb{S}=\{ X_i \in \mathbb{R}^T, i =1 \dots K \}$, and we allow $\mathbb{S}$ to have repeated time series.  $X_i$ are the observed time series, vectors of real numbers, assumed to be of the same finite length $T$ without loss of generality (e.g. one can carefully pad the shorter series). We are interested in the future of each time series until a \textit{finite} number of time steps $H$; i.e. each time series $X_i$ has a future part which is a vector in $\mathbb{R}^H$.
Forecasting functions are functions from the observed time series to the future part:
$$
  f: \mathbb{R}^T \to \mathbb{R}^H.
$$
Let $A_L$ be a local learning algorithm: a function that takes a time series $X_i$ and returns a forecasting function $f_i$.
That is,
$$
  A_L: \mathbb{R}^T \to (\mathbb{R}^T \to \mathbb{R}^H).
$$
The same local algorithm is applied to each time series in $\mathbb{S}$, producing a different function for each $X_i$, and as an algorithm it must produce the same output given the same input.

A global method $A_G$ is also a learning algorithm, but it only produces a single forecasting function, $g$, for the whole $\mathbb{S}$,
$$A_G: \mathbb{S} \to (\mathbb{R}^T \to \mathbb{R}^H)$$

\begin{proposition}[Equivalence of Local and Global Algorithms for Time Series Forecasting]

Let
\begin{itemize}
\item $\mathbb{A}_L$ = set of all local learning algorithms;
\item $\mathbb{A}_G$ = set of all global learning algorithms;
\end{itemize}
Then

\begin{enumerate}
\item For any $A_G$ in $\mathbb{A}_G$, there exists an $A_L$ in $\mathbb{A}_L$ so that $(A_G(\mathbb{S}))(X_i)=A_L(X_i)$ for all $X_i$ in $\mathbb{S} $
\item For any $A_L$ in $\mathbb{A}_L$, there exists an $A_G$ in $\mathbb{A}_G$ so that $A_L(X_i) = (A_G(\mathbb{S}))(X_i)$ for all $X_i$ in $\mathbb{S}$
\end{enumerate}

\end{proposition}

To prove Proposition~1 we study the two implications separately.

\begin{enumerate}
\item Given the function $g=A_G(\mathbb{S})$, an $A_L$ that sets every $f_i = g$ independent of the series would
produce the same forecasts.
\item In this scenario, we fix the output of the local algorithm, $\{f_i, i=1,\dots,K\}$, and show that there is a $g$ that can approximate them.
We consider two possibilities:

a) $X_i \neq X_j$ for all $i \neq j$.

This means every $X_i$ is unique in $\mathbb{S}$. The set of all $f_i(X_i)$ is finite since $\mathbb{S}$ is finite, with cardinality less than or equal to $K$. Then $g$ must be a function that maps from a finite set to another of less or equal size, and such a function exists. The codomain of $g$ is equal or smaller than its domain, so there is a function that hits every element of the codomain at least once (the pigeonhole principle). We can even explicitly construct $g$ if we are given all the $f_i$, by our favourite universal approximator: polynomials, kernels, etc.

b) $X_i = X_j$ for some $i \neq j$ but $f_i(X_i) \neq f_j(X_j)$

In this case there is no global function $g$ because it would need to produce two different values for the same input, which goes against the definition of a function.
This is why global methods are more restrictive than local for regression problems in general. In the context of supervised learning, regression deals with finding a function mapping from $X_i$, to arbitrary ``targets'' $y_i$, with both $X_i$ and $y_i$ being the inputs to the algorithm. Therefore it can happen that $X_i = X_j$ but $y_i \neq y_j$. A local method would be able to find such $f_i$ and $f_j$ so that $f_i(X_i) = y_i$ and $f_j(X_j) = y_j$, but there is no single function $g$ that can produce such a mapping.
 In time series forecasting, this situation cannot happen because there is no arbitrary mapping to be found. The only inputs to the algorithm are the $X_i$. Because a learning algorithm is also a function, equal input produces equal output. For a local algorithm $A_L$ we have $X_i = X_j \Rightarrow A_L(X_i) = f_i = A_L(X_j) = f_j$, so situation b) is not possible. $\qedsymbol$
\end{enumerate}

\subsection{Finite memory}
\label{sec:finitemem}

Proposition~1 refers to forecasting functions that take the whole series into account, its $T$ observed time periods. It is often the case that interest is restricted to forecasting functions using only the most recent $M$ observations of the series, $M<T$. These are called finite memory
models, autoregressive models or Markov models. In some contexts, $M$ is called the order of the autoregression, the size of the memory or the receptive field. Finite memory models are useful because they deliver a powerful simplification of the world, are easier to fit to data and to analyze. Many natural phenomena can be explained by these models (the ubiquitous ARIMA class is an example of finite memory).

In the setting of Proposition~1, there is no guarantee of the equivalence of local and global approaches if both produce finite memory functions. A simple example are two time series having the same $M$ recent observations, but differing at some other point in the $T$ observed time points. A local approach might produce different functions for these two series since they are different, so their forecast could be different, but any global method will produce the same forecasts for these two series. This is interesting because to guarantee equivalence, the global approach must use longer memory than the local approach. This is a key concept that leads to a guiding principle in the design of global methods: longer memory to approximate the expressive power of local methods.

\section{Generalization relationship between local and global approaches}
\label{sec:genbounds}

The main goal of forecasting is to produce accurate predictions on unseen data according to some error measure.
We have seen in the previous Section~\ref{sec:equivalence} that local and global algorithms can produce the same forecasts,
and therefore potentially achieve the same accuracy, but Proposition~1 does not tell us about how two fixed algorithms compare in practice.

\subsection{Generalization error bounds}

We will study how well local and global algorithms generalize. The goal of studying generalization is to determine about how far the expected error on unseen (out-of-sample) data can be from the error an algorithm makes in the observed (in-sample) data. The difference between out-of-sample error and in-sample error is named \textit{generalization error} or \textit{generalization gap}.

Low generalization error does not imply low out-of-sample error, only that it will be close to the in-sample. High generalization error is usually related to over-fitting. Theoretical analysis of generalization produces results in the form of probabilistic bounds for the generalization error; it gives us the probability that the out-of-sample error lies within a desired distance of the in-sample error. The bound can also be interpreted as a confidence interval for the expected out-of-sample error.

We will use a basic result in machine learning about generalization error \cite{abu2012learning} that relates how far the out-of-sample error, $E_{\textit{out}}$, can be from the in-sample error,
$E_{\textit{in}}$, in terms of an i.i.d.\ dataset of size $N$ and a model $\mathcal{H}$ (also termed hypothesis class).
The model $\mathcal{H}$ is a finite set of functions that can be potentially fit to data.
The size of $\mathcal{H}$, denoted by $|\mathcal{H}|$, is the \textit{complexity} of $\mathcal{H}$.
The basic bound has the form:
$$
  E_{\textit{out}} < E_{\textit{in}} +  \sqrt{ \frac{\log(|\mathcal{H}|) + \log(\frac{2}{\delta})}{2N}},
$$
with probability at least $1 - \delta$.

To simplify notation, we are assuming $E_{\textit{in}}$ to be the in-sample average and $E_{\textit{out}}$ to be its expectation, both calculated over loss values of the predictions of one forecasting function.
This function is chosen from $\mathcal{H}$, picked after exploring all alternatives (e.g.\ the one that minimizes $E_{\textit{in}}$). We can think of a learning algorithm as the way of picking this function from $\mathcal{H}$, though in practice a learning algorithm is tied to a specific $\mathcal{H}$. We assume that the error measure, also named loss function, takes values in $[0,1]$.

To apply this bound we \textit{only} need information about the sample size, maximum and minimum values for the error, and the number of functions which we consider in our model. It is a good compromise of generality: introducing additional assumptions makes the scope too narrow, making it more general leads to vacuous statements.

This bound assumes the worst possible situation with the information we have. It is valid for any distribution of errors taking values in $[0,1]$, in fact the worst distribution is either min or max error with half probability each. The model $\mathcal{H}$ can be any set of functions and the algorithm can pick from it in any way possible, for which the worst case is testing the error of all functions $\mathcal{H}$ before picking one.

The assumption of bounded loss seems limiting but is of great general applicability for what we get from it.
The range $[0,1]$ is arbitrary for notational convenience, when the natural range of the errors is different (e.g. in our experiments), one can apply a scaling factor.
We highlight three situations where, after rescaling, the bound can be applied naturally:
\begin{itemize}
  \item The measurement units have maximum and minimum values, e.g.\ number of patients arriving at a hospital has a clear lower bound and an upper bound can be reasonably defined.
  \item The error measure is relative, e.g.\ sMAPE is between 0 and 200 even when we do not know the bounds of the variable.
  \item We are interested in errors over a certain threshold, in which case we can assume the error is bounded. The bounds assume the worst-case scenario where maximum and minimum errors each have a 50\% chance of happening, so it is a safe worst-case scenario (depending on our needs).
\end{itemize}

More sophisticated notions of complexity other than the size of $\mathcal{H}$ exist, which can make the bound tighter. These require additional information about $\mathcal{H}$ or the way of choosing a function from $\mathcal{H}$. For example, the relation between the number of steps in stochastic gradient descent and generalization error \cite{hardt2016train} could be adapted to get equivalent analyses, provided we know certain properties of our models. In the context of time series, \cite{mariet2019foundations} analyze the sequence-to-sequence models for time series forecasting using the concepts of Rademacher complexity and discrepancy, a notion of the nonstationarity series that includes model and loss function.

The insights derived from our results do not depend on an accurate notion of complexity, since they come from examining \textit{relative} performance. This bound is particularly useful for reasoning at a general level, we do not need information about the distribution of the data (just min and max error values) and about the hypothesis class (as long as we know its finite size).

To use this bound in the context of a time series, where observations are not i.i.d, we need to make the extra assumption of $N$ being an \textit{effective} sample size, which somehow grows with the length of the series $T$, but is less than $T$ in general. This is a feasibility requirement; if we cannot guarantee that the longer we observe a time series, the better we can approximate it, then it is not possible to learn from time series data. See Theorem~1 in \cite{kuznetsov2016forecasting} for a discussion on the notion effective sample size in non i.i.d processes; also \cite{mcdonald2017nonparametric}(Theorem 4) and \cite{kuznetsov2017generalization}.
In our case $N$ ``encapsulates'' or ``abstracts away'' important details about the underlying processes (degree of nonstationarity, dependence structure, etc.) which must be assumed otherwise. Arguably, we use automatic time series forecasting \textit{precisely} because we do not know these details. Moreover, because we are focusing on the relative performance of local and global approaches, it is not important to know the exact effective sample size, as $N$ will be the same for both.

\subsection{Comparing local and global methods on a single time series}

 We can directly apply this bound to compare local and global approaches in the context of out-of-sample error of one isolated time series in the set. We will adopt the common practice of considering one approach is better than the other if its bound is better. This is a heuristic, better bounds do not guarantee better out-of-sample error. We can view this heuristic as a form of penalization for model selection, similar to Akaike's Information criterion, though we will not use it as such.

We can show a result that confirms intuition and the main rationale behind the preference for local methods:
\begin{quote}\itshape
``For a time series in isolation and local and global algorithms with the same hypothesis class,\\
the local approach has a better worst-case out-of-sample error than the global.''
\end{quote}
This occurs because the best local in-sample error will be at most the same as the best global in-sample loss.
The global approach must pick the function from $\mathcal{H}$ that minimizes the in-sample loss of all time series in the set, while the local approach only focuses on the current time series of interest. The remaining terms in the bound are equal for both approaches, so it follows that the bound is better for the local. One could argue that the hypothesis class of the global approach is effectively smaller because it needs to fit to other data. Nothing can be said in this regard at this level of generality, since we are making no assumptions on the distributional relationship \textit{between} series in the set, not even independence. Without the independent series assumption, the set could have been chosen to be adversarial for global methods.
The assumption of independence is necessary in the mainstream machine learning literature in order to make claims about generalization (e.g.\ we are already assuming "approximately" independent observations within a series), so it follows that it should be introduced at the between-series level.

\subsection{Comparing local and global methods on groups of independent time series}

When we introduce the assumption of series in $\mathbb{S}$ being independent and focus on average error across all time series in the set, the comparison changes. The assumption of independence between series is not strong, in the sense that dependent time series can be analyzed from a multivariate point of view, e.g.\ with vector autoregression, rather than with a global model. On the other hand, averaging errors across different time series might be pointless in some scenarios, such as when time series have different scales or are measuring different units.

We name $E_{\textit{in}}^{\textit{Local}}$ the average in-sample error across all time series of the local approach, and $E_{\textit{out}}^{\textit{Local}}$ its out-of-sample expectation.
The equivalent terms for the global approach are $E_{\textit{in}}^{\textit{Global}}$ and $E_{\textit{out}}^{\textit{Global}}$. The local approach uses hypothesis class $\mathcal{H}_i$ for time series $X_i$, and the global approach uses only one hypothesis class, ${\mathcal{J}}$, for the whole set of series.

This scenario results in the following bounds for local and global approaches.

\begin{proposition}[Generalization bounds for local and global algorithms of finite hypothesis class]

Local and global approaches have the following generalization bounds:
\begin{align*}
  E_{\textit{out}}^{\textit{Local}} &< E_{\textit{in}}^{\textit{Local}} +  \sqrt{ \frac{\log\big(\prod_{i=1}^K|\mathcal{H}_i |\big) + \log(\frac{2}{\delta})}{2NK}} \\
  E_{\textit{out}}^{\textit{Global}} &< E_{\textit{in}}^{\textit{Global}} +  \sqrt{ \frac{\log(|{\mathcal{J}}|) + \log(\frac{2}{\delta})}{2NK}},
\end{align*}
with probability at least $1 - \delta$.
\end{proposition}

Assuming independence between time series in $\mathbb{S}$, and effective sample size $N$ in each time series, we can follow the same reasoning that results in the original classical bound. This follows from the Hoeffding inequality \cite{hoeffding1963inequality}, restated here as:
$$
  P(\bar{Y} - \mathbb{E}(\bar{Y}) \geq t ) \leq 2e^{-2Nt^2},
$$
for independent random variables $Y_1,\dots,Y_N$, where $Y_i$ might come from any probability distribution with bounded support in $[0,1]$, $\bar{Y} = N^{-1}\sum Y_i$, and $t > 0$. We can apply the Hoeffding inequality by considering $Y_i$ \textit{equivalent} to the in-sample errors made by one a priori fixed function, in a time series of effective sample size $N$. Then $\bar{Y}$ will be the average in-sample error and its expectation $\mathbb{E}(\bar{Y})$ the expected out-of-sample error.

Instead of fixing the hypothesis, if we pick one hypothesis over all possible in a set $\mathcal{H}$, the bound relaxes. We can apply the union bound (Boole's inequality) to bound the error when picking between many functions, getting:
$$
  P(\bar{Y} - \mathbb{E}(\bar{Y}) \geq t ) \leq |\mathcal{H}|2e^{-2Nt^2}.
$$
This last step is similar to the Bonferroni correction in multiple hypothesis testing.

When we move from one time series to averaging over $K$ independent time series in the set, we get $NK$ samples, assuming for simplicity that all series have the same effective sample size.

For the local approach, the size of the hypothesis class is the size of the Cartesian product of all the local hypotheses $\mathcal{H}_i$ used to fit the series, while for the global approach, the size is $|{\mathcal{J}}|$.
For example, if the local approach picks between three functions for each series ($|\mathcal{H}_i|=3$), it is equivalent to picking between $3^K$ functions for the full set of $K$ series.

For the local approach:
$$
  P(E_{\textit{out}}^{\textit{Local}} - E_{\textit{in}}^{\textit{Local}} \geq t ) \leq \prod_{i=1}^K |\mathcal{H}_i | 2e^{-2KNt^2}.
$$
To get to Proposition~2, we rewrite the right-hand side of the previous inequality as our tolerance level $\delta$.
$$
  \prod_{i=1}^K |\mathcal{H}_i| 2e^{-2KNt^2} = \delta \qquad\Rightarrow\qquad t = \sqrt{ \frac{\log\big(\prod_{i=1}^K|\mathcal{H}_i|\big) + \log(\frac{2}{\delta})}{2NK}}.
$$
So
$$
  P \Bigg (E_{\textit{out}}^{\textit{Local}} - E_{\textit{in}}^{\textit{Local}} \geq \sqrt{ \frac{\log\big(\prod_{i=1}^K|\mathcal{H}_i|\big) + \log(\frac{2}{\delta})}{2NK}} \Bigg ) \leq \delta.
$$
The complementary event corresponds to the result in Proposition~2. The reasoning for the global approach is analogous.

Key to this result is the fact that the Hoeffding bound does not require samples to be identically distributed, so series in $\mathbb{S}$ may vary wildly, can have \textit{any} distribution as long as they are independent (and the error is bounded between 0 and 1). The bound is applicable to \textit{independent copies} of $\mathbb{S}$. To apply it to \textit{new} observations from the same series in $\mathbb{S}$, these should come in the same proportion as the effective sample sizes.

The bounds derived in Proposition~2 allow us to compare how local and global algorithms generalize in a (probabilistic) worst-case sense before we even see the data and hypothesis classes. We can predict how altering the elements involved in these bounds will affect generalization, and propose algorithms inspired on these predictions.

\subsection{Relative complexity of local and global approaches hypothesis classes}
\label{sec:altcomplex}

Imagine we are given a local algorithm and we want to find a global algorithm that has the same or better performance in a given dataset. Under the paradigm (or heuristic) of ``better bounds means better algorithm'', we can try to find a global algorithm that produces the same in-sample error as the local reference, and then from Proposition~2 it follows that the relative performance is totally determined by the respective complexity terms. Proposition~1 becomes relevant now because it shows that a global algorithm exists that can match the errors of a fixed local approach.
In practice, we can think of having a ``budget of complexity'' to reach the in-sample loss equivalent to the local. If we get to it within the budget, we get better generalization than the local alternative and therefore better performance. We aim for equal in-sample loss to isolate the effect of the complexity, which is what we can control before seeing the data.

Assuming $E_{\textit{in}}^{\textit{Local}} = E_{\textit{in}}^{\textit{Global}}$ and $E_{\textit{out}}^{\textit{Local}} = E_{\textit{out}}^{\textit{Global}}$, we can match the two bounds in Proposition~2 to get:
$$
  \prod_{i=1}^K|\mathcal{H}_i |= |{\mathcal{J}}|,
$$
with the $=$ symbol \textit{overloaded} to mean they have the same worst-case guarantees.
We can read this result as saying that the complexity of the local approach grows with the size of the set, while the complexity of the global approach remains constant. Assuming a state-of-the-art local algorithm as baseline (such as an ARIMA fit to each series), a global approach can be at least as complex as the sum of all individual ARIMAs in the set. 

This leads us to prescribe the use of comparatively more complex hypothesis classes for global algorithms. The common ways of increasing complexity is by adding features/variables (such as nonlinear transformations) or changing the model family altogether (e.g.\ from linear to a neural network).
The experiment in Section~\ref{sec:expalter} shows empirical evidence of this principle. This result can serve as an explanation on the recent success of global models, which are generally more complex that the traditional local models. The relationship between global and local models is qualitatively the same as multi-task and single-task learning frameworks \cite{zhang2015multi}, as well as sequence-to-sequence in time series \cite{mariet2019foundations}. The important difference coming from Proposition~1 that global models can approximate as well as local so they can be actually compared in a general setting. We can consider global models as a specific class within the larger framework of multi-task learning and potentially some of the results in the latter could be transferred to the former.

\subsection{Relative memory of local and global autoregressive models}
\label{sec:largemem}

In our analysis of finite memory models (Section~\ref{sec:finitemem}) we saw that global autoregressive methods must have larger memory than local methods to be able to express the same forecasts. Increasing the memory of an autoregressive model can be analyzed from Proposition~2, it increases the complexity of the hypothesis class. It can be seen as a form of featurization more specific to time series than other forms of complexity control.

We can use a reasoning analogous to Section~\ref{sec:altcomplex} to isolate how memory affects generalization. Assume the local and global approaches use autoregressive models (which can be nonlinear), with one parameter per order of autoregression (or size of the memory). Then the complexity of the hypothesis class is completely determined by the order of the autoregression.
Even though in theory there are infinite autoregressive models that can be fit to data (thus the bounds in Proposition~2 become infinitely loose), we can restrict ourselves to the realistic scenario of using double floating point numbers for the parameters. This limits the number of possible models that can be fit, since there are roughly $2^{64}$ possible values for each parameter. If we consider the memory of the local model fit to each series $X_i$ to be $L_i$ and the memory of the global approach $G$, we can explicitly compare the complexities for the local and global algorithms. Substituting in the result of Section~\ref{sec:altcomplex}:
$$
  \prod_{i=1}^K 2^{64L_i} = 2^{64G},
$$
and then taking logarithms and simplifying, we obtain
$$
  \sum_{i=1}^K L_i = G.
$$

\begin{quote}\itshape
``Local and global autoregressive algorithms with the same performance have the same total memory''
\end{quote}

As in Section~\ref{sec:altcomplex}, it follows that the global approach can have much larger memory than that used by a local approach for each time series. While other complexity-controlling procedures such as adding nonlinearities
can be borrowed directly from the common machine-learning literature, this result is more specific to time series.
Proposition~1 indicates that global models need longer memory and we show now that can they afford much more memory than local alternatives. This leads us to prescribe the use of global algorithms with relatively larger memories.
Empirical results of this prescription can be found in Section~\ref{sec:experires}, particularly in Section~\ref{sec:largememexp}, highlighting the importance of controlling this parameter.

\subsection[Relative Complexity of Partitioning S]{Relative Complexity of Partitioning $\mathbb{S}$}
\label{sec:parti}

From a higher level of abstraction, we can see local and global as special cases of an algorithm that acts on a \textit{partition of} $\mathbb{S}$, instead of on $\mathbb{S}$. This algorithm pools together the series in each element of the partition of $\mathbb{S}$, fits a single model within each partition, and forecasts each series in that subset using that function. Therefore, the ``local version'' of this algorithm works on the \textit{atomic} partition ($\{\text{all subsets of } \mathbb{S}\text{ of size }1\}$) and the ``global version'' works on the \textit{trivial} partition ($\left\{\mathbb{S}\right\}$).

The bounds in Proposition~2 can be adapted to this new level of generality. Introducing the cardinality of the partition as $P$, the bounds become:
$$
  E_{\textit{out}}^{Part} < E_{\textit{in}}^{Part} +  \sqrt{ \frac{\log(\prod_{i=1}^P|\mathcal{H}_i |) + \log(\frac{2}{\delta})}{2NK}},
$$
with probability at least $1 - \delta$.
As in previous bounds, to simplify notation we are assuming that all $P$ elements of the partition contain the same number of series. For the trivial partition $P=1$ we recover the global algorithm,
for the atomic partition $P=K$ we recover the local algorithm.

Considering all other possible partitions of $\mathbb{S}$ leads to another way of controlling complexity. A global algorithm may increase its complexity by partitioning, in addition to exploring alternate hypothesis classes or increasing its memory. Similarly, local algorithms can reduce their complexity by pooling together time series in small groups.

Partitioning a set is also specific to time series forecasting, similar to the result in Proposition~1, because we are not interested in forecasting new time series, just the unobserved part of the given set.

These bounds refer to the complexity for a specific partition of $\mathbb{S}$ fixed before seeing the data, in the same sense as the Hoeffding inequality considers a single hypothesis set beforehand. Choosing between different partitions based on data is commonly called clustering and leads to an increase in complexity. While clustering can be analyzed with Hoeffding-based bounds, the worst-case analysis (agnostic of the clustering algorithm) leads to vacuous bounds due to the combinatorial explosion of the number of partitions in a set. Partitioning is also related to multi-task learning (as a form of soft partitioning), discussed in Section~\ref{sec:relatparti}.

We leave clustering outside the scope of the paper. Instead, to showcase the practical usefulness of the principle of partitioning, we propose the most naïve form partitioning: random partitioning into subsets of equal size. Even this simple idea can be effective when the global model cannot reach good in-sample loss levels due to the size of the set.
Section~\ref{sec:exparti} shows empirical evidence of this prescription.

\section{Experimental Setup}
\label{sec:experisetup}

We describe in this section the common elements of the experiments, such as the local and global algorithms we are going to compare, datasets and error metrics.

\subsection{Local benchmark methods}

We consider some methods implemented in the \texttt{forecast} R package \cite{hyndman2008forecast,Rforecast} for the benchmark local methods.
\begin{itemize}
\item \textbf{auto.arima}: Seasonal ARIMA with model selection by AICc.
\item \textbf{ets}: Variants of Exponential Smoothing selected by AICc.
\item \textbf{theta}: Implementation of the Theta method, equivalent to simple exponential smoothing with drift.
\item \textbf{TBATS}: Automatic multi-seasonal state space model with ARMA errors.
\item \textbf{STL-AR}: STL decomposition with AR errors.
\end{itemize}

The first three (auto.arima, ets and theta), are general-purpose methods particularly well-suited for monthly, quarterly and annual data. TBATS and STL will also handle multiple seasonalities such as arise in daily and weekly data. All these methods require the frequency of the series to be pre-specified, unlike the global methods we will consider.

We consider these methods state-of-the art based on the results of the M4 competition \cite{makridakis2020m4}. The top entries in that competition achieved their accuracy through ensembling these methods \cite{montero2020fforma, pawlikowski2020weighted, jaganathan2020combination} or by combining them with neural networks \cite{smyl2020hybrid}.

We do not consider ensembling as a benchmark local method, since ensembling is a technique that can also be applied to global methods. The more accurate the individual models that enter the ensemble, the more accurate the results of the ensemble, so the target accuracy is that of the individual methods.

\subsection{Global methods}

All global methods are based on autoregressive or finite memory models. Each series is lag-embedded into a matrix at the given AR order and then these matrices are stacked together to create one big matrix, achieving data pooling. The different model classes minimize a loss on this large global matrix as in a standard regression problem, with the final column of the matrix representing the target values or dependent variable and the remaining columns the input or predictor variables. All global models produce one-step-ahead forecasts, longer forecasting horizons are achieved through recursive forecasting.

The specific global methods we use are as follows.
\begin{itemize}
\item \textbf{Linear Autoregressive}: A linear model fit by least squares. We consider this method to be the baseline of global models. Linear models fit by least squares are important as a benchmark because they do not include any advanced machine learning technique (such as implicit regularization by SGD, bagging, or ensembling) and overlap the model class with the ARIMA model (a common local approach). Therefore, they are ideal to isolate the effect of globality.
\item \textbf{Featurized Linear Autoregressive}: A linear model fit by least squares with increased complexity by adding nonlinear features in the form of polynomials of degree 2 and 3. This is a relatively naïve way of increasing complexity, and subsumes the linear autoregressive model.
\item \textbf{Deep Network Autoregressive}: Deep Networks are a model class that achieves outstanding results in many applications. The architecture is a ReLu MLP with 5 layers, each of 32 units width, with a linear activation in the final layer, fit by the Adam optimizer with default learning rate. The implementation is Keras \cite{chollet2015keras}. Fitting is done by early stopping on a cross-validation set at 15\% of the dataset (the stacked matrix). The batch size is set to 1024 for speed, and the loss function is the mean absolute error (MAE).
\item \textbf{Regression Tree Autoregressive}: Regression trees are another model class that produces great success in machine learning. Implementation is XGboost \cite{chen2016xgboost}, with default parameters \texttt{subsampling=0.5} and \texttt{col\_sampling=0.5}. Fitting is done by early stopping on a cross-validation set at 15\% of the dataset. The loss function is RMSE and the validation error is MAE.
\item \textbf{Partitioned Linear}: The dataset is initially randomly partitioned into subsets, then a global linear AR model is applied to each subset. This method is a simple representative of the alternative way of increasing model complexity suggested by our theory; i.e. by partitioning instead of featurization or alternative model class.
\end{itemize}

We highlight the simplicity or naïvety of each model. We use no seasonality information, tests for stationarity, batch normalization, dropout, residual layers, advanced architectures such as LSTM or convolutional layers, attention, time dilation, ensembling, etc. No hyperparameter search has been used.

Other notable families of methods, such as nearest neighbors or kernel methods, are left out for computational reasons, but we expect qualitatively equivalent results.

\subsection{Errors and preprocessing}

We focus on MASE as a popular scale-free error measure which rarely divides by zero, and is therefore less prone to numerical errors than SMAPE\@. Furthermore, the scaling done by the MASE can be applied to data as a preprocessing normalizing step, which we apply to each series. Then minimizing the MASE is equivalent to minimizing the mean absolute error in the preprocessed time series. Compared to SMAPE, MASE is easier to fit with off-the-shelf optimizers (it is convex), and it therefore isolates better the effect of the model class rather than the optimization procedure used to fit it.

\subsection{Datasets}

We compare the methods across a collection of datasets which have been used in many peer-reviewed publications as standard benchmarks, from local methods to recent literature for global models. We have added some examples of our own. We loosely and subjectively categorize each dataset as heterogeneous or homogeneous, based on the nature of the domains (single vs multiple domains) and the descriptions by the original publishers.

\begin{itemize}
\item \textbf{M1}: Heterogeneous dataset from a forecasting competition, 1001 series, subdivided into Monthly, Yearly and Quarterly periodicity.
\item \textbf{M3}: Similar to M1, with 3003 time series and an extra ``Other'' category of periodicity.
\item \textbf{tourism}: Homogeneous series from a tourism forecasting competition, subdivided into Monthly, Yearly and Quarterly data.
\item \textbf{M4}: Heterogeneous dataset of 100,000 time series, subdivided into Monthly, Yearly, Quarterly, Weekly, Daily and Hourly periodicities.
\item \textbf{NN5}: From the webpage of the competition \cite{nn5url}: The data consists of 2 years of daily cash demand at various automated teller machines (ATMs, or cash machines) at different locations in England. Homogeneous in data types but heterogeneous in patterns. 111 time series.
\item \textbf{Wikipedia}: Daily visits of pages within the Wikipedia domain, from a Kaggle Competition \cite{wikikaggle}. 115084 time series of 803 observations each. Homogeneous in data type, heterogeneous patterns.
\item \textbf{FRED-MD}: 107 monthly macroeconomic indicators from the Federal Reserve Bank, already differenced and log transformed according to literature consensus \cite{mccraken2018fredmd}. Heterogeneous.
\item \textbf{weather}: Daily weather variables: ``rain'', ``mintemp'', ``maxtemp'', ``solar radiation'' measured at weather stations across Australia. 3010 time series from the Bureau of Meteorology and the \texttt{bomrang} R package \cite{Rbomrang}. Heterogeneous.
\item \textbf{dominick}: 115704 Time series of weekly profit of individual SKUs from a retailer. Homogeneous. From \cite{dominicks}, similar dataset to \cite{gasthaus2019probabilistic}.
\item \textbf{traffic}: Each series in the set is hourly occupancy in a highway lane. Between 0 an 1. Homogeneous.
\item \textbf{electricity}: Hourly time series of the electricity consumption of 370 customers. Homogeneous.
\item \textbf{car parts}: Monthly sales car parts. 2674 series. Jan 1998 - Mar 2002. Intermittent demand. From the \texttt{expsmooth} R package \cite{Rexpsmooth}.
\item \textbf{hospital}: Monthly patient count for products that are related to medical problems. There are 767 time series that had a mean count of at least 10 and no zeros. From the \texttt{expsmooth} R package \cite{Rexpsmooth}.
\item \textbf{CIF2016}: From \cite{vstvepnivcka2017cif}: \textit{``72 time series with monthly frequency, from which 24 were real time series from the banking domain and 48 artificially generated. The real banking domain time series had length from 23 to 69 values and the forecasting horizon was 8 or 16 values.''}. Heterogeneous.
\item \textbf{Pedestrian}: Hourly pedestrian count across different streets in Melbourne. Homogeneous. 66 series ranging from 264 to 35000+  hours.
\item \textbf{Double Pendulum}: Experiment generating a chaotic time series \cite{asseman2018learning}. Positions of the pendulums were extracted from the video at 400Hz, the task is to predict 200 future positions from past positions. Each time series represents a different run of the experiment or the same run sufficiently separated in time. The example of homogeneous series all come from the same ``generative process''.
\item \textbf{COVID19}: Per-region daily cumulative COVID19 deaths, each of the 56 time series representing one region.
Dataset taken from the Johns Hopkins repository, featuring the first 90 days since 22 January 2020. Forecast horizon is the last 14 days.
\end{itemize}

\section{Experiments}
\label{sec:experires}

We show empirical evidence of the principles derived in Section~\ref{sec:genbounds} by comparing forecast accuracy state-of-the-art local methods to global methods. For the sake of clarity we analyze some results from our experiments, a comprehensive summary of experimental results is provided in the Appendix.

\subsection{Large memory}
\label{sec:largememexp}

We first illustrate the effect in forecasting accuracy of increasing the memory of a global model, one of the principles coming from the complexity analysis in Section~\ref{sec:largemem}.
We test this design principle in the most basic class of models, linear models.
Given a dataset, we fit a global linear model for all possible orders of autoregression, and compare the average forecast error against the state-of-the art local models. For each dataset, the maximum order of autoregression is the length of the shortest time series in the set, because it cannot be time-delay embedded at larger orders.

\begin{figure}[!htb]
  \centering
  \includegraphics[width=0.75\textwidth]{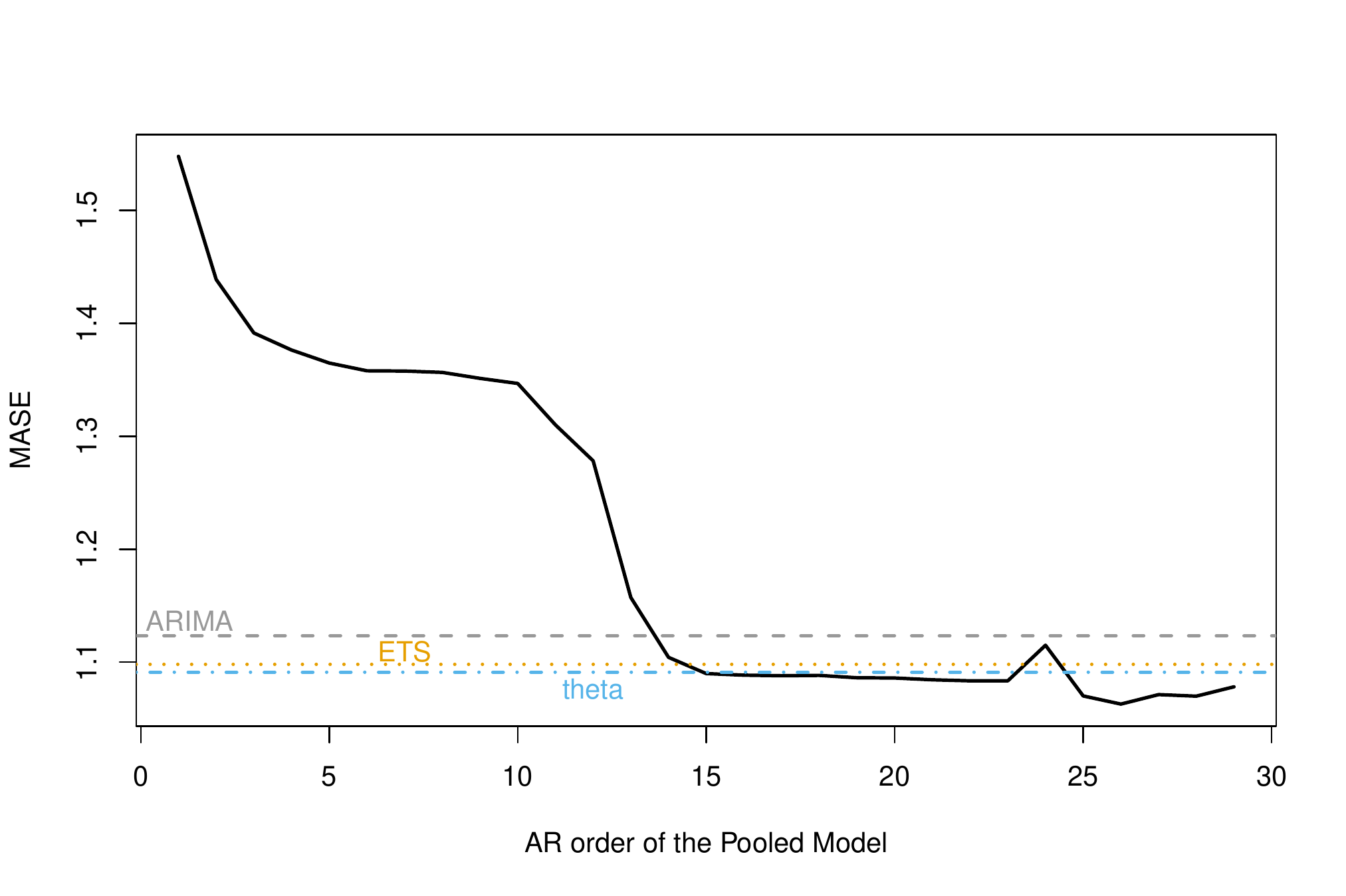}
  \caption{The MASE of a global linear AR model as its memory increases, compared to local methods, for the M1 Monthly data. The forecast accuracy increases as the memory of the global model becomes larger, with noticeable jumps when the memory is close to seasonal periods (multiples of 12 months). A global linear model of order 25 fitted to 617 time series outperforms state-of-the-art local methods, resulting in better forecast accuracy and a simpler overall model.}
  \label{fig:outsampleM1}
\end{figure}

\begin{figure}[!p]
  \centering
  \includegraphics[width=1.0\textwidth]{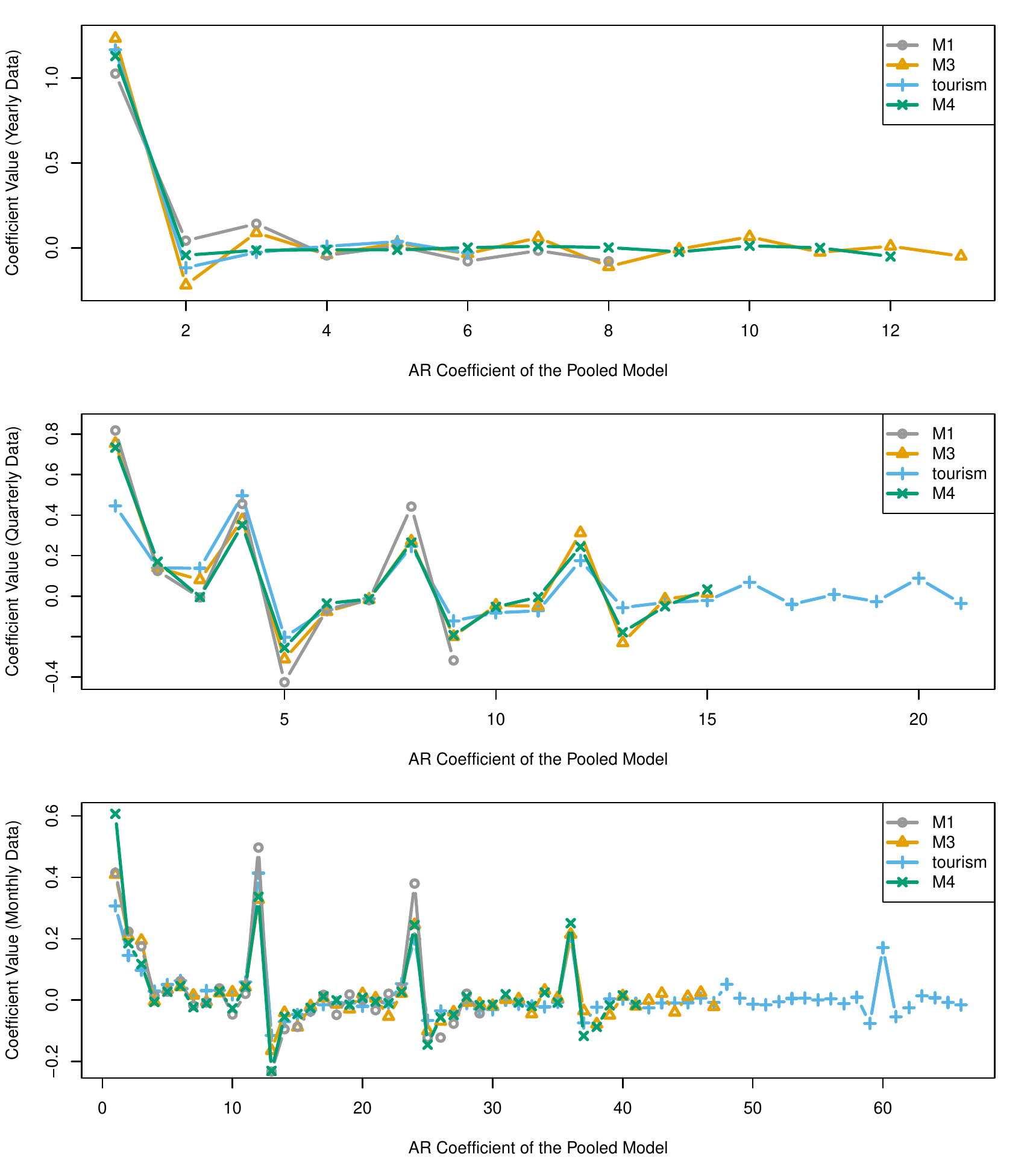}
  \caption{Coefficients of the global linear AR model for Yearly, Quarterly and Monthly series across the M1, M3, tourism and M4 datasets. The fitted coefficients are similar for all datasets. The global linear AR model has large coefficients for long memory levels for both Quarterly and Monthly data.}
  \label{fig:ARcoef}
\end{figure}

We show the results for the M1 Monthly dataset in Figure \ref{fig:outsampleM1}.
We observe a clear pattern of reduction of the out-of-sample error as the order of the global AR model increases (solid black line).
The global model outperforms all local benchmark methods (horizontal lines) at about AR(25).
The M1 Monthly dataset has been available since the 1980s and the technology behind the global model is even older: linear least squares.
This dataset has been extensively used for testing new methods, yet a naïve global method, derived from the basic bounds, is able to outperform state-of-the-art local forecasting algorithms, which leverage sophisticated ideas from time series analysis and are based on accumulated empirical evidence.
The M1 Monthly dataset has 617 time series.
A single AR(25) --- i.e., 25 parameters --- can ``summarize'' the dataset better than models that at least use 1 parameter per series, e.g.\ auto.arima needs approximately 1600 parameters to fit the whole set.
This means a reduction in the number of parameters close to 2 orders of magnitude while achieving better accuracy.
Intuitively, the M1 dataset can be considered heterogeneous; it exhibits different patterns (seasonality, long time trends, structural breaks, etc.) and represents different domains (macro, micro, demographics, etc.).

\paragraph{Longer memory leads global linear models to outperform local}

We extend the analysis to the rest of the datasets in the experimental setting.
As mentioned earlier, when a dataset contains time series of different lengths the maximum memory of a global AR model is limited to the length of the shortest time series. This artificial difficulty can be overcome with recurrent algorithms such as Recurrent Neural Networks, but this would obfuscate the effects of the memory. To study the effects of longer memories we modify the experiment in the following way.
For a given memory level, we analyze on each dataset only the series with length above this level. This methodology is only applied to the M1 Quarterly, M3 Monthly and M4 Quarterly and M4 Monthly datasets.

The experimental results can be seen in the Appendix\@. A global linear model outperforms local approaches for virtually all datasets, including all frequencies of M1, M3, tourism and M4 (Figure~\ref{fig:longmemextra}).
It also holds for the NN5 (Figure~\ref{fig:nn5}), electricity (Figure~\ref{fig:electricity}), traffic (Figure~\ref{fig:traffic}), hospital (Figure~\ref{fig:hospital}), parts (Figure~\ref{fig:parts}), dominick  (Figure~\ref{fig:dominick}), Wikipedia  (Figure~\ref{fig:wiki}), weather  (Figure~\ref{fig:weather}), pendulum (Figure~\ref{fig:pendulum}), pedestrian (Figure~\ref{fig:pedestrian}) and COVID19 (Figure~\ref{fig:COVID19}) datasets.
The exceptions are CIF2016 (Figure~\ref{fig:CIF2016}) and FREDMD-Monthly (Figure~\ref{fig:FREDMD}); even in these datasets the global linear model outperforms two of the three local methods.

The pattern of improving accuracy when increasing memory holds for the majority of datasets. At very long memories, we see some form of convergence in accuracy until the error explodes because even the global model gets into an over-fitting regime. This behavior is exemplified in the hospital dataset, Figure~\ref{fig:hospital}, but would happen for all datasets when the potential maximum memory is much larger than the number of series in the set. This type of over-fitting can be prevented in practice by simple cross-validation or regularization.

Notable exceptions to the pattern of ``longer memory, better accuracy'' are the dominick and parts datasets. We do not have a good explanation of why this happens, but both these datasets are intermittent data. It is also interesting that very low memory orders of a global model outperform local models by a great margin.

Another peculiar effect can be seen in datasets with heavy seasonal components, such as monthly and quarterly periodicities. In these datasets, the accuracy improves with the memory, but then it stops improving or suddenly degrades at memory levels that are multiple of the seasonal frequency. After the memory level is over a seasonal frequency multiple, the accuracy substantially improves. Figure \ref{fig:outsampleM1} clearly illustrates this effect, at around 10 lags the accuracy does not improve and then it improves almost by 30\% between lags 12 and 13. At lag 24 (two seasonal patterns) the accuracy experiences a ``bump'' and then improves again. The electricity hourly dataset (Figure~\ref{fig:electricity} showcases a similar phenomenon, the accuracy improvement stops at lag 24 (1 day cycle) and then improves again. When the memory gets close to a one week cycle (lag 168) the accuracy degrades and after that it gets another big improvement.
We explain this effect as an over-fitting phenomena related to the heterogeneity of seasonal strengths in the datasets. In those datasets that contain both strongly seasonal and non seasonal series (or different seasonal periods), a global model tends to favor the seasonality at memory levels that are multiples of the frequency because it produces good results in-sample. When the memory is over the seasonal period, the global model can use its coefficients to fit both seasonal and non seasonal patterns.

\paragraph{Global models fit long memory patterns}
Fitting long memory patterns with local models in a completely data-driven way is difficult. For example, local models require observing several complete seasonal cycles to discover if a series has strong seasonality. Alternatively, a priori knowledge of the seasonal cycle can be added to the local model. One of the advantages of global models is the ability to pick these patterns without prior knowledge, or requiring less data per series.

We analyze how global models capture long memory by studying the fitted coefficients.
Figure~\ref{fig:ARcoef} shows the coefficients of the global linear model for the M1, M3, tourism and M4 datasets on yearly, quarterly and monthly frequencies. We can see that the coefficients are very similar among datasets, and the effect is consistent across datasets.
The coefficients corresponding to long memory levels are large for quarterly and monthly frequencies, particularly at lags coinciding with seasonal periods, indicating that longer memories get significant contribution to the predictions. There is a slight damping effect for longer memories. The seasonality is being automatically discovered by the global model, and is not part of the model specification. The memory orders that produce the best of accuracy cannot be fitted locally without prior information. For example, to fit a basic local linear AR(40) to a series we would need 80 observations. Realistically, we would need 10--100 times more observations than parameters in a single series to fit it by ordinary least squares. Global methods use the whole dataset to fit these coefficients.

\subsection{Nonlinear models: polynomials, deep networks and regression trees}
\label{sec:expalter}

This experiment considers increasing complexity by changing the model class of global models.
We add nonlinear features such as polynomials of orders 2 and 3, and consider deep networks and regression trees instead of linear combinations of past values.
Polynomials and deep networks represent an incremental or parsimonious increase in complexity with model classes that are a superset of the linear  class.
Regression trees represent a different model family.
We test the nonlinear models for all valid orders of autoregression to isolate the effects of the model class from the effects of expanding memory.

\begin{figure}[!htb]
  \centering
  \includegraphics[width=0.75\textwidth]{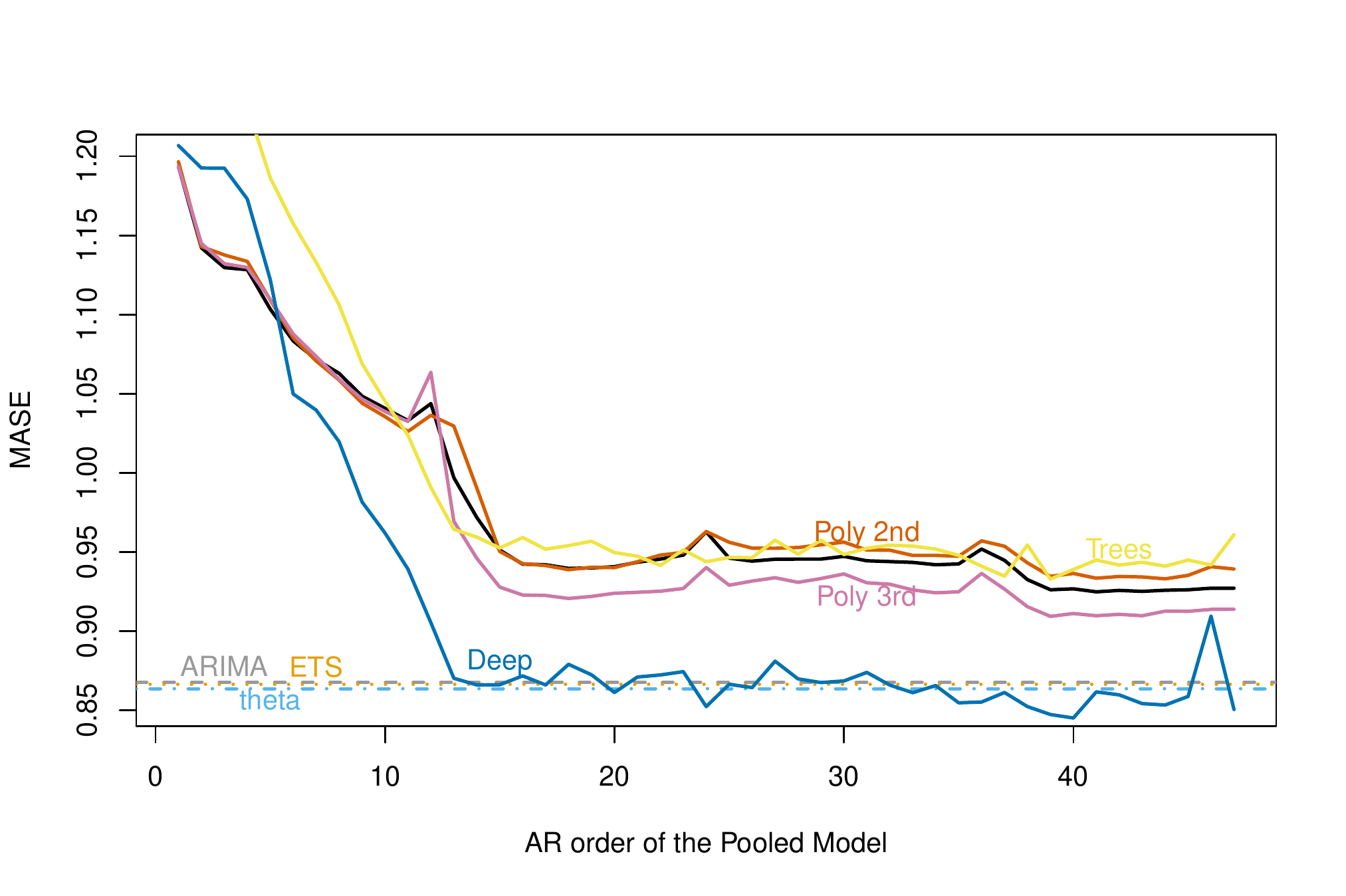}
  \caption{The performance of nonlinear autoregressive models on the M3 Monthly data. Solid lines are the global methods, horizontal dotted lines the local methods. Black:  baseline global linear model; orange: polynomial of degree 2; pink: polynomial of degree 3; yellow: regression trees; dark blue: deep network. The M3 Monthly dataset contains 1428 time series. The global linear model increases accuracy as the AR order increases, but it is limited by the maximum order possible in this dataset. Increasing complexity through polynomials is effective for polynomials of degree 3, which improve accuracy over the baseline global linear for almost all AR orders. Regression trees do not improve over linear, although the performance also increases with memory. The deep network gets the most improvement over linear, reaching a level of accuracy comparable to local.}
  \label{fig:nonlinear}
\end{figure}

Figure~\ref{fig:nonlinear} shows the effects of changing the model class on the M3 Monthly dataset.
The global linear model (black solid line) improves in accuracy as its memory increases until it reaches the maximum order for this dataset (equal to the length of the shortest series).
Unlike the M1 results shown earlier, the global linear model applied to the M3 Monthly data does not get close to the accuracy of local benchmark methods (horizontal dashed lines).
The downward trend pattern of the error is suddenly stopped when we reach the maximum memory level in this dataset, suggesting that the linear model is limited in complexity, it is under-fitting.
Increasing the model complexity by adding nonlinear features in the form of polynomials of degree 2 (orange line) has a small negative impact on accuracy relative to the simple linear model.
A polynomial of degree 3 (pink line) has clearly better performance than the linear model.

Recursive forecasting makes polynomials numerically unstable (e.g.\ an 18-step-ahead forecast for this dataset is equivalent to a 54 degree polynomial) and are not recommended in the literature for automatic time series forecasting. Nevertheless, we show that even polynomials can improve results over linear models when applied globally. Regression Trees (yellow line) are also more complex than linear models but have worse performance in this dataset. They are an example of inappropriate model class, we can detect it in-sample through simple cross-validation. Finally, the Deep Network (dark blue line) also improves over linear, getting the best results of the global models, close to the local models in this scenario. Deep networks also suffer from the problem of instability for recursive forecasting, and the literature recommends direct forecasting (modeling each horizon separately) \cite{taieb2012review}, yet we show here that already recursive forecasting is competitive.

Both polynomials and deep networks demonstrate that nonlinear model classes can improve accuracy.
In this example there is a nonlinear model outperforming the linear models for all values of memory over 8.

We show the results of an extended experiment in the Appendix, applying the same methodology to the M1, M3, tourism and M4 datasets.
Figure~\ref{fig:totalyearly} shows the Yearly results, Figure~\ref{fig:totalquart} the Quarterly, Figure~\ref{fig:totalmonth} the Monthly and Figure~\ref{fig:totalweek} the Weekly. Daily and Hourly are not considered for computational reasons, but results of the global linear can be seen in Figure~\ref{fig:longmemextra} and Figure~\ref{fig:partitioningextra}.

In agreement with the M3 Monthly, global models such as third degree polynomials and deep networks outperform linear models for the majority of datasets (the only exception is Tourism yearly).
Comparing to local models, the nonlinear global models are able to outperform local models in the majority of datasets. The results of global models are especially strong in the large datasets of the M4 competition. In the M4 yearly an AR(12) polynomial of second degree not only outperforms the local benchmarks, it would have scored second in the M4 competition according to its MASE of 3.01.
This means that 24 numbers that specify the model can forecast the 23000 time series of the M4 Yearly better than most of the methods that participated in the competition. For the M4 Quarterly, the deep neural network would achieve third best ranking.

We interpret these results as strong empirical evidence towards globality for the design of new learning algorithms. We can increase the complexity of a baseline global model (linear) in naïve ways (polynomials) and the learning algorithm such as least squares can succeed in finding better forecasting functions. Modern machine learning algorithms such as deep networks produce outstanding results consistently across multiple datasets.

These results should not be interpreted as a recommendation or preference of specific model class over the others, e.g. regression trees could perform better with some hyperparameter tuning.

\par{\textbf{Model complexity vs memory}}
In terms of the effect of memory we highlight two phenomena.
The main is that both sources of complexity, memory and model class, are effective at improving forecast accuracy. When both are included, model class interacts with memory in different ways depending on the dataset, but in general a more complex model class requires less memory to achieve the same level of accuracy than the more simple class.

\subsection{Partitioning}
\label{sec:exparti}

We experiment with another method of increasing complexity: partitioning the set of series. We split the dataset \textit{randomly} into 10 groups of equal size, each group is fit by and forecasted by a global model.

\begin{figure}[!htb]
  \centering
  \includegraphics[width=0.75\textwidth]{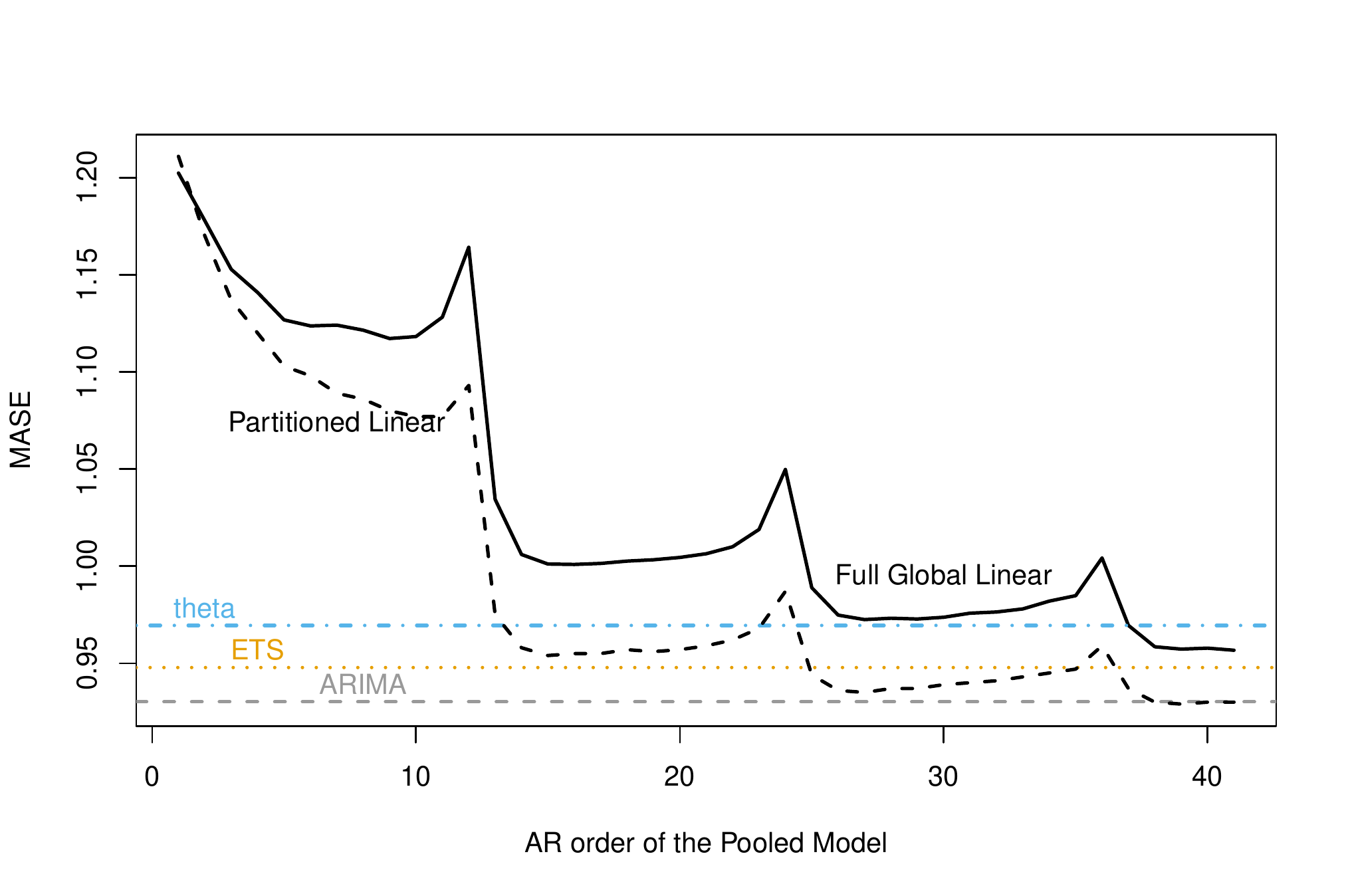}
  \caption{The MASE of global linear models when complexity is increased by partitioning the set. The M4 Monthly is randomly partitioned into 10 parts of equal size, a separate global model is applied to each subset of the partition. Solid line: MASE of a global linear AR; black dashed line: a global linear AR after partitioning.  Partitioned linear models outperform local methods and the non-partitioned global model. An AR of order 20 applied on each partition is already good.}
  \label{fig:partition}
\end{figure}

The implications of the experiment shown in Figure~\ref{fig:partition} are quite interesting, since partitioning is equivalent to \textit{reducing} the sample size, something not usually recommended in statistics and machine learning. Since partitioning is done at random (not clustering), it is strong evidence towards the benefit of raw complexity in a model rather than finding the ``right'' model, and it shows that having similar series is not the cause of the good performance of global methods.
Beyond the possibility of more accurate results, partitioning might also result in faster training times through ``embarrassing'' parallelism.

More examples of this effect for the M1, M3, M4 and tourism data can be found in the Appendix (Figure~\ref{fig:partitioningextra}), but the effect holds (with more or less beneficial impact) in the majority of datasets where the global model is complexity-constrained.

The experiments in this section are meant to showcase the strength of theoretical results because they predict a counter-intuitive effect, they are not a specific recommendation.

\subsection{In-sample vs out-of-sample loss}
\label{sec:insample}
This experiment shows that the ability of simple local methods (e.g.\ ets) to approximate the data generating processes is not the main limiting factor of their performance in many practical situations.
The superior accuracy of global methods is more likely attributable to their better generalization.
This can be considered one of the main predictions from the theory: when we introduce some complexities for specific local and global algorithms in the bounds of Proposition~2, we get that the aggregate complexity of local models is higher than global.

We can expect the following.
\begin{itemize}
\item Local models should get smaller in-sample error than global for comparable model classes (e.g.\ in the linear case, auto.arima and the global linear).
\item Higher complexity implies higher risk of over-fitting, we should see more over-fitting from local models (difference between in-sample and out-of-sample errors).
\end{itemize}

We compare the average in-sample loss that local and global methods produce against their respective one-step-ahead out-of-sample loss for the M1, M3, tourism and M4 datasets.
We use in-sample loss as a measure of the capacity of a given model class to approximate the data.
It measures how well an ideal function within the model class can fit the data, even though in practice we might not be able to find this function.
We focus on one-step-ahead forecast loss, instead of longer horizons, because it is the measure that both local and global methods minimize when training. Theoretical results relate training to testing losses when they measure the same quantity and at the same time we remove the confounding factors of longer horizons.

\begin{figure}
  \centering
  \includegraphics[width=0.6\textwidth]{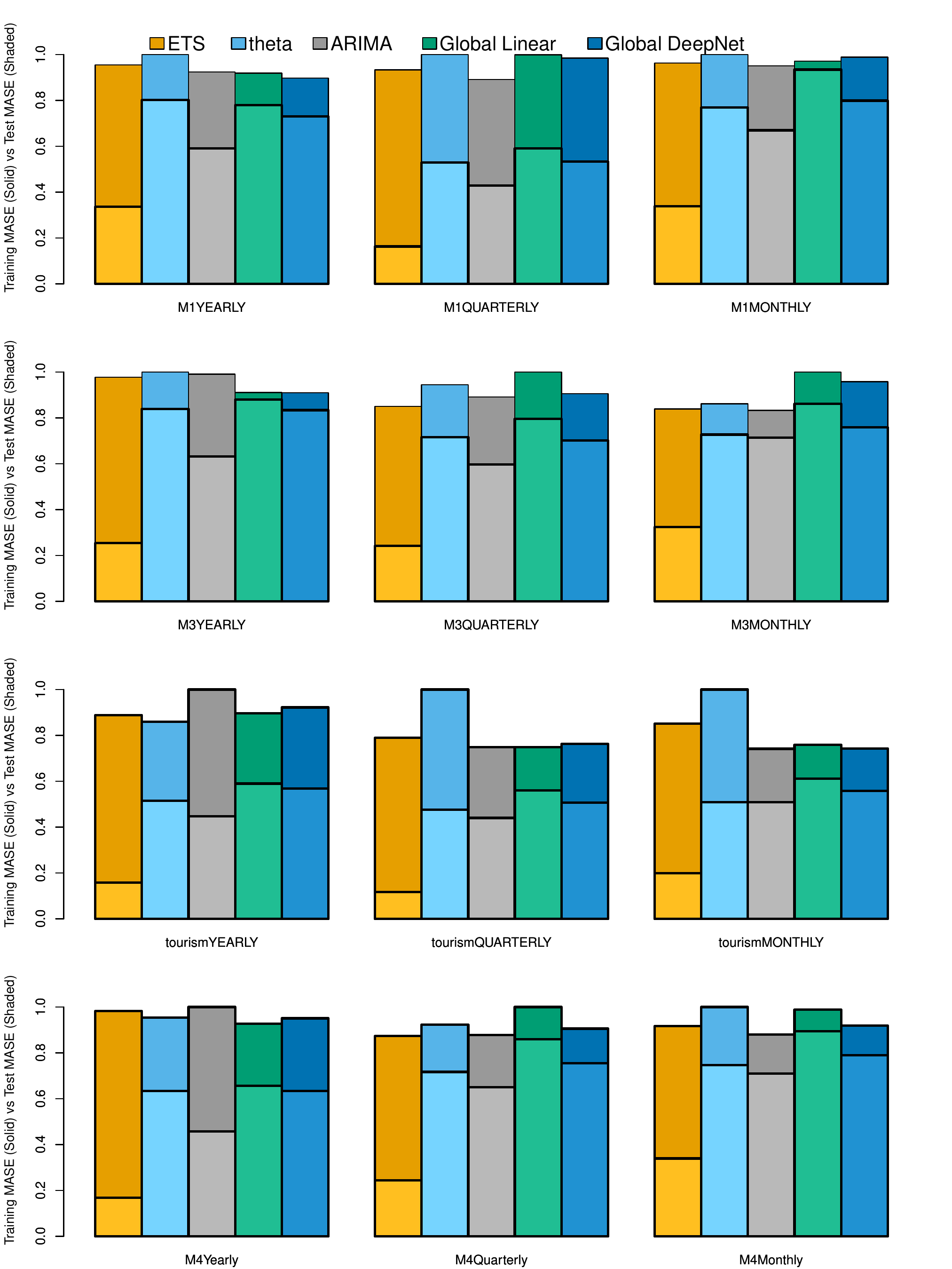}
  \caption{Generalization Error of Local and Global Methods in the M1, M3, tourism and M4 datasets. Measured as in-sample MASE in lighter shaded bars, superimposed over their respective out-of-sample MASE in darker bars, on one-step-ahead forecasts.
  State-of-the-art Local Models (ets, ARIMA, Theta) tend to have better in-sample loss than global models, suggesting they do not suffer from under-fitting. However, generalization error of global models is superior to the local alternatives (in-sample and out-of-sample errors are closer). Even a deep network with thousands of parameters gets worse in-sample error but is competitive out-of-sample. Global methods are in fact simpler than local at per-series complexity.}
  \label{fig:invsout}
\end{figure}

When a model is limited in capacity, it is not able to fit the data well so its in-sample error is large.
We see in Figure~\ref{fig:invsout} the average in-sample error (light bar) superimposed over the out-of-sample error (darker bar) for each method and each datasets. The difference between these two bars is the generalization error. Local models have better average in-sample loss than the global alternatives. Local models suffer from over-fitting, rather than from being unable to fit complex series.
The global linear model exhibits the smallest generalization error, a result that can be predicted by the theory, since it is by far the simplest of the model classes.
Even a global deep network has less difference between in-sample and out-of-sample errors than local models, although it is a quite complex model class.

The more common explanation for global models (see Section~\ref{sec:relglobloc}) is that classical local models cannot pick ``complex patterns'' in the data. However, this experiment suggests a different explanation for the empirical performance of global and local methods. The more likely reason for performance is the over-fitting of local models; we can fit the series well with simple local models but do not know how to control their complexity. For global methods, standard complexity control techniques are more successful or not even needed.

\subsection{Heterogeneous datasets}
\label{sec:heterogexp}

 We show empirical evidence of the capacity of global methods to adapt to heterogeneous datasets by increasing the complexity of the model class. We consider the \textit{full} M4 dataset as the heterogeneous set for our tests.
 When we take the M4 as a single set of time series, we expect is to be highly heterogeneous, because it contains 100K time series from different domains (measuring different units), and at different frequencies (from hourly to yearly measurements).

We compare average out-of-sample error of the one-step-ahead forecasts for the local benchmark methods and the global linear and deep network, fixed at a memory level of AR(12), we are limited by the length of the shortest series in the set.
We add two additional global models that introduce information about the heterogeneity.
\begin{itemize}
\item The first model clusters the data by its frequency before fitting a global model to each cluster.
\item The second model adds information about the frequency and other features \cite{hyndman2019tsfeatures} as extra features of the model in addition to the last 12 observations, but it does not partition the dataset.
\end{itemize}

The purpose is to compare ``homogeneous manual clusters'' with ``heterogeneous with some manual information''. The dataset with the added features is still heterogeneous, and all extra features are functions of the data (but the frequency).

We can see in Table~\ref{tab:clusterheterog} that the ``plain'' global models are outperformed by the local models. Even though they are overshadowed by local models, this result is already interesting because global models outperform naïve forecasting by a large margin (a naïve seasonal method would have a MASE of roughly 1), something that is not trivial. In particular the global linear is fitting 12 numbers to 100000 time series, more than 4 orders of magnitude fewer parameters that an ARIMA.

When we cluster the datasets, the global models substantially improve performance, in particular the deep network already outperforms two of the three local models. These results are to be expected because we know from other experiments that global models outperform local in the M4 for each individual frequency, but here we are severely limiting them in terms of memory.
When we add the extra features, we see that the deep network improves accuracy, outperforming all local and the clustered global. We see this result as strong evidence towards the power of complex model classes for global models; they can automatically find good forecasting functions even in very heterogeneous datasets.

\begin{table}[ht]
\centering
\begin{tabular}{lr}
  \toprule
Method                    & MASE \\
  \midrule
auto.arima                & 0.7618 \\
ets                       & 0.7644 \\
theta                     & 0.7745 \\
Global linear             & 0.9279 \\
Cluster linear            & 0.8559 \\
Global DeepNet             & 0.8149 \\
Cluster DeepNet           & 0.7632 \\
\textbf{Glob. Deep+Feat.} & \textbf{0.7596} \\
   \bottomrule
\end{tabular}
\caption{MASE of one-step-ahead forecasts in the M4 dataset. For local, cluster by period and fully global methods. Clustering by periodicity reduces heterogeneity of the dataset and improves forecast accuracy over naïve global methods. Augmenting a fully global neural network with features produces equivalent results as clustering.}
\label{tab:clusterheterog}
\end{table}

\subsection{Scale normalization as preprocessing}
\label{sec:normalizexp}

We compare the effect of scaling as a means of preprocessing the dataset before applying global models, to show that scale information can be relevant. Scale normalization is one of the main forms of preprocessing applied in global models, however one of the benefits of global models is precisely that they offer a way of modeling information coming from scale in a data-driven way, when local models only have information within each series. For example, a global model can learn that time series in the traffic dataset only take values between 0 and 1, but this information might not exist within a individual traffic time series so it should be manually added to local models.

Scale invariance is appropriate in most benchmark datasets because they have been arbitrarily re-scaled or mix series with different units (e.g.\ tourist arrivals and unemployment rates). Moreover, popular error measures (MASE or SMAPE) are scale invariant, artificially emphasizing the impact of scale normalization. We expect real applications to exhibit none of these issues. Since there seem to be no real reason for scale invariance and some good reasons for not normalizing (or removing scale information), it is interesting to experiment with scale invariance.

In order to keep the ``best of both worlds'' we compare a method that does scale normalization against a method that scale-normalizes but adds the scale of each series back as an additional variable to each series. These two methods are compared to a third that does not normalize the data at all.
 We use the global deep network of AR(12) (we cannot use linear methods because they are scale invariant) as the model for all three methods, the only aspect that changes is whether a scale-normalization preprocessing is added.
 For each of the datasets we measure the error in a one-step-ahead forecast horizon. Table~\ref{tab:scalenorm} shows average errors, both in MASE and also plain mean average error (MAE). The errors of the scale-normalized and the ``scale-normalized with scale features'' are shown relative to the non-scaled version: the error for each dataset is divided by the respective error of the non-normalized method and then averaged across all datasets. 

We can see that scale-normalization has a positive effect relative to the not-normalized version (errors smaller than 1 indicate better perforamnce than the non-normalized), but also adding scale information as an additional feature has a further positive effect in reducing the error. Then non-normalized version also captures scale information but is very sensitive to the influence of series in the set with higher scales, because of their weight in the loss function. 
This experiment shows that scale information can be meaningful, but is not a definitive prescription on how to model it. Specific ways of capturing it will require further work or fine-tuning, for example in the network architecture or loss function, which might allow for bettern performance on the non-scaled version.

\begin{table}[ht]
\centering
\begin{tabular}{lrr}
\toprule
     & Normalization & Normalization + Scale as Feature \\
\midrule
Relative MASE & 0.9497        & 0.9479 \\
Relative MAE  & 1.0213        & 0.9959 \\
\bottomrule
\end{tabular}
\caption{Forecast errors of two types of normalization preprocessing, relative to the non-normalized version. Averaging across datasets. The model is a deep network of AR(12). Adding scale information has a small positive impact in accruracy in both MASE and MAE errors, measured relative to error produced by a raw non-normalized deep network.}
\label{tab:scalenorm}
\end{table}

\section{Discussion and related work}
\label{sec:related}

\subsection{Global and local}
\label{sec:relglobloc}

There is a rich literature on forecasting groups of time series dating back to at least 1962 \cite{zellner1962efficient}, where a form of data pooling is studied. We refer to the excellent paper by Duncan et al. \cite{duncan2001forecasting} for an introduction to the key elements in the literature.

Global methods for forecasting sets of time series can be classified into ``classic'' and ``modern'' categories.
The classic methods include those based on the use of state-of-the-art, well-tested models in combination with data pooling, exemplified by \cite{duncan2001forecasting} and \cite{trapero2015identification}.
Classic methods argue for the use of global models but without increasing the complexity with respect to local models (resulting in simple global models), and support this claim with empirical evidence.
While this prescription is true in some scenarios (e.g.\ Trapero et al. \cite{trapero2015identification} use a global model only for short time series where a basic model is already complex), our work gives a theoretical motivation in the \textit{opposite} direction. Global models can be more complex than individual models in a local algorithm and still benefit from better generalization.
Empirical evidence of this claim appears in Section~\ref{sec:experires}.

Modern methods include those based on the use of recent machine learning models (e.g.\ deep nets) as a global model. Exemplified by \cite{salinas2019DeepAR}, \cite{laptev2017extreme} and \cite{bandara2020cluster}, global models are prescribed with higher complexity than the state-of-the-art local individual models, achieving outstanding empirical performance, though lacking theoretical motivation. Our work serves as a theoretical justification of this prescription: a local algorithm applied to a dataset grows its complexity with the number of series, so a global model should increase its complexity to keep up.

Our results also extend the prescriptions in modern approaches. The main motivation for the use of complex models given in \cite{salinas2019DeepAR} and \cite{laptev2017extreme} is the ability of complex models to pick ``complex patterns'' (e.g.\ ``nonlinear feature interactions'') shared across series in the set. This again might be true in some scenarios, but our work suggests a more comprehensive rationale. In practice, local-and-simple models are able to fit the data better (i.e.\ with lower in-sample loss) than global-and-complex models. In other words, pattern complexity is not a problem for time series forecasting (unlike other learning problems) --- a local linear model with a few parameters can produce results so accurate as to be considered unbelievable. If we want to estimate a complex pattern in a series in ideal conditions of sample size, then we are better off with local models. The (often better) performance of global-and-complex models is more likely due to a better generalization; patterns picked by a global-and-complex method are in fact simpler than the local-and-simple method when taken individually. Evidence in Section~\ref{sec:insample} shows in-sample error of local models being lower than complex global models use, implying they can fit the data better. Therefore, the prescription of the modern literature papers is limiting in the sense that global-and-complex models can be used not only when the set of series involves similar complex patterns. Our theoretical justification shows that it can also be used when there are many unrelated simple patterns (which will happen more often).

Another highlight of our analysis is the emphasis on memory, we show that global models require more memory than local to guarantee the same approximation capacity. This contrasts with the classic global models that assume all time series follow the same model as the local alternative. Longer memories are required for global models even when they have other means of increasing complexity over local alternatives (e.g. by adding nonlinearies).

\subsection{General applicability of global models}

Under the traditional statistical learning framework of consistent estimators and true data generating processes, global methods should not be used because they cannot approximate as well as local models.
A synthetic scenario can be carefully crafted in which \textit{a} global method performs much worse than \textit{a} local method.
Proposition~1, by changing the setting to a more realistic one (finite data and unknown ground truth), challenges the applicability of the traditional framework. Proposition~1 shows that ideally global and local methods are able to produce the same forecasts. It is also specific for time series forecasting, since it does not happen for regression tasks in general. The opposite is also true, the property of consistency of an estimator (or ``universal approximator'' of neural networks) cannot be invoked as a rationale for the use as global models in the traditional setting (no single function exists which can approximate a set of time series in general).

Even though global models were being used years before the motivation from Proposition~1, the value of this theoretical result goes beyond being an interesting curiosity because the restrictive traditional framework is still implicitly carried into practice. For example, the literature on global models encourages its use for groups of time series which are similar or analogous, but not for general sets. From Proposition~1 we can derive that global models can approximate general sets as well as local models, without requiring similarity nor other ``tricks'' such as adding dummy variables to identify each series in the set to be able to fully pick the patterns as in \cite{salinas2019DeepAR} (Section 3.4) or \cite{oord2016wavenet}. While these techniques can and will lead to better performance in some scenarios, they also limit generalization and interpretability, so it is interesting to know that they are not a strict requirement for approximation. For example, adding dummy variables identifying each series in the set to a global linear model with interactions is equivalent to a local linear model.

Proposition~1 can be exploited further because it is valid for any (practical) set of time series, which in fact means that successful global models can be applied even to single series by artificially adding \textit{any} series to form a set. A clear consequence is data augmentation: series can be added to a set while guaranteeing that a global algorithm exists which fits the augmented set as well as a local model fits the original set.

We hope that Proposition~1 will enable exploration of ideas that may have been subconsciously inhibited by the restrictive traditional framework, encouraging a deeper study of global models and the proposal of new global learning algorithms.

\subsection{Partitioning, relatedness and clustering}
\label{sec:relatparti}

Global models have been introduced to address the problem of forecasting sets of series which are ``related'', ``analogous'' or have ``similar patterns'', but not for general sets \cite{duncan2001forecasting, salinas2019DeepAR, bandara2020cluster}. This notion of relatedness has been introduced mostly through examples, but never formally defined. In order to achieve or increase relatedness of a set, several notions of ``clustering by relatedness'' have been introduced.
The most prevalent clustering type is judgemental clustering \cite{duncan2001forecasting}. Judgemental clustering is applied manually and is based on natural characteristics of the series, such as what units of measure (sales of products and precipitation series belong to different groups), or the frequency of the series (monthly, daily, quarterly series in separated groups). Algorithmic clustering (as in as k-means or partitioning against medoids) based on features of the series (e.g.\ autocorrelation coefficients) have also been applied \cite{bandara2020cluster}. Both forms of clustering often result in improved forecasting accuracy, but their nature is mostly intuitive or heuristic (there is no general guarantee that they affect forecasting accuracy) and as far as they impose a reduction of sample size they can also be detrimental.

Our work can serve to analyze relatedness and clustering.
First, from Proposition~1 we show that relatedness, and therefore clustering, is not essential to forecasting sets, we can think of clustering+modelling as another type of model. The bounds of Proposition~2 and Section~\ref{sec:parti} show clustering as a form of set partitioning and inform us about the trade-offs compared to other methods (such as purely local or purely global).

Second, the formalism of Proposition~2 and Section~\ref{sec:parti} allows us to define a concrete notion of relatedness. The elements involved in relatedness are clearly identified: a partition of a set, a learning algorithm (including model class), and a loss function. We cannot measure the relatedness of a partition without specifying the forecasting model. Given a set of time series and two ways of partitioning it, say $P_1$ and $P_2$, partition $P_1$ is more related than $P_2$ if its bound is better. When $P_1$ and $P_2$ have all sub-elements of equal size, it reduces to which has better in-sample error.

This view of partitioning links to the main formulation on multi-task learning \cite{thung2018brief}, with each of the time series representing a task. It is an extension of partitioning or clustering: prior information on task similarity is expressed through a weight matrix, a weighted graph which can represent every partition from global to local, clusters of the set and even soft clustering.
From this point of view, we can see clustering as a ``layer'' or preprocessing step in our architecture which infers this weight matrix either manually or from features. Compared to increasing the complexity of the model, clustering has less flexibility and does not directly minimize the loss function. In a way, any extra information (either judgemental or features) that is considered by the clustering step can be added via data augmenting to a fully data-driven, end-to-end algorithm and let the data speak for itself. We show in Section~\ref{sec:heterogexp} that a global deep network model is able to produce good forecasts for a highly heterogeneous dataset without requiring judgemental or algorithmic clustering; it is in fact able to fit the data surprisingly well. With the methodological advances in machine learning, computing power and the increase in the availability of data, we expect that general-purpose machine learning models will get diminishing benefits from adding ad~hoc clustering methods on top of them.

\subsection{Modern practices and model classes for global methods}

We now provide some critical analysis of specific recommendations in ``modern'' approaches based on our results.

\begin{itemize}
\item \textbf{Model Classes}: Most ``modern'' methods prescribe the use of a specific deep network architecture. Our work decouples the complexity from a particular model, and therefore motivates and justifies the use of alternative model classes (e.g.\ trees in \cite{mafildes2019customer}, featurization, kernel methods or nearest neighbors). The justification increases the ``toolbox'' available to practitioners. For example, gradient boosted decision trees also benefit from the theoretical properties presented here, and have been known to produce very good results in forecasting competition, such as the GEFCom2014 \cite{landry2016probabilistic} and the recent M5 \cite{M5kaggle}, in particular when adding external information.

\item \textbf{Memory}: Many Global Models are based on Long Short-term Memory (LSTM) or other Recurrent Deep Network Architectures. Despite their name, these methods have difficulties learning long memory patterns without manual tuning \cite{trinh2018learning}. One common way of tackling this limitation is the use of larger input window, making them more similar to a purely autoregressive network. We show even non-recurrent, global AR models can successfully fit long explicitly specified memory orders, and we expect that the concept of really long input windows within a global method combined with a recurrent architecture can produce benefits while retaining the good properties of recurrence. In Section~\ref{sec:largememexp} and the Appendix we show the effect of long memory on forecasting accuracy. Other recent architectures such as temporal convolutions with time dilation \cite{oord2016wavenet, borovykh2018dilated} also go in the direction of fitting really large memory levels. Convolutions on the other hand are successful because they substantially reduce the capacity of the model class, so there is the danger of underfiting when we move towards larger, more heterogenous datasets.

\item \textbf{Scale Normalization as Preprocessing}: Several forms of preprocessing have been recommended for global methods \cite{duncan2001forecasting, salinas2019DeepAR, smyl2020hybrid, hewamalage2019recurrent, rabanser2020effectiveness}. They are considered an important component of the performance of the methods. The main such preprocessing step is normalizing the scale of each series: each series is transformed by removing its scale before model fitting and forecasting, then the forecasts are back-transformed. Scaling can help deal with outlier series (e.g. one series has much larger scale than the rest, dominating the fitting) and in the case of deep networks makes them faster to train.

This form of scaling the series locally makes \textit{any} model scale invariant, a strong assumption which reduces the complexity of the model. Interestingly, it is one of the two properties that makes linear maps. The trade-off therefore is between how relevant is the information about scale compared to how it reduces over-fitting or deals with outliers, but in general we know that time series forecasting is not scale invariant. For example, the series in the traffic dataset have a clear maximum value of 1 implying maximum occupancy in the lane. When series in this dataset take values close to this maximum we know that they cannot go beyond that value and this is valuable information for forecasting. After scaling, series close to the maximum value may look exactly the same as those in the middle range, so this relevant information is lost. On the other hand, there are example of processes which would clearly benefit from scale normalization, for example, if our set of time series is comprised of exponential growth patterns, each time series can be scale-normalized without loss of information, and the pattern will be much easier to pick for a global model.  

Coincidentally, the errors commonly used for benchmarking (such as SMAPE or MASE) are scale invariant, so this form of preprocessing also helps for optimizing the correct loss function. However many real situations will not need this scale invariant error measures, so scale invariance becomes a way of complexity reduction. Because as we have show global models can afford to be quite complex, it is worth exploring alternative ways of dealing with specific problems that are solved by scale normalization (such as the weight of outliers, training speed, etc.) rather than making everything scale invariant. An experiment on the effect of scaling can be found in Section~\ref{sec:normalizexp},  showing that scaling can convey important information in some cases, it is not a definite argument on the use of normalization.

\item \textbf{Seasonality}: Another form of preprocessing is modeling seasonal patterns individually, subtracting them from each series and then modeling the residuals with a global method. A rationale for this form of preprocessing can be found in \cite{smyl2020hybrid}: ``Neural networks do not learn different seasonality'' (page 4) and in \cite{bandara2019lstm}. Our empirical results suggest that global neural networks do not suffer from an inability to pick seasonal patterns, and the theoretical justification of previous empirical evidence points towards the models used there not being complex enough or having limited memory. Removing seasonality from the data can erase some relevant information which might be exploited by global models, as with scale normalization.

\end{itemize}

\subsection{Multivariate, Covariables and Sequence models}

There is a large overlap between global models and multivariate forecasting models, because both work on groups of time series.
In practical terms, global models are univariate time series models, they only require observations of a single time series for forecasting, they use the group for finding good predictive functions. Consequently, global models are more generally applicable because they do not require observations from multiple time series \textit{at the time of forecasting}.
Multivariate models work on groups that are supposed to have some form of interaction or joint evolution, global models work on any group. For example, forecasting a pandemic with a global model can benefit from time series measured at different historic periods (separated by decades), controlled experiments or even different pathogens, while a multivariate model is limited to the evolution of the current one and factors that influence it. Another example could be electricity consumption patterns, a global model could pool data from different places around the world to find a shared pattern, while it is clear that completely unrelated households at different corners of the world will not benefit from a multivariate approach that used their joint evolution.

When there is inter-dependence (mutivariate) or depencence (covariate and a main time series) in the dataset, global methods do not capture it directly, though they could help it in capturing it in a implicit manner, by the same reasoning that allows time series to be predicted from their own past observations (time-delay embedding) even when they depend on other unobserved processes. This effect is of course not unique to global models, but they can contribute in that they could help in picking more complex patterns that local alternatives.

From a more theoretical point of view, the generalization of global models that we show here assumes groups of independent time series, under heavy dependence global models loose their beneficial performance guarantees (though practicioners could still try them). Multivariate methods specifically deal with this strong dependence case through heavily reducing the complexity, either by considering that the joint group of time series can be represented at a much lower dimensionality or through other forms of regularization \cite{sen2019think, wang2019deep}.

Ultimately globality and multivariate are kind of orthogonal concepts, globality representing a ``same pattern of evolution'' and multivariate representing ``joint or dependent evolution''. For example, one can have a multivariate global model, i.e. a group of different multi-variable processes that are assumed to evolve in the same way. For example, pairs of energy consumption per household and income per household time series represent a multivariate process, and we can have a dataset with many of these multivariate processes, for example households sampled all around the planet. Each multivariate process does not necessary affect the other, but we could assume that they follow the same pattern. Our results can even be adapted to this more general case by assuming that all time series in a multivariate joint has a finite support, so the multivariate can be encoded in a single number producing a univariate time series.

The main property of equal approximation of global and local methods that we show for the univariate case will be inherited by the multivariate global and more interestingly by the case of univariate time series with covariates. The latter case highlights the difference between time series and other regression scenarios, even sequence-to-sequence. If we want to forecast a main sequence based only on other sequences, without using the main sequence series as input to the prediction function, we are in a standard regression and there is no guarantee that global will be able approximate as well as local. For example, we want to predict sales of ice-cream based on temperature data and sales of umbrellas based on temperature data by pooling both time series in a global model. It is clear that these two time series are different functions of the temperature, so a global model will not be able to approximate as well as a local one. We we add the main time series as the extra input, we get the equal approximation capacity, but the benefit comes from the principle that we show in Proposition~1 for the univariate case.

\section{Conclusions}

We have shown that global and local methods for forecasting sets of time series are equally general. Globality is neither restrictive nor requires ``similarity'' or ``relatedness'' in the set. We have then compared the relative complexities of local and global methods, showing that global methods can afford to be more complex than local because they generalize better.
The combination of these two findings strongly supports recent empirical results and motivates further exploration of modern machine learning methods for time series forecasting.

The theoretical symmetry between global and local models can however be broken by considering some practical issues.
\begin{itemize}
\item While better generalization of global models does not imply better performance, it is already useful because it makes it easier to detect when an algorithm fails (e.g.\ by cross-validation or simply because in-sample loss does not reach satisfactory levels). Arguably, this benefit is even more important in time series forecasting than in other more stable machine learning problems (see \cite{makridakis2018statistical}, Table 10).
\item Global models can increase model complexity compared to individual local models.
 Modern machine learning has developed an impressive array of techniques for increasing and controlling model complexity (e.g.\ deep networks, regularization) that can be applied to global models. Meanwhile finding the right simple models to fit locally requires domain knowledge or prior information, which can be difficult to obtain.
On the other hand, when prior information is already available, we can usually express it only through local models.
\end{itemize}

Global models for time series forecasting is a relatively new field, and there is an opportunity for researching ways of expressing prior information for global models.
Ultimately, the seemingly superior performance of global models is a question of context --- most problems for which we have simple, domain-based local models are not usually considered for automatic forecasting because they are already solved. For example, we use automatic forecasting for sales because of the lack of domain understanding, but for weather forecast we have domain-based models so we do not consider this dataset for benchmarking data-driven univariate forecasting.

Our complexity analysis comes from the derivation of basic generalization bounds.
The bounds can be used to identify the main elements involved in the generalization of globality and locality at a general level.
From the bounds we derived three parsimonious principles for designing new algorithms.
The first principle points towards increasing complexity of model classes for global methods (nonlinearities, deep networks, etc.) and serves as a strong motivation and support for recent results.
The second principle of increasing the memory for global methods has strong implications for forecasting by showing that global models might be able to incorporate larger memory much more effectively.
The third principle of partitioning a group of series bridges the gap between local and global approaches, and serves to link the related literature, highlighting the hidden costs of methods such as clustering. Beside the basic principles, the proposed framework can be used to analyze concepts such as relatedness, specific architectural decisions such as scale-normalization, recurrence or manually added seasonality, and extends the interpretation of the success of global methods to more heterogeneous settings.

Because of their generality, the bounds are necessarily loose; their benefit comes from the insights derived from their analysis.
Tools are needed to navigate the space of all possible models or architectures \textit{before we see the data}, otherwise it becomes impossible to do it efficiently through trial-and-error and we are at the mercy of spurious results.
For tighter bounds, particularly in time series where effective sample size is difficult to estimate, we recommend cross-validation \cite{tscv}.
Cross-validation is powerful at the global level, and current state of technology relies on testing many different models and hyperparameters, which is safe to do globally (unlike locally).

We have presented ample empirical evidence comparing global and local models and the algorithms we have suggested.
We show that a parsimonious increase in complexity of global models makes them highly competitive, outperforming local state-of-the-art on most datasets.
In our estimation, the strongest empirical claim comes from the linear case.
 While outperformed by advanced methods such as deep networks, global linear models achieve comparable or superior accuracy to state of the art local models, but require \textit{orders of magnitude fewer parameters than the competing local alternatives}.
It is both simpler and more accurate.
There is no danger of data snooping or cherry picking, we are showing every possible linear model that can be fit to each dataset, and the results present a consistent picture.
Linear models are well understood, it is trivial to create situations where global linear models fail, but it seems that these situations are difficult to find in practice.
From this fact, it is easy to believe that a global model which is slightly more flexible than linear will produce even better results.
The strength of the empirical result challenges the notion of ``relatedness'' or heterogeneity of datasets with thousands of time series such as the M4.

\bibliographystyle{unsrt}
\bibliography{global_forecast}  

\begin{thebibliography}{10}

\bibitem{salinas2019DeepAR}
David Salinas, Valentin Flunkert, Jan Gasthaus, and Tim Januschowski.
\newblock Deepar: Probabilistic forecasting with autoregressive recurrent
  networks.
\newblock {\em International Journal of Forecasting}, 36(3):1181--1191, 2020.

\bibitem{smyl2020hybrid}
Slawek Smyl.
\newblock A hybrid method of exponential smoothing and recurrent neural
  networks for time series forecasting.
\newblock {\em International Journal of Forecasting}, 36(1):75--85, 2020.

\bibitem{rabanser2020effectiveness}
Stephan Rabanser, Tim Januschowski, Valentin Flunkert, David Salinas, and Jan
  Gasthaus.
\newblock The effectiveness of discretization in forecasting: An empirical
  study on neural time series models.
\newblock {\em arXiv preprint arXiv:2005.10111}, 2020.

\bibitem{laptev2017extreme}
Nikolay Laptev, Jason Yosinski, Li~Erran Li, and Slawek Smyl.
\newblock Time-series extreme event forecasting with neural networks at uber.
\newblock In {\em International Conference on Machine Learning}, volume~34,
  pages 1--5, 2017.

\bibitem{makridakis2020m4}
Spyros Makridakis, Evangelos Spiliotis, and Vassilios Assimakopoulos.
\newblock The {M4} competition: 100,000 time series and 61 forecasting methods.
\newblock {\em International Journal of Forecasting}, 36(1):54--74, 2020.

\bibitem{montero2020fforma}
Pablo Montero-Manso, George Athanasopoulos, Rob~J Hyndman, and Thiyanga~S
  Talagala.
\newblock {FFORMA}: Feature-based forecast model averaging.
\newblock {\em International Journal of Forecasting}, 36(1):86--92, 2020.

\bibitem{gasthaus2019probabilistic}
Jan Gasthaus, Konstantinos Benidis, Yuyang Wang, Syama~Sundar Rangapuram, David
  Salinas, Valentin Flunkert, and Tim Januschowski.
\newblock Probabilistic forecasting with spline quantile function {RNNs}.
\newblock In {\em The 22nd International Conference on Artificial Intelligence
  and Statistics}, pages 1901--1910, 2019.

\bibitem{oreshkin2019nbeats}
Boris~N Oreshkin, Dmitri Carpov, Nicolas Chapados, and Yoshua Bengio.
\newblock N-beats: Neural basis expansion analysis for interpretable time
  series forecasting.
\newblock In {\em International Conference on Learning Representations}, 2019.

\bibitem{zhang2015multi}
Yu~Zhang.
\newblock Multi-task learning and algorithmic stability.
\newblock In {\em Proceedings of the Twenty-Ninth AAAI Conference on Artificial
  Intelligence}, pages 3181--3187, 2015.

\bibitem{mariet2019foundations}
Zelda Mariet and Vitaly Kuznetsov.
\newblock Foundations of sequence-to-sequence modeling for time series.
\newblock In {\em The 22nd International Conference on Artificial Intelligence
  and Statistics}, pages 408--417, 2019.

\bibitem{abu2012learning}
Yaser~S Abu-Mostafa, Malik Magdon-Ismail, and Hsuan-Tien Lin.
\newblock {\em Learning from data}, volume~4.
\newblock AMLBook New York, NY, USA:, 2012.

\bibitem{hardt2016train}
Moritz Hardt, Ben Recht, and Yoram Singer.
\newblock Train faster, generalize better: Stability of stochastic gradient
  descent.
\newblock In {\em International Conference on Machine Learning}, pages
  1225--1234, 2016.

\bibitem{kuznetsov2016forecasting}
Vitaly Kuznetsov and Mehryar Mohri.
\newblock Forecasting non-stationary time series: From heory to algorithms.
\newblock Technical report, 2016.
\newblock \url{https://cims.nyu.edu/~vitaly/pub/fts.pdf}.

\bibitem{mcdonald2017nonparametric}
Daniel~J McDonald, Cosma~Rohilla Shalizi, and Mark Schervish.
\newblock Nonparametric risk bounds for time-series forecasting.
\newblock {\em The Journal of Machine Learning Research}, 18(1):1044--1083,
  2017.

\bibitem{kuznetsov2017generalization}
Vitaly Kuznetsov and Mehryar Mohri.
\newblock Generalization bounds for non-stationary mixing processes.
\newblock {\em Machine Learning}, 106(1):93--117, 2017.

\bibitem{hoeffding1963inequality}
Wassily Hoeffding.
\newblock Probability inequalities for sums of bounded random variables.
\newblock {\em Journal of the American Statistical Association},
  58(301):13--30, 1963.

\bibitem{hyndman2008forecast}
Rob Hyndman and Yeasmin Khandakar.
\newblock Automatic time series forecasting: The forecast package for {R}.
\newblock {\em Journal of Statistical Software, Articles}, 27(3):1--22, 2008.

\bibitem{Rforecast}
Rob Hyndman, George Athanasopoulos, Christoph Bergmeir, Gabriel Caceres, Leanne
  Chhay, Mitchell O'Hara-Wild, Fotios Petropoulos, Slava Razbash, Earo Wang,
  and Farah Yasmeen.
\newblock {\em {forecast}: Forecasting functions for time series and linear
  models}, 2020.
\newblock R package version 8.12. \url{http://pkg.robjhyndman.com/forecast}.

\bibitem{pawlikowski2020weighted}
Maciej Pawlikowski and Agata Chorowska.
\newblock Weighted ensemble of statistical models.
\newblock {\em International Journal of Forecasting}, 36(1):93--97, 2020.

\bibitem{jaganathan2020combination}
Srihari Jaganathan and P.K.S. Prakash.
\newblock A combination-based forecasting method for the {M4}-competition.
\newblock {\em International Journal of Forecasting}, 36(1):98 -- 104, 2020.
\newblock M4 Competition.

\bibitem{chollet2015keras}
Fran{\c{c}}ois Chollet et~al.
\newblock Keras: Deep learning library for theano and tensorflow, 2015.
\newblock \url{https://keras.io}.

\bibitem{chen2016xgboost}
Tianqi Chen and Carlos Guestrin.
\newblock Xgboost: A scalable tree boosting system.
\newblock In {\em Proceedings of the 22nd acm sigkdd international conference
  on knowledge discovery and data mining}, pages 785--794, 2016.

\bibitem{nn5url}
{NN5} forecasting competition for artificial neural networks and computational
  intelligence, 2018.
\newblock \url{http://www.neural-forecasting-competition.com/NN5/}.

\bibitem{wikikaggle}
Web traffic time series forecasting, 2017.
\newblock \url{https://www.kaggle.com/c/web-traffic-time-series-forecasting}.

\bibitem{mccraken2018fredmd}
Michael~W. McCracken and Serena Ng.
\newblock Fred-md: A monthly database for macroeconomic research.
\newblock {\em Journal of Business \& Economic Statistics}, 34(4):574--589,
  2016.

\bibitem{Rbomrang}
Adam~H. Sparks, Jonathan Carroll, James Goldie, Dean Marchiori, Paul Melloy,
  Mark Padgham, Hugh Parsonage, and Keith Pembleton.
\newblock {\em {bomrang}: Australian Government Bureau of Meteorology (BOM)
  Data Client}, 2020.
\newblock R package version 0.7.0.
  \url{https://CRAN.R-project.org/package=bomrang}.

\bibitem{dominicks}
Dominicks dataset, 2019.
\newblock \url{https://www.chicagobooth.edu/research/kilts/datasets/dominicks}.

\bibitem{Rexpsmooth}
Rob~J Hyndman.
\newblock {\em expsmooth: Data sets from "Exponential smoothing: a state space
  approach" by Hyndman, Koehler, Ord and Snyder (Springer, 2008)}, 2020.
\newblock R package version 2.4. \url{http://pkg.robjhyndman.com/expsmooth}.

\bibitem{vstvepnivcka2017cif}
Martin {\v{S}}t{\v{e}}pni{\v{c}}ka and Michal Burda.
\newblock On the results and observations of the time series forecasting
  competition cif 2016.
\newblock In {\em 2017 IEEE International Conference on Fuzzy Systems
  (FUZZ-IEEE)}, pages 1--6, 2017.

\bibitem{asseman2018learning}
Alexis Asseman, Tomasz Kornuta, and Ahmet Ozcan.
\newblock Learning beyond simulated physics.
\newblock In {\em Modeling and Decision-making in the Spatiotemporal Domain
  Workshop}, 2018.

\bibitem{taieb2012review}
Souhaib~Ben Taieb, Gianluca Bontempi, Amir~F Atiya, and Antti Sorjamaa.
\newblock A review and comparison of strategies for multi-step ahead time
  series forecasting based on the {NN5} forecasting competition.
\newblock {\em Expert systems with applications}, 39(8):7067--7083, 2012.

\bibitem{hyndman2019tsfeatures}
Rob Hyndman, Yanfei Kang, Pablo Montero-Manso, Thiyanga Talagala, Earo Wang,
  Yangzhuoran Yang, and Mitchell O'Hara-Wild.
\newblock {\em tsfeatures: Time Series Feature Extraction}, 2020.
\newblock R package version 1.0.2.
  \url{https://pkg.robjhyndman.com/tsfeatures/}.

\bibitem{zellner1962efficient}
Arnold Zellner.
\newblock An efficient method of estimating seemingly unrelated regressions and
  tests for aggregation bias.
\newblock {\em Journal of the American Statistical Association},
  57(298):348--368, 1962.

\bibitem{duncan2001forecasting}
George~T Duncan, Wilpen~L Gorr, and Janusz Szczypula.
\newblock Forecasting analogous time series.
\newblock In J~Scott Armstrong, editor, {\em Principles of Forecasting}, pages
  195--213. Springer, 2001.

\bibitem{trapero2015identification}
Juan~R Trapero, Nikolaos Kourentzes, and Robert Fildes.
\newblock On the identification of sales forecasting models in the presence of
  promotions.
\newblock {\em Journal of the Operational Research Society}, 66(2):299--307,
  2015.

\bibitem{bandara2020cluster}
Kasun Bandara, Christoph Bergmeir, and Slawek Smyl.
\newblock Forecasting across time series databases using recurrent neural
  networks on groups of similar series: A clustering approach.
\newblock {\em Expert Systems with Applications}, 140:112896, 2020.

\bibitem{oord2016wavenet}
Aaron van~den Oord, Sander Dieleman, Heiga Zen, Karen Simonyan, Oriol Vinyals,
  Alex Graves, Nal Kalchbrenner, Andrew Senior, and Koray Kavukcuoglu.
\newblock Wavenet: A generative model for raw audio.
\newblock {\em arXiv preprint arXiv:1609.03499}, 2016.

\bibitem{thung2018brief}
Kim-Han Thung and Chong-Yaw Wee.
\newblock A brief review on multi-task learning.
\newblock {\em Multimedia Tools and Applications}, 77(22):29705--29725, 2018.

\bibitem{mafildes2019customer}
Shaohui Ma and Robert Fildes.
\newblock Forecasting third-party mobile payments with implications for
  customer flow prediction.
\newblock {\em International Journal of Forecasting}, 36(3):739--760, 2020.

\bibitem{landry2016probabilistic}
Mark Landry, Thomas~P Erlinger, David Patschke, and Craig Varrichio.
\newblock Probabilistic gradient boosting machines for gefcom2014 wind
  forecasting.
\newblock {\em International Journal of Forecasting}, 32(3):1061--1066, 2016.

\bibitem{M5kaggle}
M5 forecasting competition, 2017.
\newblock \url{https://www.kaggle.com/c/m5-forecasting-accuracy}.

\bibitem{trinh2018learning}
Trieu~H Trinh, Andrew~M Dai, Minh-Thang Luong, and Quoc~V Le.
\newblock Learning longer-term dependencies in {RNNs} with auxiliary losses.
\newblock {\em arXiv preprint arXiv:1803.00144}, 2018.

\bibitem{borovykh2018dilated}
Anastasia Borovykh, Sander Bohte, and Cornelis~W Oosterlee.
\newblock Dilated convolutional neural networks for time series forecasting.
\newblock {\em Journal of Computational Finance}, 22(4):73--101, 2018.

\bibitem{hewamalage2019recurrent}
Hansika Hewamalage, Christoph Bergmeir, and Kasun Bandara.
\newblock Recurrent neural networks for time series forecasting: Current status
  and future directions.
\newblock {\em International Journal of Forecasting}, 37(1):388 -- 427, 2021.

\bibitem{bandara2019lstm}
Kasun Bandara, Christoph Bergmeir, and Hansika Hewamalage.
\newblock Lstm-msnet: Leveraging forecasts on sets of related time series with
  multiple seasonal patterns.
\newblock {\em arXiv preprint arXiv:1909.04293}, 2019.

\bibitem{sen2019think}
Rajat Sen, Hsiang-Fu Yu, and Inderjit~S Dhillon.
\newblock Think globally, act locally: A deep neural network approach to
  high-dimensional time series forecasting.
\newblock In {\em Advances in Neural Information Processing Systems}, pages
  4837--4846, 2019.

\bibitem{wang2019deep}
Yuyang Wang, Alex Smola, Danielle~C Maddix, Jan Gasthaus, Dean Foster, and Tim
  Januschowski.
\newblock Deep factors for forecasting.
\newblock {\em arXiv preprint arXiv:1905.12417}, 2019.

\bibitem{makridakis2018statistical}
Spyros Makridakis, Evangelos Spiliotis, and Vassilios Assimakopoulos.
\newblock Statistical and machine learning forecasting methods: Concerns and
  ways forward.
\newblock {\em PLOS One}, 13(3), 2018.

\bibitem{tscv}
Christoph Bergmeir, Rob~J Hyndman, and Bonsoo Koo.
\newblock A note on the validity of cross-validation for evaluating
  autoregressive time series prediction.
\newblock {\em Computational Statistics \& Data Analysis}, 120:70--83, 2018.

\end{thebibliography}

\clearpage

\appendix
\section{Appendix: Extra results}

\begin{figure}[!hb]
  \centering
  \includegraphics[width=0.8\textwidth]{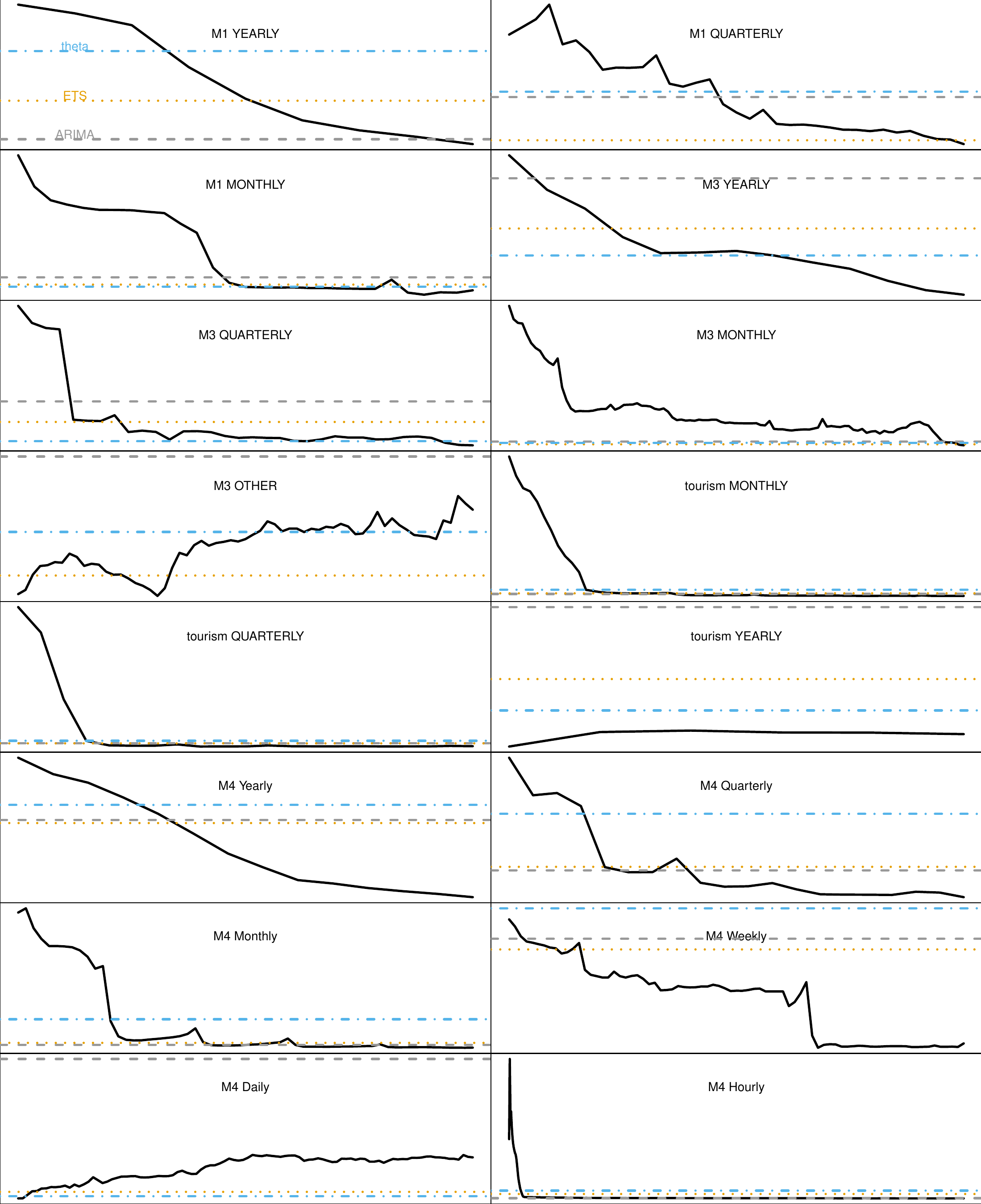}
  \caption{MASE error of a global linear AR model. When the memory is allowed to increase, the global model outperforms local alternatives for all frequencies in the M1, M3, tourism and M4 datasets. The results extends to most of the additional datasets. In this experiment, the M4 dataset was subsampled to 2000 series per frequency. Differences with previous Figures of the same datasets are due to the fact that we consider longer memories by subsetting each dataset to the series longer that each given memory level in order to fit longer memories.}
  \label{fig:longmemextra}
\end{figure}

\begin{figure}
\centering
\subfloat[]{
  \includegraphics[width=87mm]{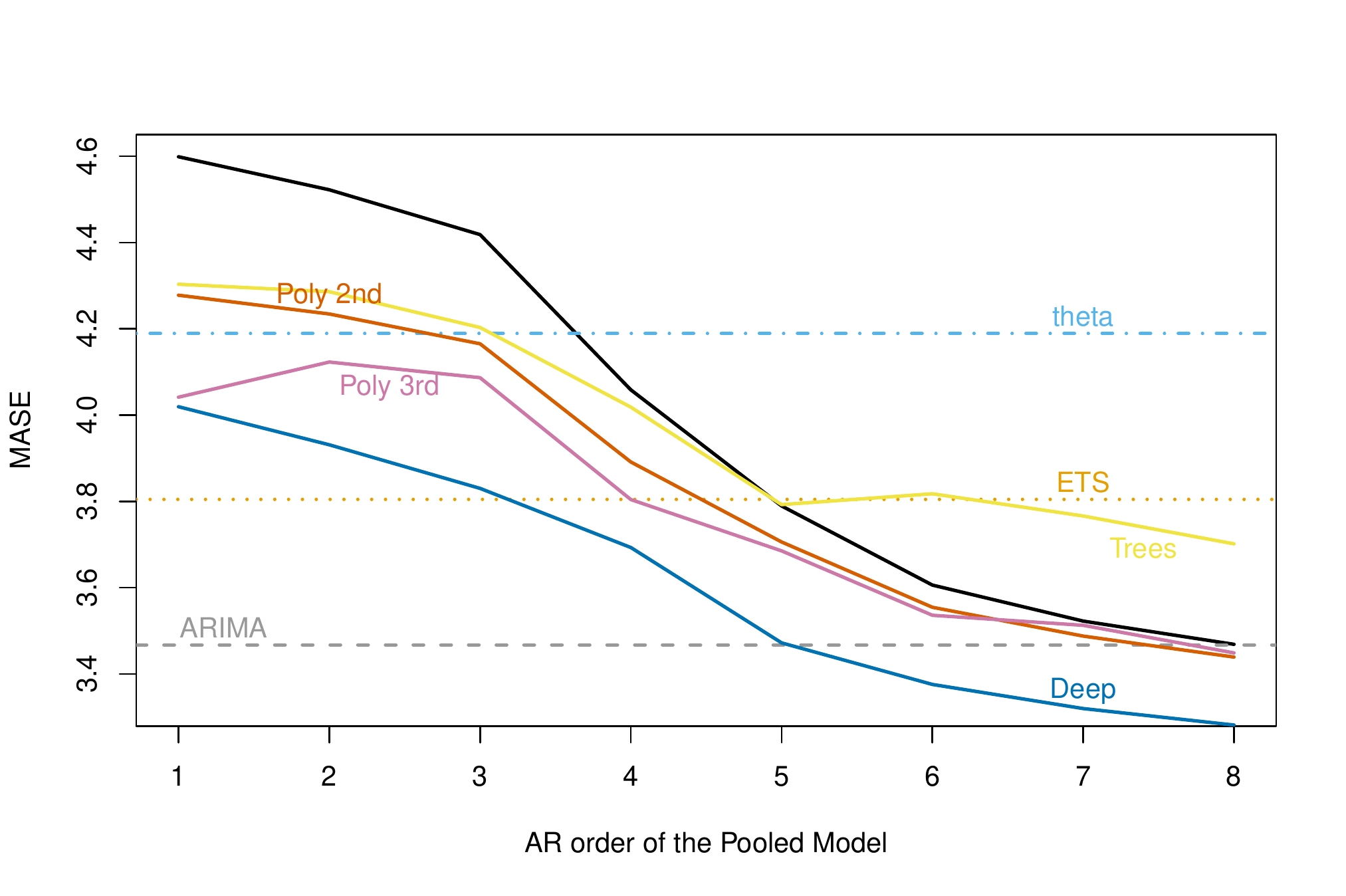}
  }
\subfloat[]{
  \includegraphics[width=87mm]{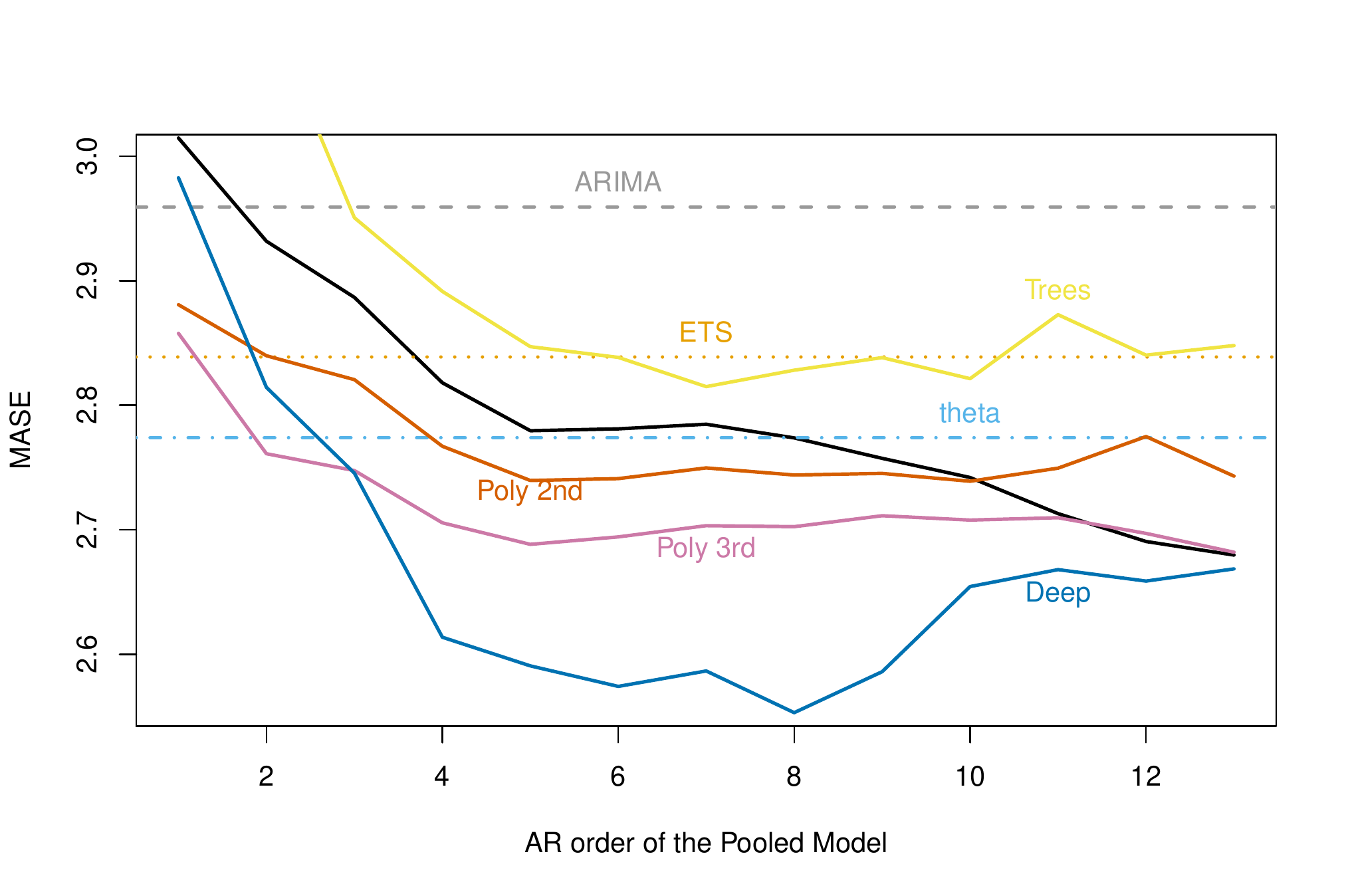}
  }
\hspace{0mm}
\subfloat[]{
  \includegraphics[width=87mm]{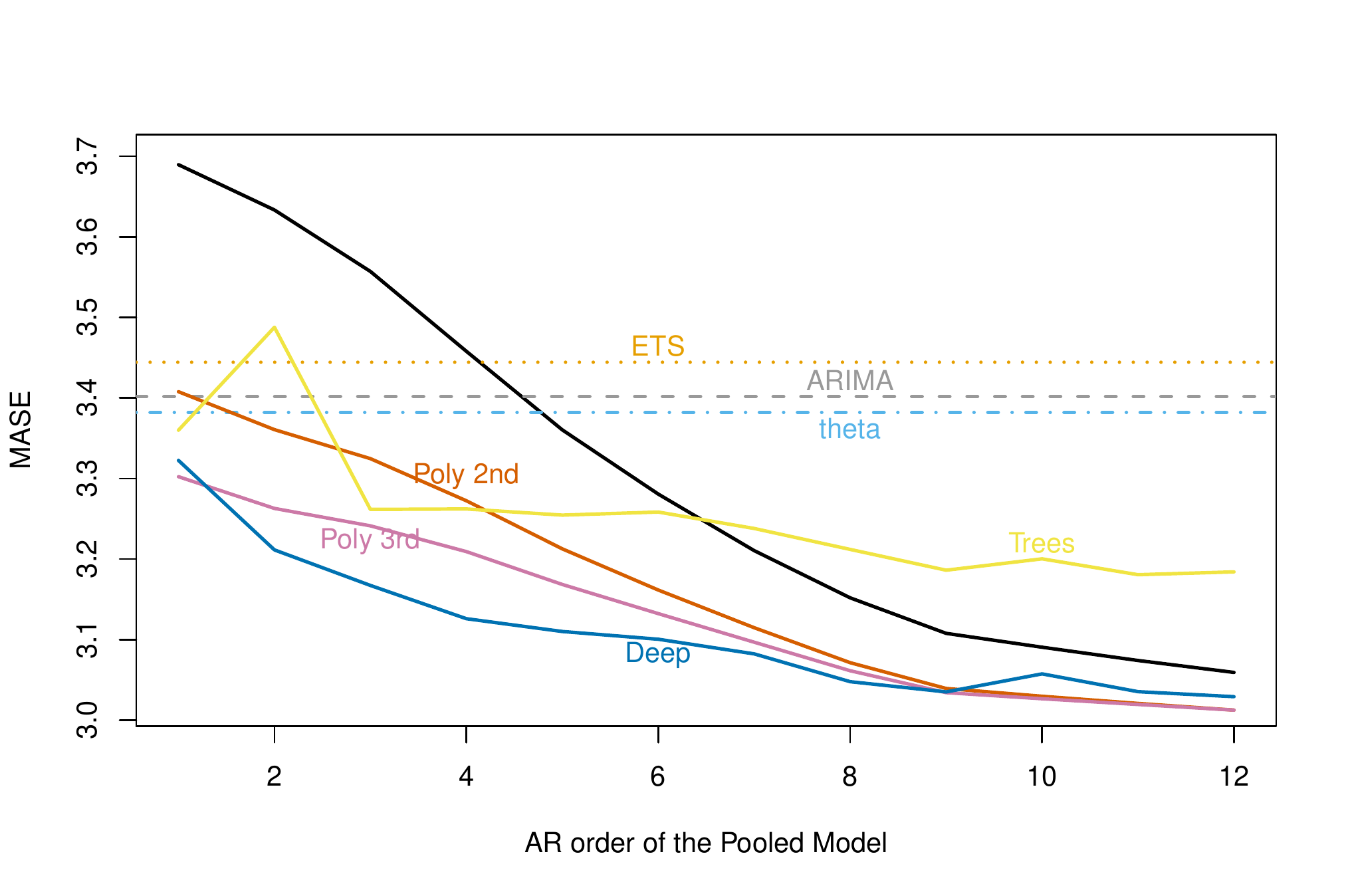}
}
\subfloat[]{
  \includegraphics[width=87mm]{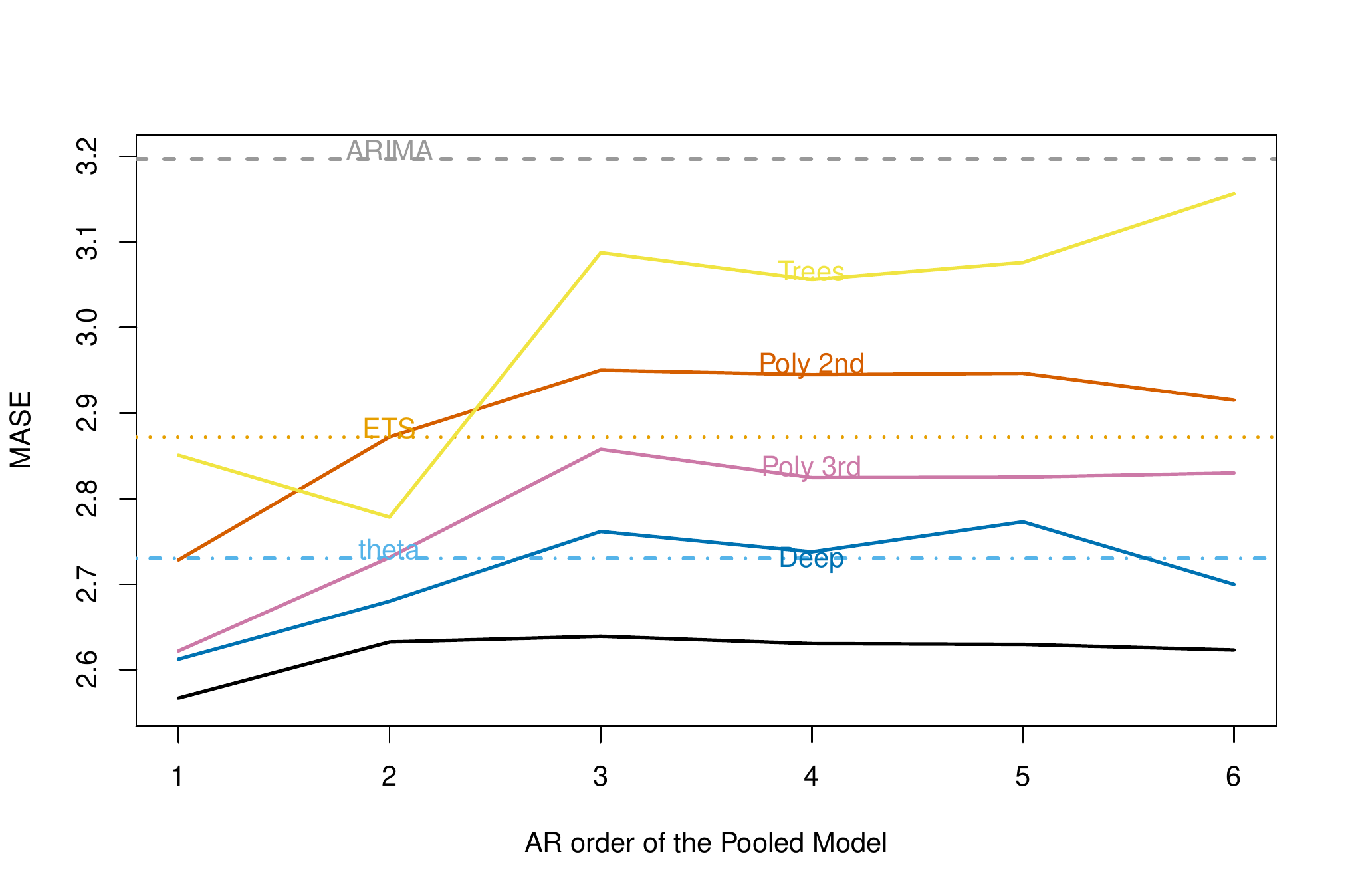}
}
\caption{MASE error for Yearly series for the M1 (a), M3 (b), M4 (c) and Tourism (d) datasets. Global models outperform the local state-of-the-art alternatives. The increase of complexity of the global nonlinear models over the global results in a significant improvement in accuracy for most datasets.}
\label{fig:totalyearly}
\end{figure}

\begin{figure}
\centering
\subfloat[]{   
  \includegraphics[width=87mm]{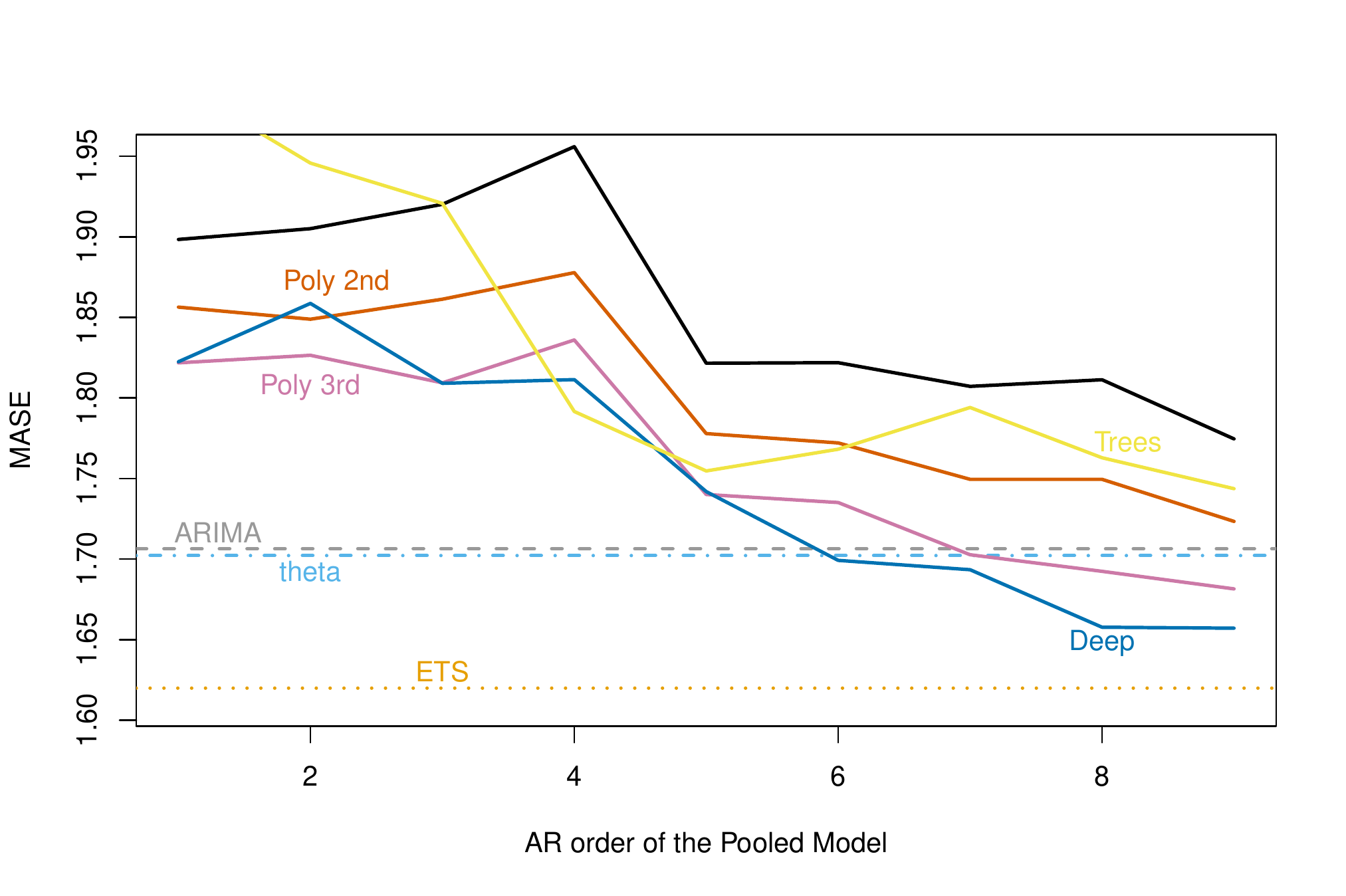}
}
\subfloat[]{
  \includegraphics[width=87mm]{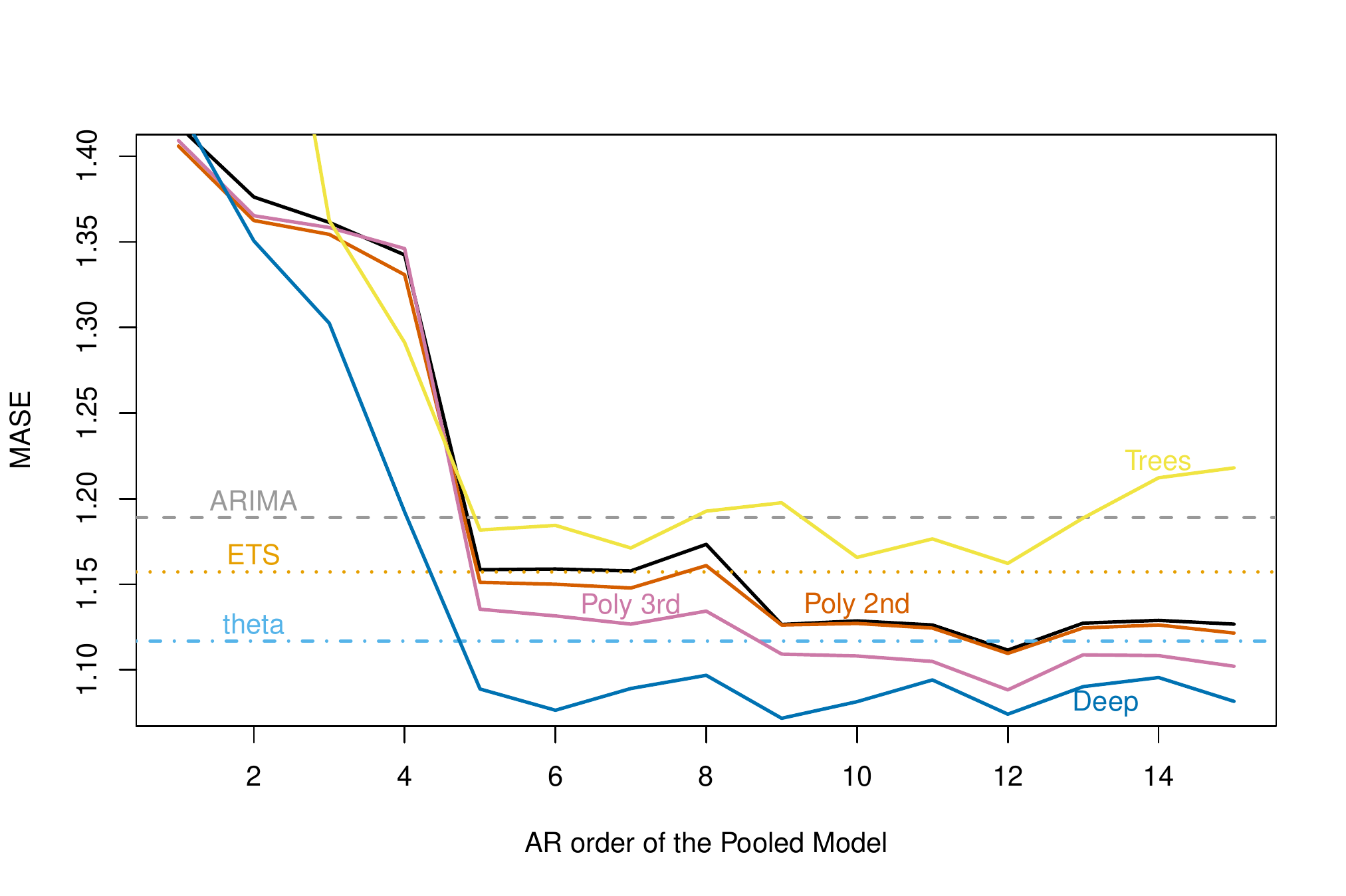}
}
\hspace{0mm}
\subfloat[]{   
  \includegraphics[width=87mm]{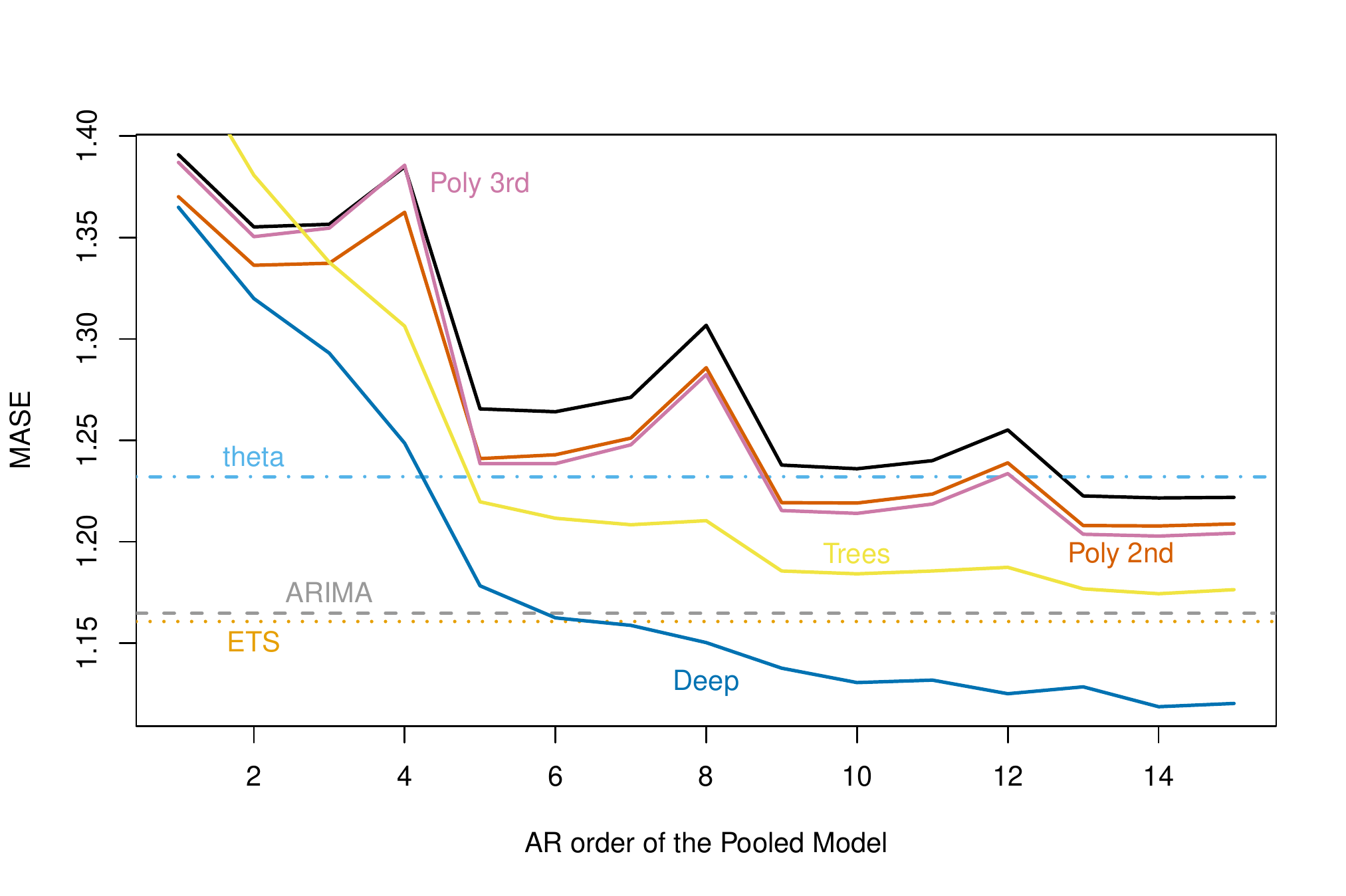}
}
\subfloat[]{
  \includegraphics[width=87mm]{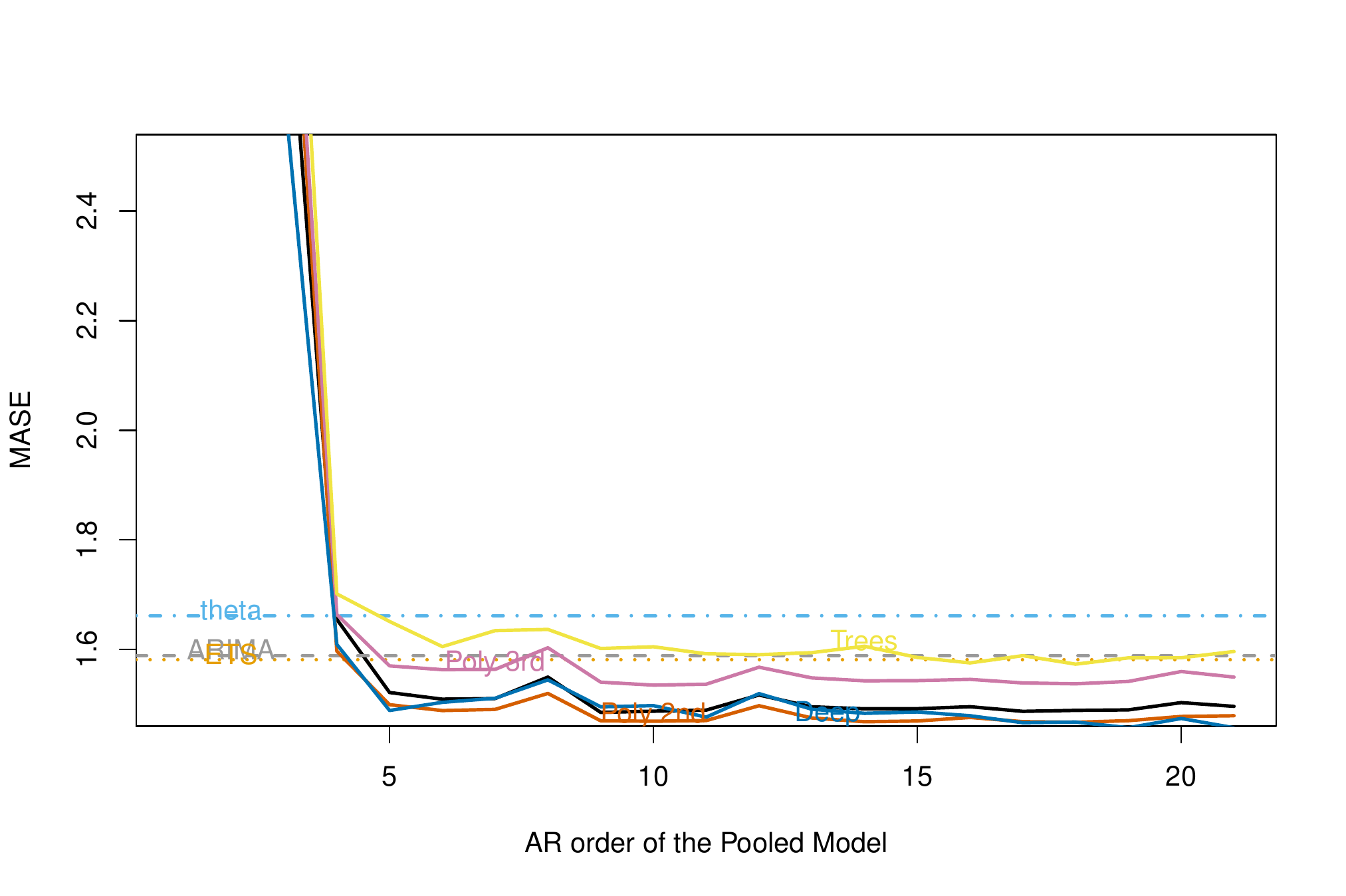}
}
\caption{MASE error for Quarterly series for the M1 (a), M3 (b), M4 (c) and Tourism (d) datasets. A global linear model (solid black line) compared to the local state-of-the-art alternatives (horizontal lines) and nonlinear models (colored lines). As memory / order of the autoregression increases, the forecasts accuracy of the global models improves. Notably, for the M4 Quarterly, 24000 time series in total, a global linear AR(14) model (only 14 numbers to forecast the whole dataset) is able to outperform a competitive local method (theta). Nonlinear global models improve over the global linear in terms of overall accuracy. The accuracy of nonlinear global models improves faster with memory, less memory is required to reach a good accuracy compared to linear.}
\label{fig:totalquart}
\end{figure}

\begin{figure}
\centering
\subfloat[]{
  \includegraphics[width=87mm]{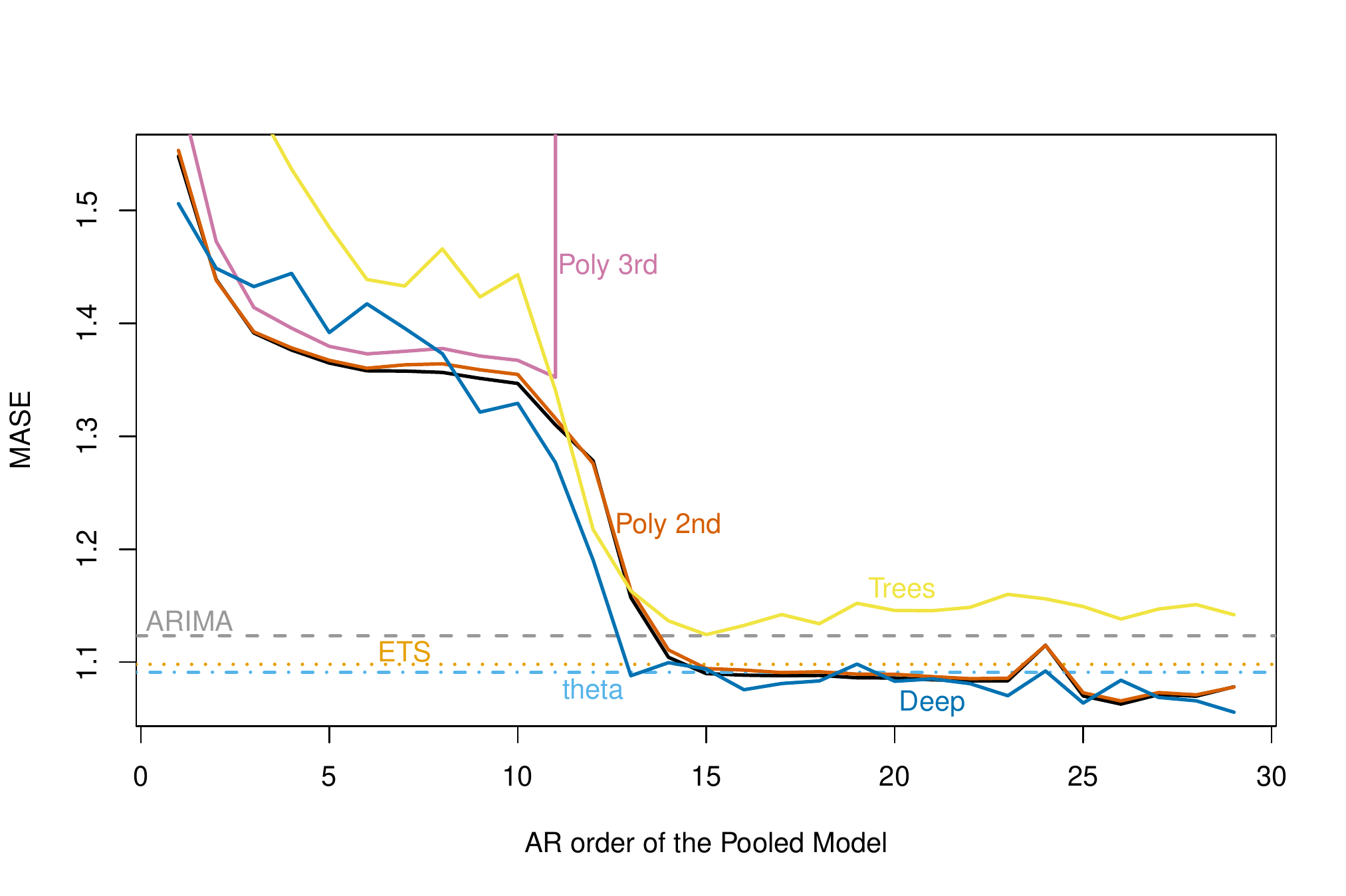}
}
\subfloat[]{
  \includegraphics[width=87mm]{./fig/MCompandTour/M3MonthNonlin.pdf}
  \label{fig:M3monthshort}
}

\hspace{0mm}
\subfloat[]{
  \includegraphics[width=87mm]{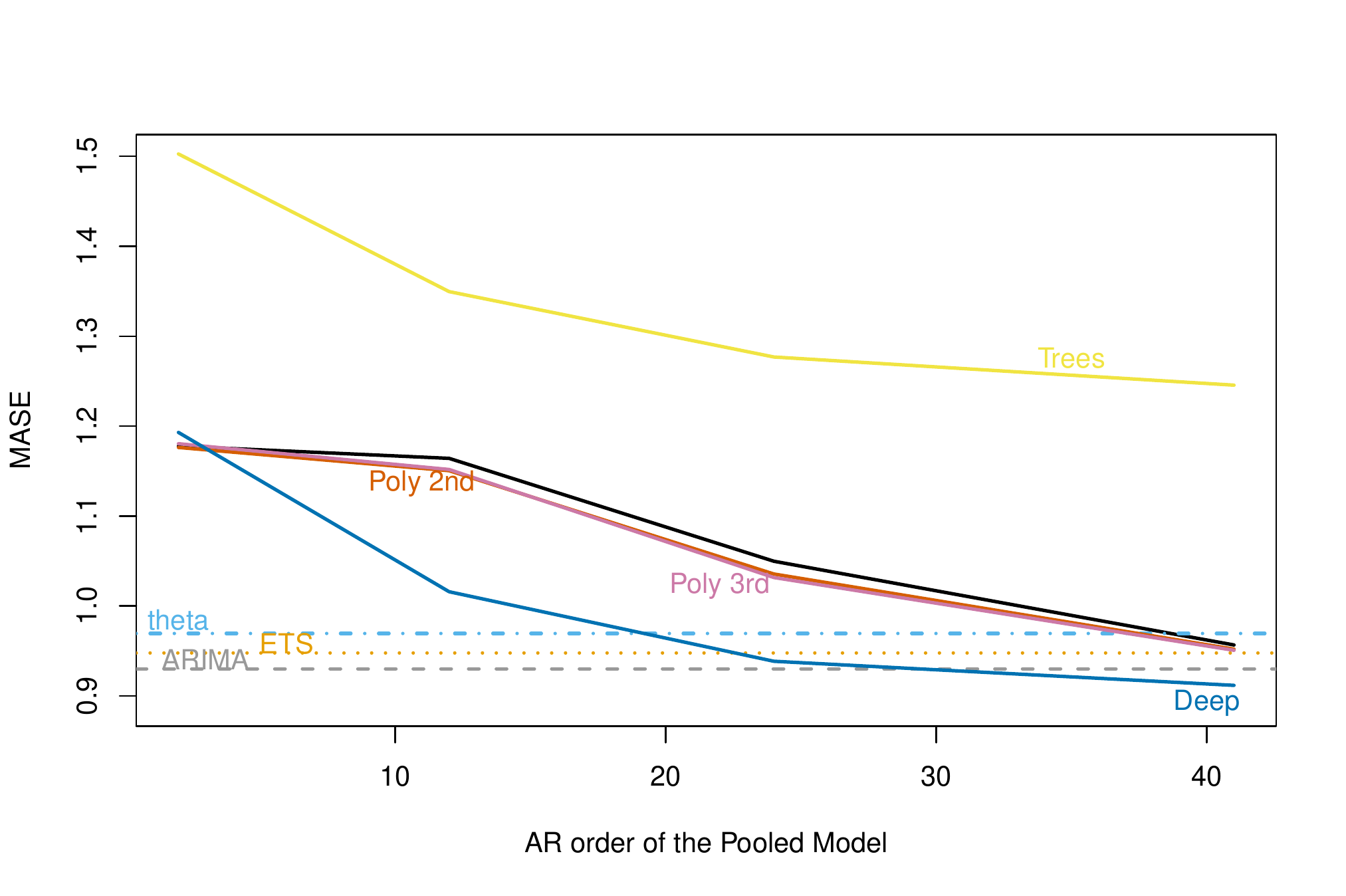}
}
\subfloat[]{
  \includegraphics[width=87mm]{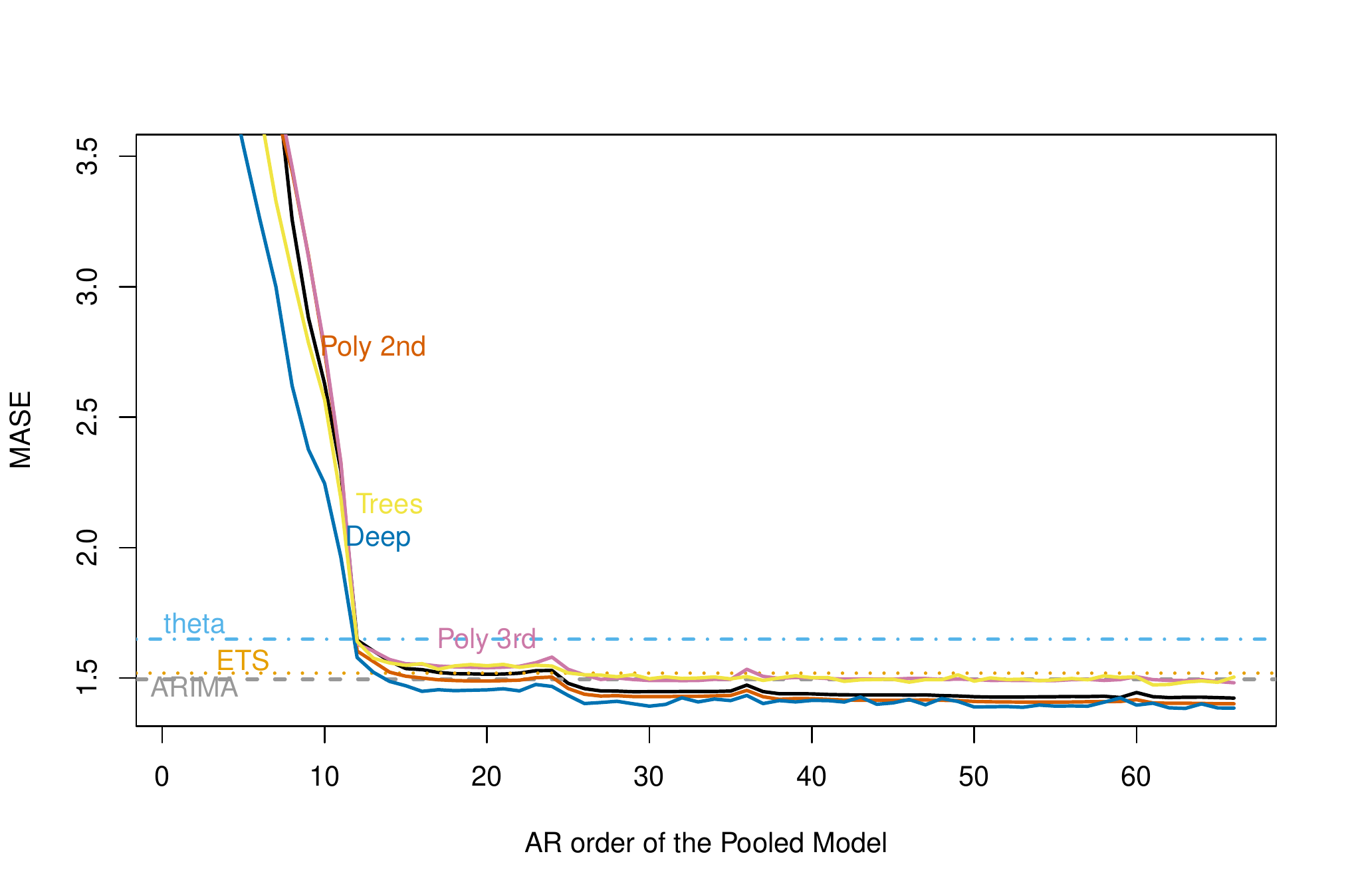}
}

\caption{MASE error for Monthly series for the M1 (a), M3 (b), M4 (c) and Tourism (d) datasets. A global linear model (solid black line) compared to the local state-of-the-art alternatives (horizontal lines) and nonlinear models (colored lines). The effect is similar to the Quarterly frequency case of Figure~\ref{fig:totalquart}, global linear models can reach comparative levels of performance than the local state-of-the-art alternatives with orders of magnitude fewer parameters. The global nonlinear models improve over the global linear in terms of accuracy and the memory required to achieve good results. The polynomial of 3rd degree model explodes in the M1 Monthly (a), due to numerical instability (a 18 step-ahead recursive forecast of a 3rd degree polynomial is a 54-degree polynomial) but it is stable for the other datasets, which have more data to fit the models.}
\label{fig:totalmonth}
\end{figure}

\begin{figure}
  \centering
  \includegraphics[width=0.75\textwidth]{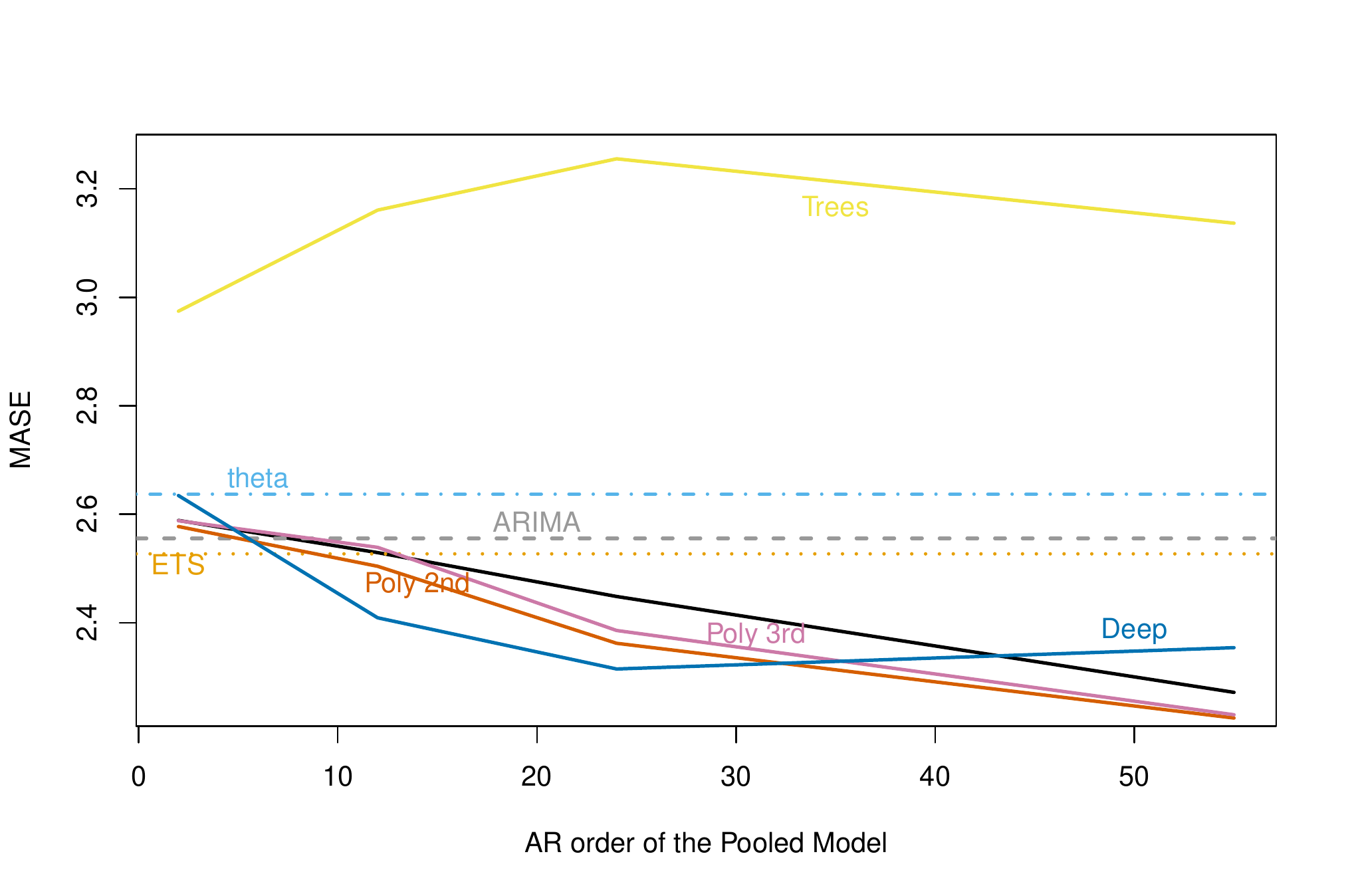}
  \caption{MASE for the M4 Weekly dataset.A global linear model (solid black line) compared to the local state-of-the-art alternatives (horizontal lines) and nonlinear models (colored lines). Due to computational requirements, lags considered are (2,12,24,55) instead of the full range (1 to 55). Global models quickly outperform local alternatives.}
  \label{fig:totalweek}
\end{figure}

%
%

\begin{figure}
  \centering
  \includegraphics[width=0.8\textwidth]{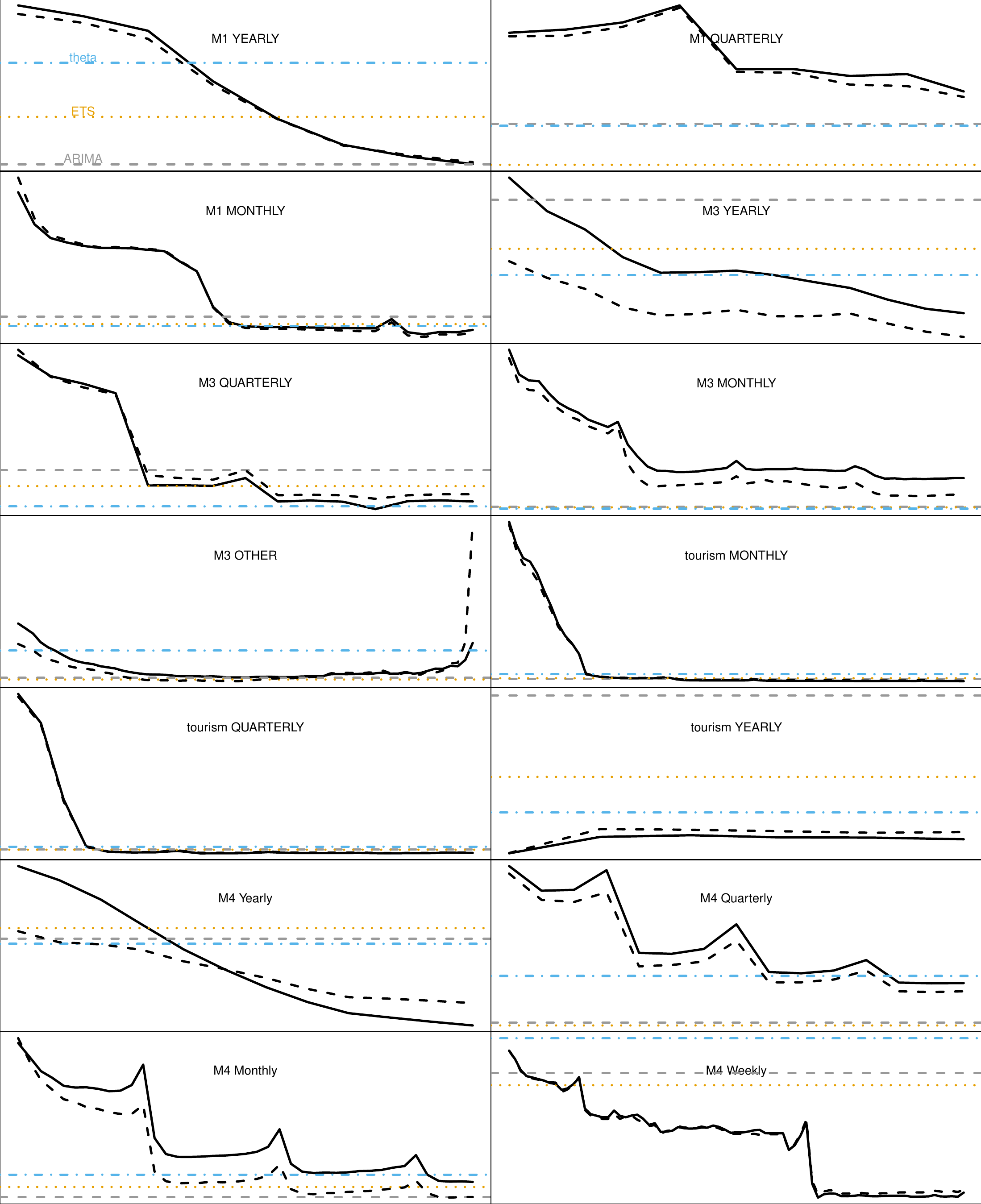}
  \caption{Partitioning experiment in the M1, M3, tourism and M4 datasets. Global linear model (solid black line) compared to the partitioned linear model (dashed black line) and the state-of-the-art local alternatives. The partitioned linear model randomly partitions each dataset into 10 disjoint subsets and fits a linear model to each subset (a ``global'' model within each the subset), increasing the complexity of the overall model compared to the global linear. In the majority of datasets, the Partitioned model outperforms the unpartitioned one, showing that the global linear is constrained by its simplicity (under-fitting).}
  \label{fig:partitioningextra}
\end{figure}

\begin{figure}
  \centering
  \includegraphics[width=0.75\textwidth]{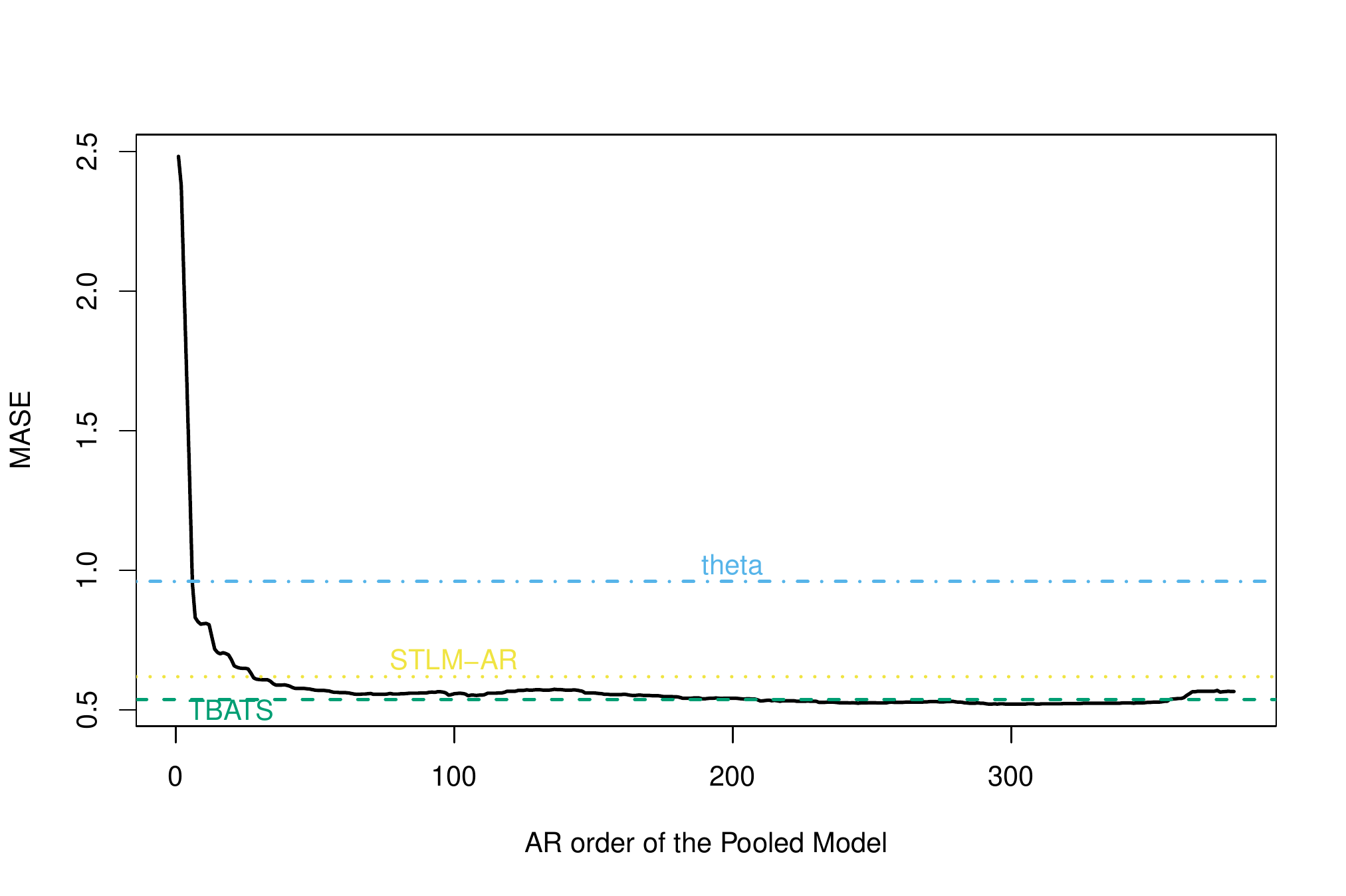}
  \caption{MASE for 111 daily time series, NN5 competition. Compared to tbats, stlm and theta (in order of accuracy). The data consists of 2 years of daily cash demand at various automatic teller machines (ATMs, or cash machines) at different locations in England. Global linear model shows better accuracy than local methods, and a clear pattern of improvement of accuracy the large the memory of the global model. To highlight the relevance/counter-intuitiveness of these results, refer to the quotation from the competition webpage:
    \textit{``The data may contain a number of time series patterns including multiple overlying seasonality, local trends, structural breaks, outliers, zero and missing values, etc\dots\ requiring an individual modeling approach for each time series.''}
}
  \label{fig:nn5}
\end{figure}

\begin{figure}
  \centering
  \includegraphics[width=0.75\textwidth]{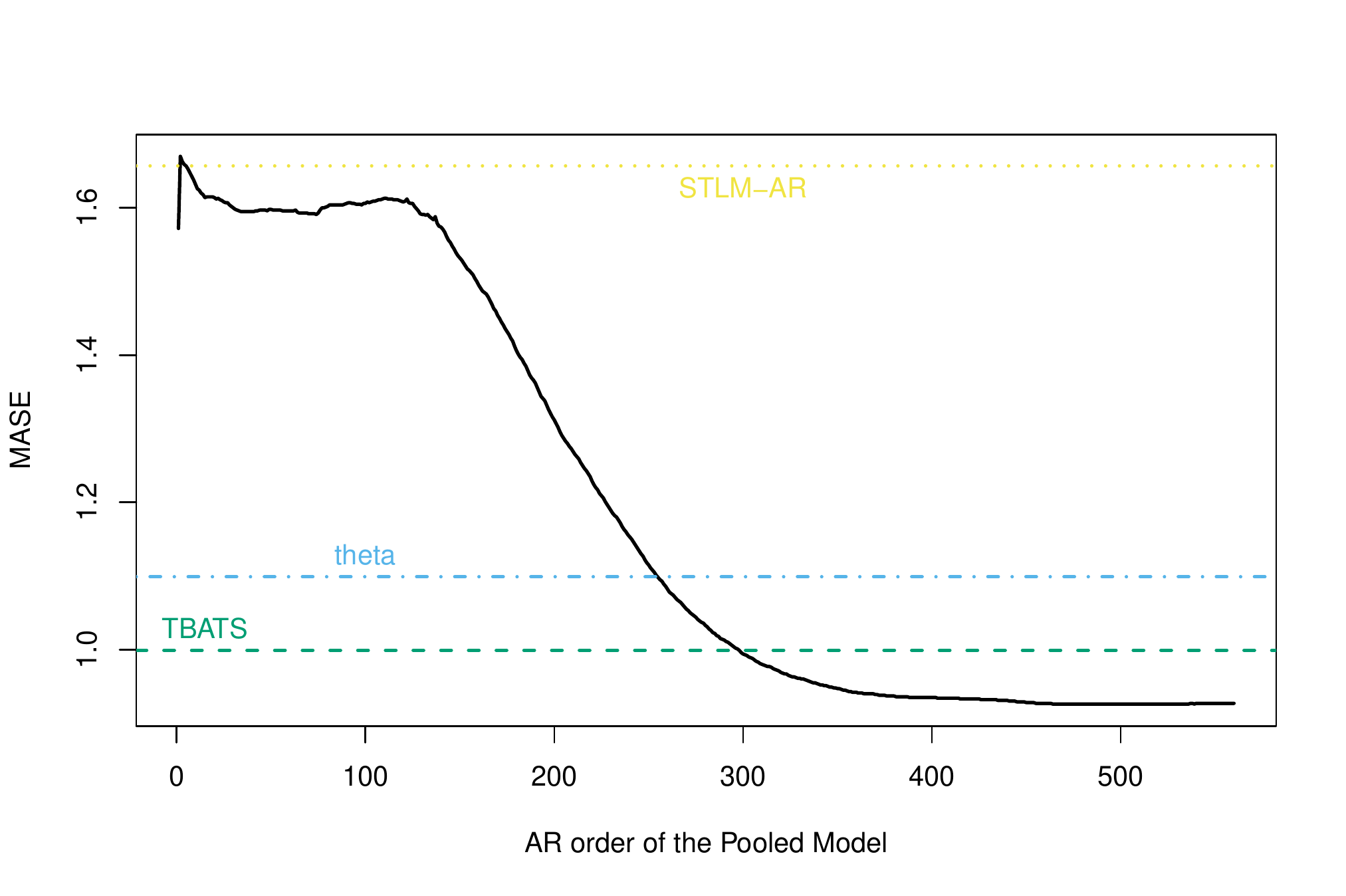}
  \caption{MASE of llobal linear model for Weather dataset, 3010 daily time series of average precipitation, temperature and solar radiation. Compared to tbats, theta and stlm in order of performance. There is a clear pattern of improvement in accuracy when the memory is increased, somehow ``saturating'' at memory levels corresponding to one year cycle (365).}
  \label{fig:weather}
\end{figure}

\begin{figure}
  \centering
  \includegraphics[width=0.75\textwidth]{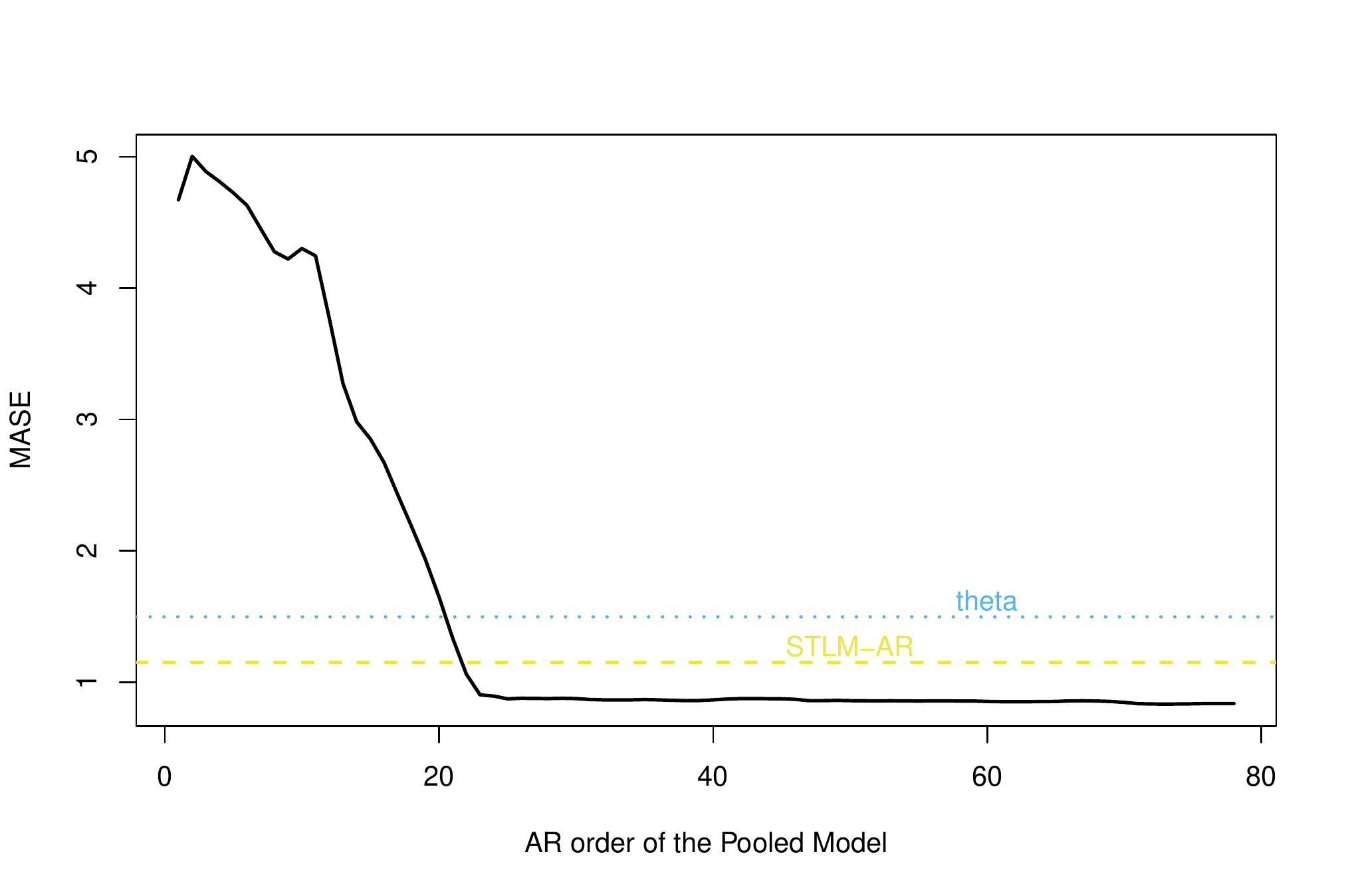}
  \caption{MASE for the traffic dataset, 963 hourly time series, each of 10392 observations, of occupancy rate in lanes of highway. Series clipped at the 600 last observations for performance purposes. Stlm and theta used as benchmark, other methods do not yield results in reasonable time. A global linear model outperforms local alternatives, saturating at an AR order of 24, one day cycle. }
  \label{fig:traffic}
\end{figure}

\begin{figure}
  \centering
  \includegraphics[width=0.75\textwidth]{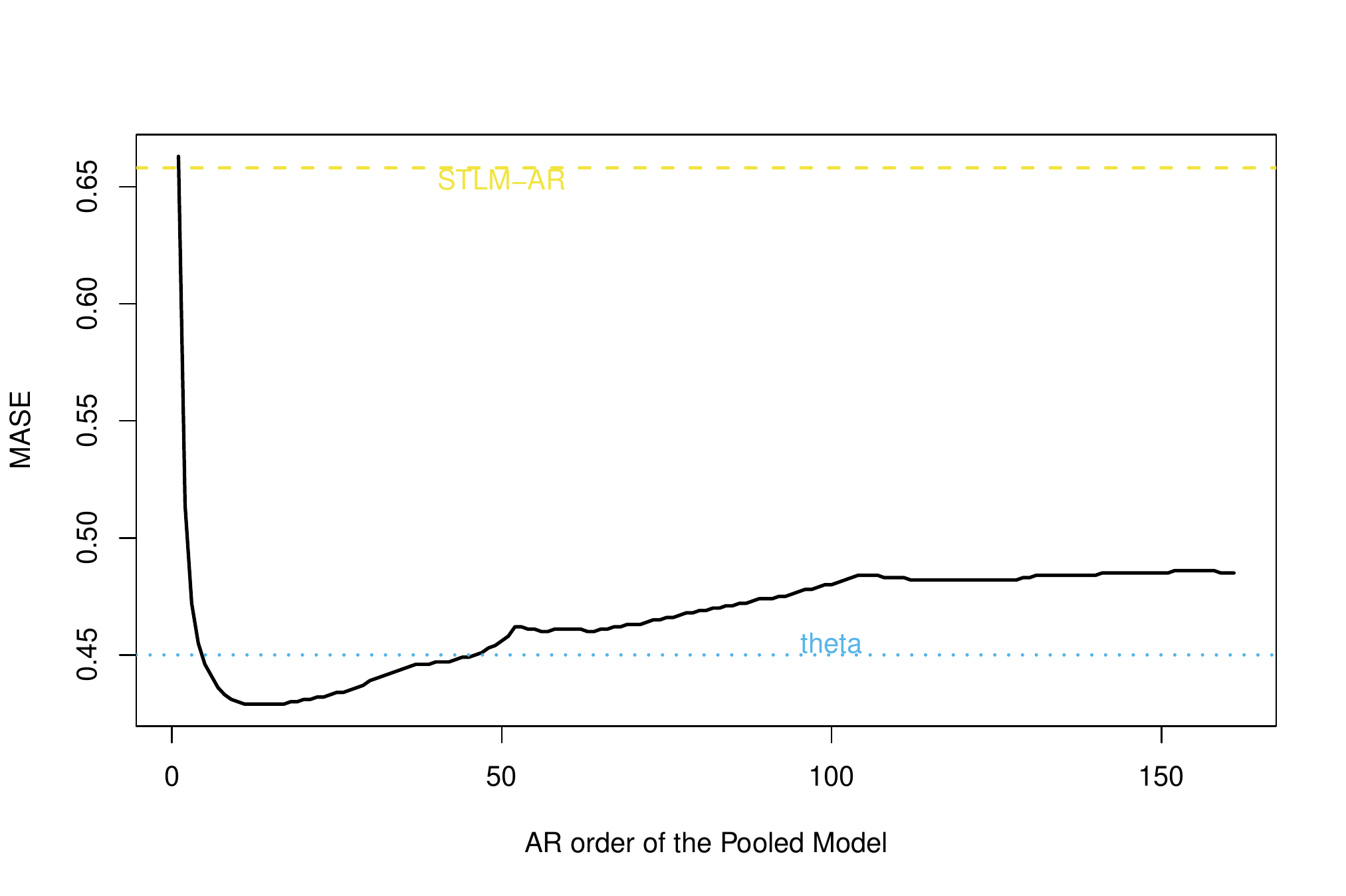}
  \caption{MASE for the Dominick dataset, 100K weekly time series of profits. A subset of the whole dataset for performance reasons.
  Each series is the profit of one product offered by a retailer.
  Global Linear AR compared to theta and stlm in order of better performance.
  The effect of increasing memory has a negative impact in performance accuracy, opposite to the vast majority of datasets.}
  \label{fig:dominick}
\end{figure}

\begin{figure}
  \centering
  \includegraphics[width=0.75\textwidth]{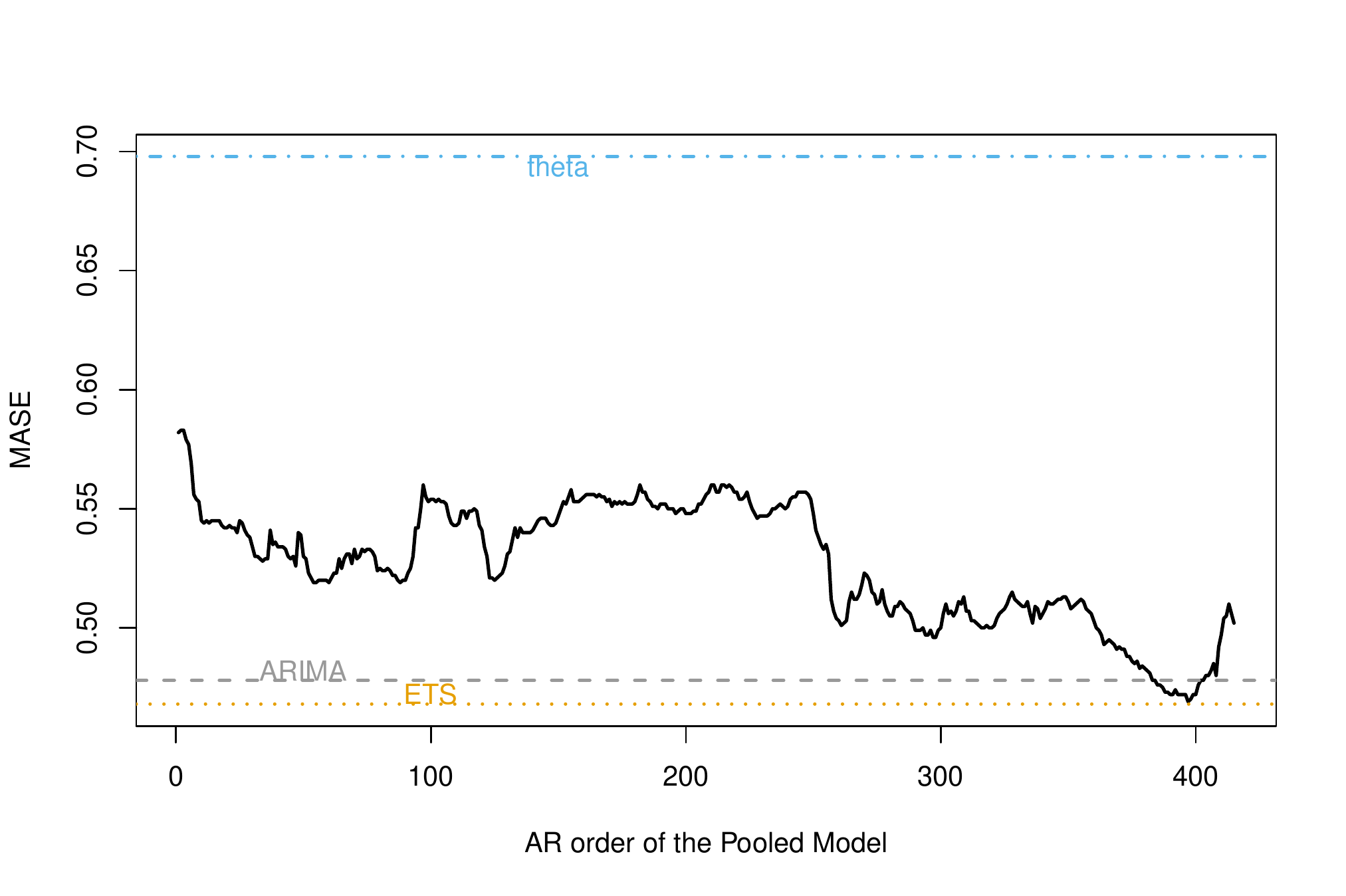}
  \caption{MASE for the FREDMD, a monthly dataset, 109 monthly time series of macroeconomic indicators. This datasets shows a more erratic pattern of accuracy with respect to the memory of the global model, but it is a positive correlation. The memory levels in this datasets are large, accuracy of the global model peaking at close to an AR(400), this model consider over 30 years of past observations.}
  \label{fig:FREDMD}
\end{figure}

\begin{figure}
  \centering
  \includegraphics[width=0.75\textwidth]{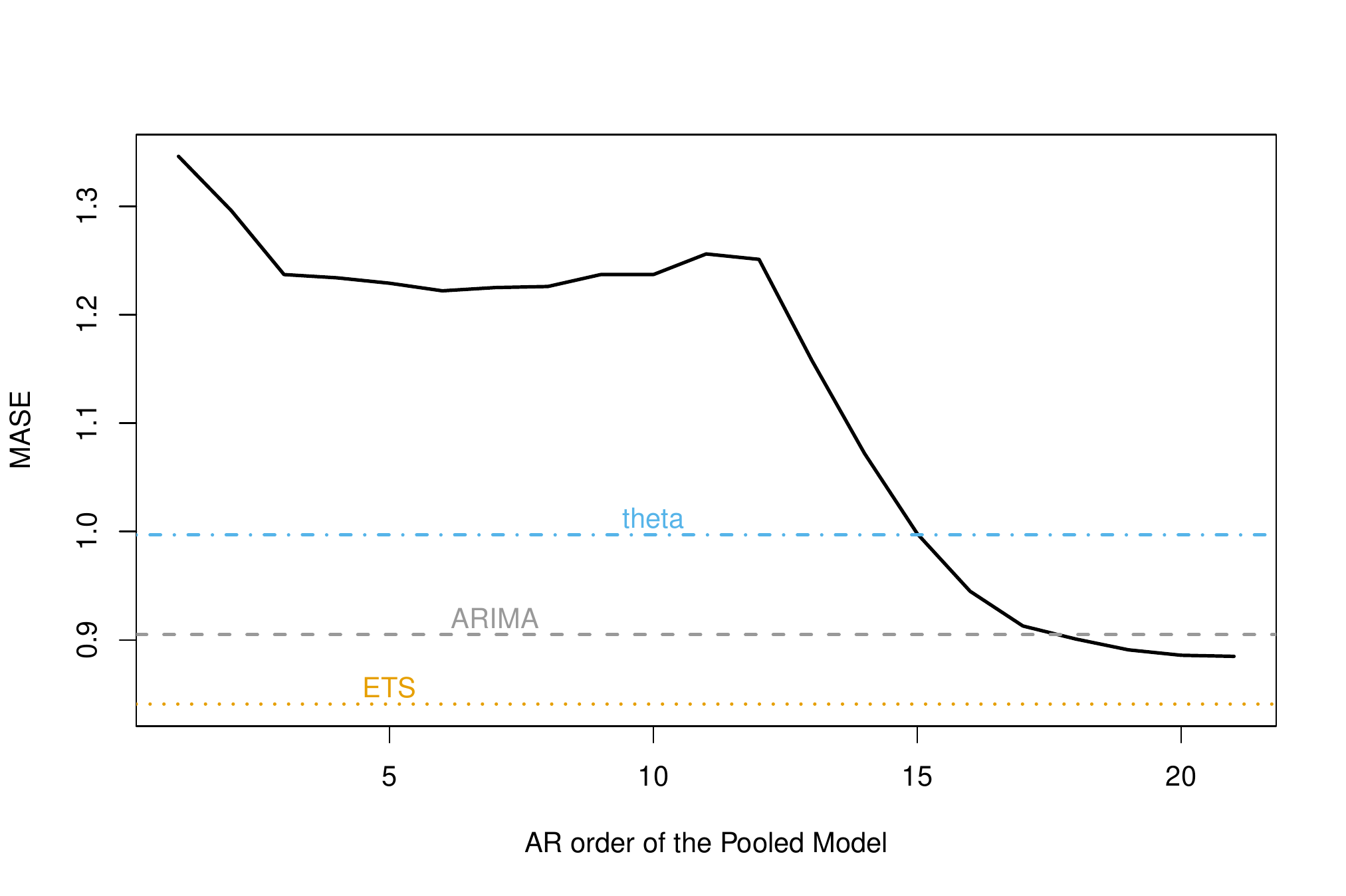}
  \caption{MASE for the CIF2016, a monthly dataset, 72 monthly time series from banking and simulations. Global linear AR compared to ets, auto.arima, and theta in order of better performance. The pattern of increasing accuracy the longer the memory also appears in this dataset, experience a big improvement when the level of memory passes the yearly cycle (12). }
  \label{fig:CIF2016}
\end{figure}

\begin{figure}
  \centering
  \includegraphics[width=0.75\textwidth]{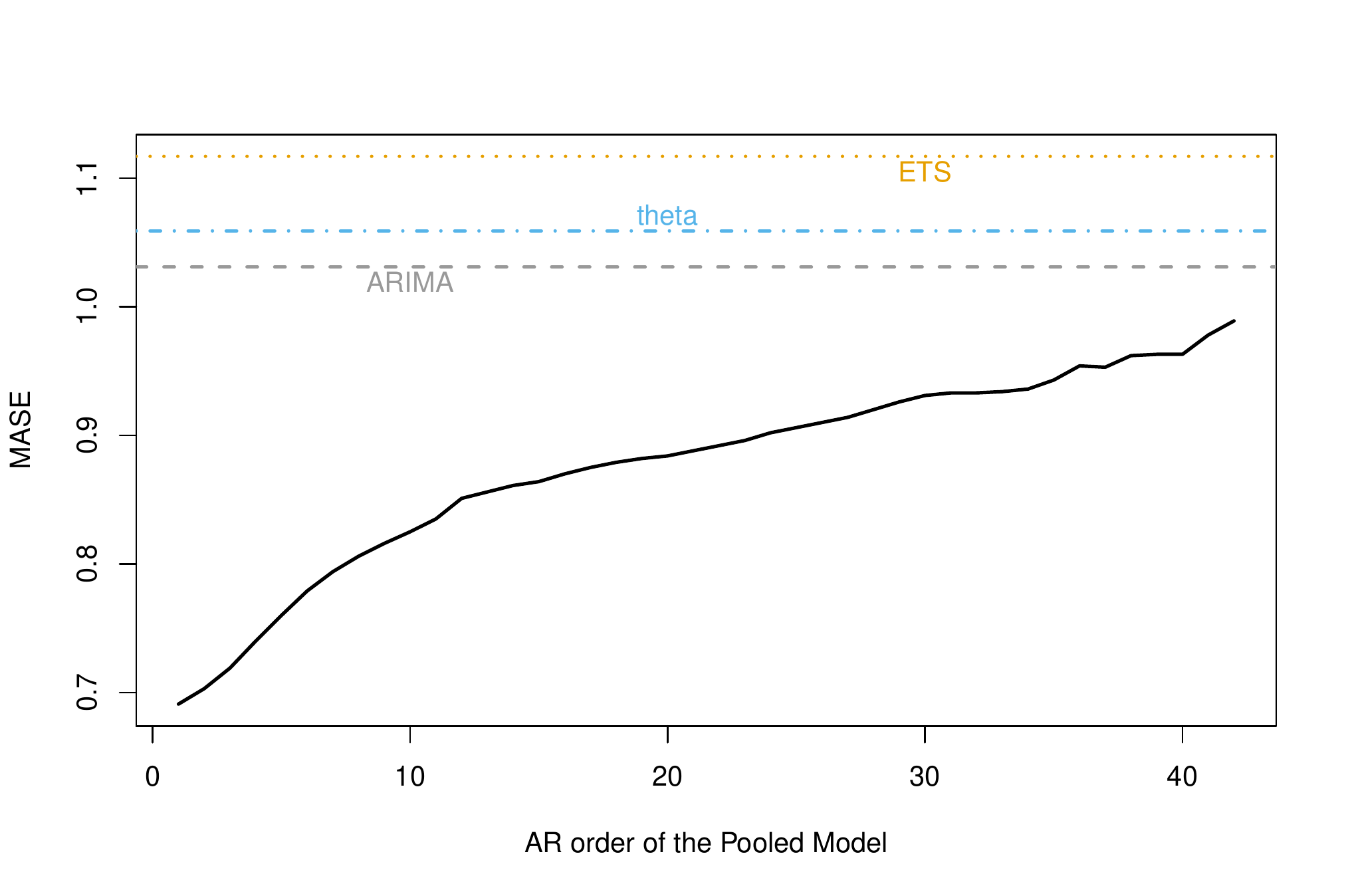}
  \caption{MASE for the parts monthly dataset, 2500 monthly time series of sales of car parts. Global linear AR compared to auto.arima, theta and ets in order of better performance. As in the Dominick's dataset, increasing memory decreases performance of the global model.}
  \label{fig:parts}
\end{figure}

\begin{figure}
  \centering
  \includegraphics[width=0.75\textwidth]{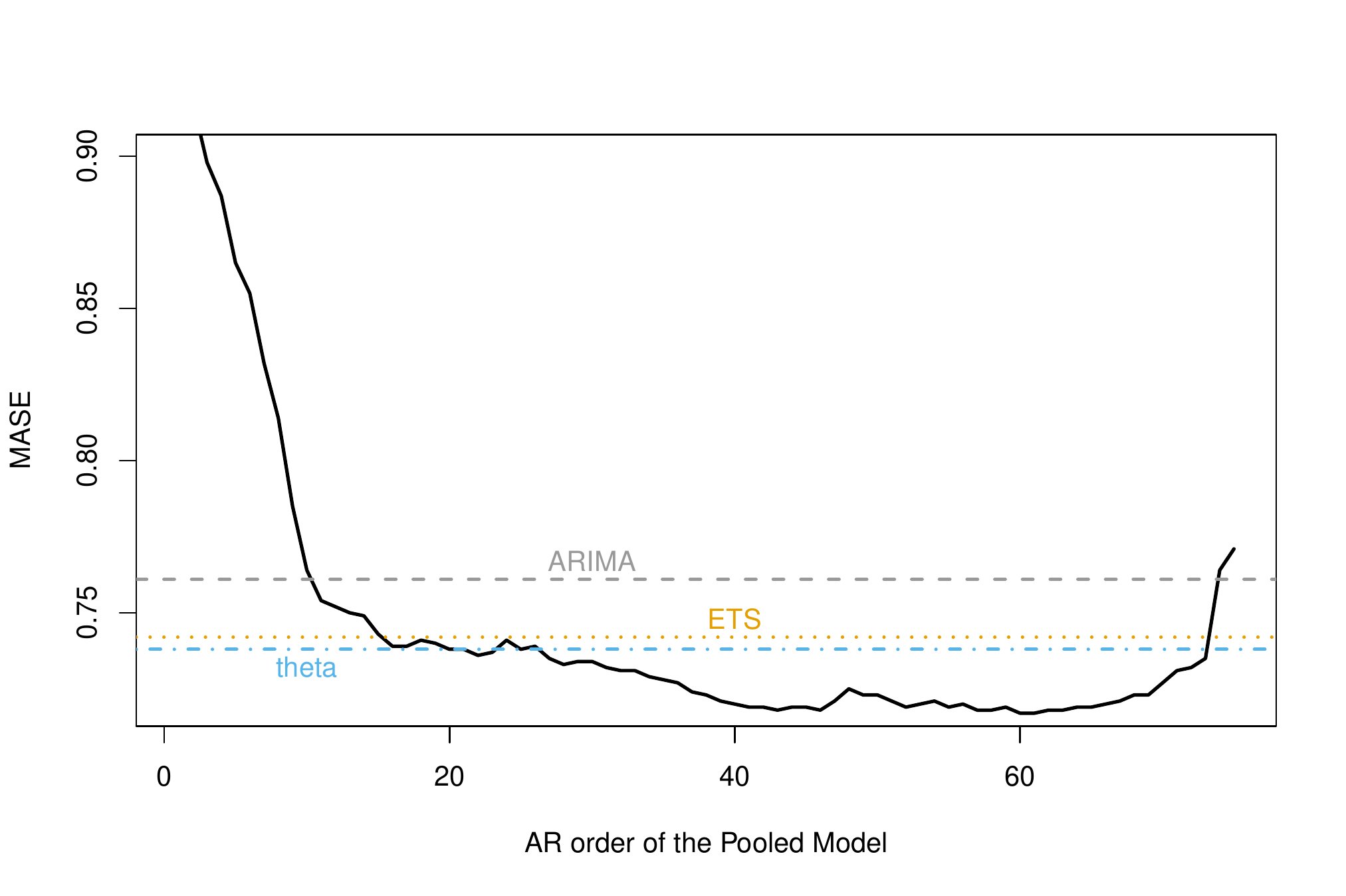}
  \caption{MASE for the hospital, a monthly dataset, 767 monthly time series of patient count for products related to medical problems. Global linear AR compared to theta, ets and auto.arima in order of better performance.}
  \label{fig:hospital}
\end{figure}

\begin{figure}
  \centering
  \includegraphics[width=0.75\textwidth]{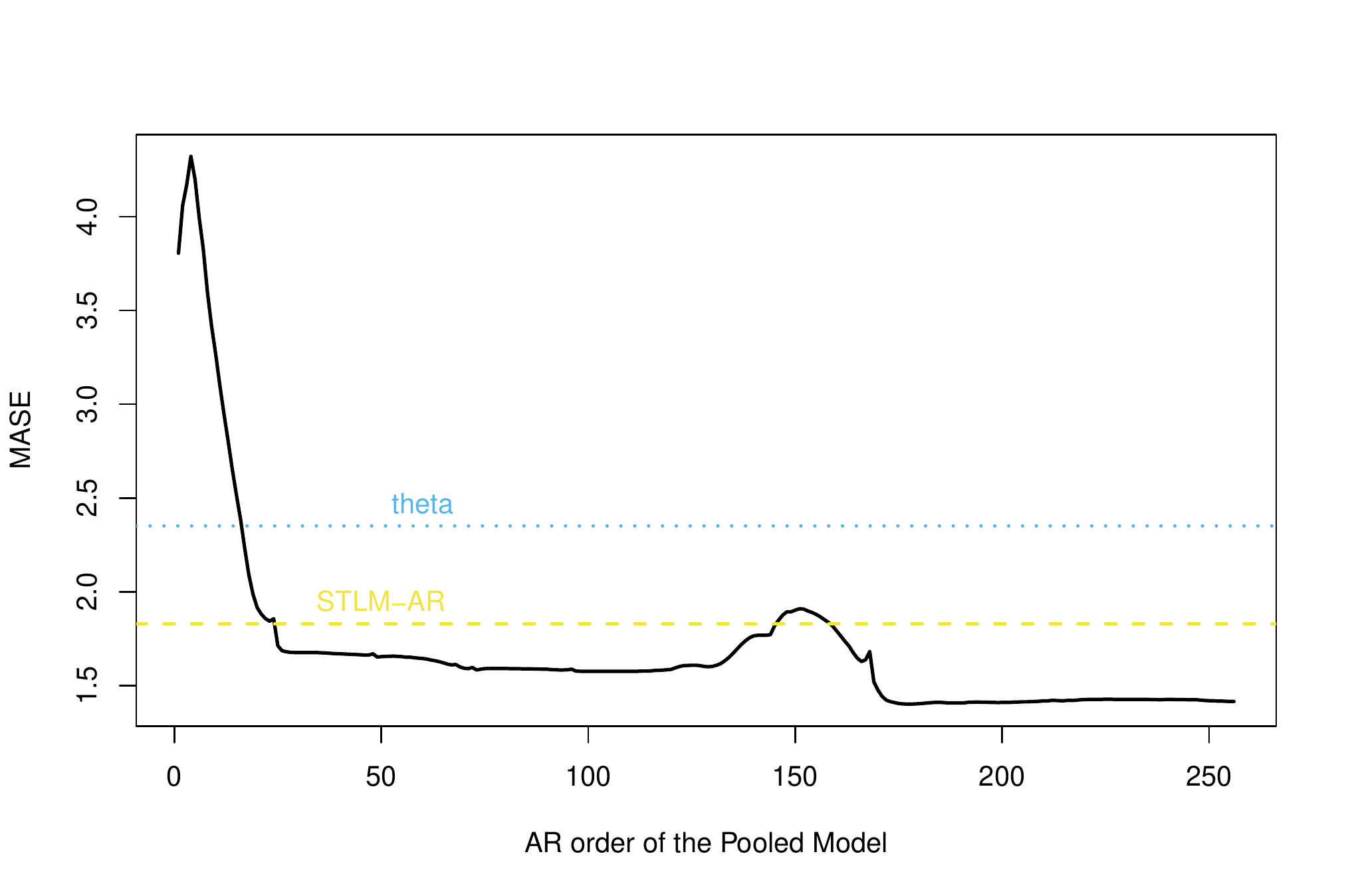}
  \caption{MASE for the electricity dataset. 370 series of hourly electricity demand, 26280 observations each.
  global linear AR compared to stlm and theta in order of performance. Accuracy of the global model increases quickly at  the AR(24) level (daily cycle) and experiences another big improvements at a memory level close to 168 (a weekly cycle).}
  \label{fig:electricity}
\end{figure}

\begin{figure}
  \centering
  \includegraphics[width=0.75\textwidth]{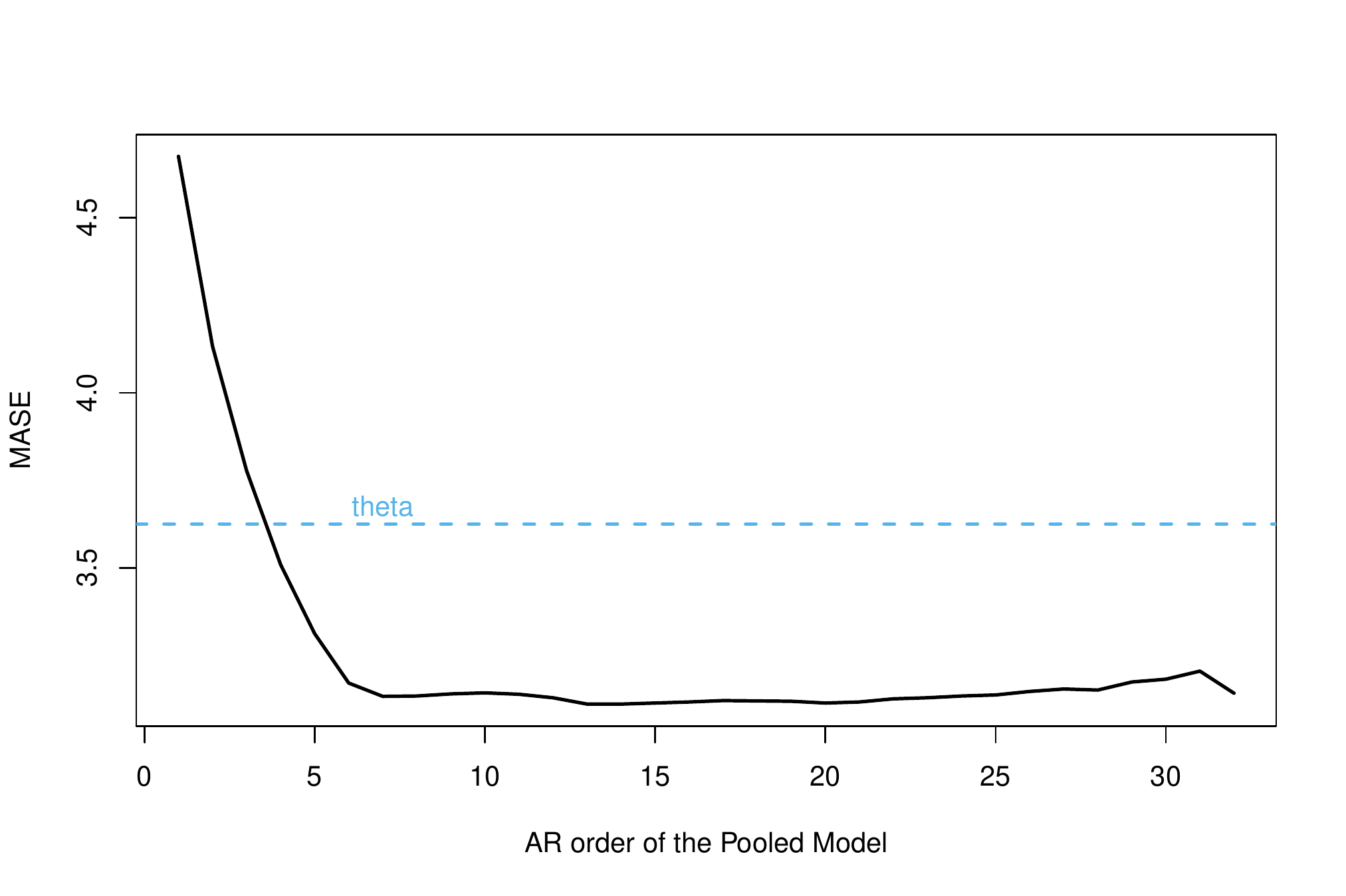}
  \caption{MASE for the wiki dataset. 100000 time series of daily visits. Each series represents on page within the Wikipedia domain. Compared to theta method. (ets did not produce realistic results and arima-based are too slow.).
  Series length is capped at 366 days, longer memory levels are not explored for performance reasons.}
  \label{fig:wiki}
\end{figure}

\begin{figure}
  \centering
  \includegraphics[width=0.75\textwidth]{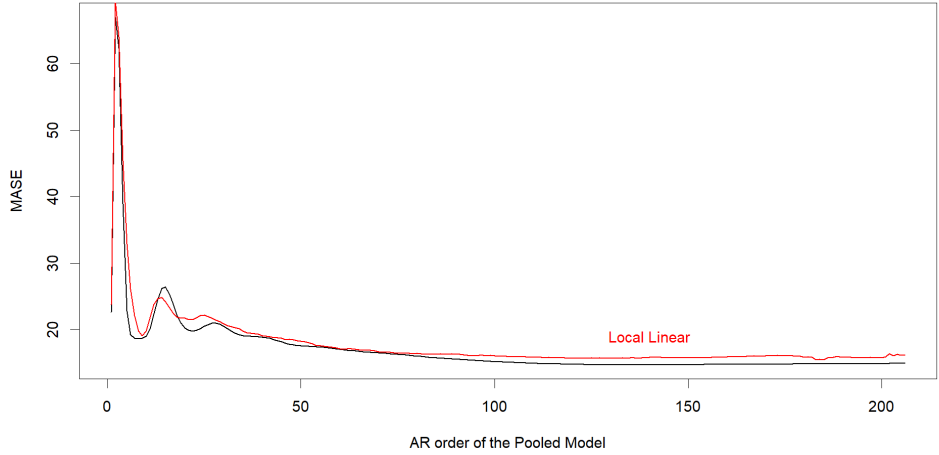}
  \caption{MASE for the double pendulum experiment. An example of a totally homogeneous dataset, all series come from repeating the double pendulum experiment, very long chaotic time series with little noise. Each series in the dataset comes from the same experiment, measured at different times / initial conditions.
  The local method (red) is a pure linear AR model of the same order as the global (black). Both local and global belong to the same model class, the only parameter that changes in this experiment is locality/globality. At the majority of memory levels, global outperforms local, while we see a clear pattern of increasing accuracy with longer memory in both.
  State-of-the-art local models did not produce accurate results in this experiment.}
  \label{fig:pendulum}
\end{figure}

\begin{figure}
  \centering
  \includegraphics[width=0.75\textwidth]{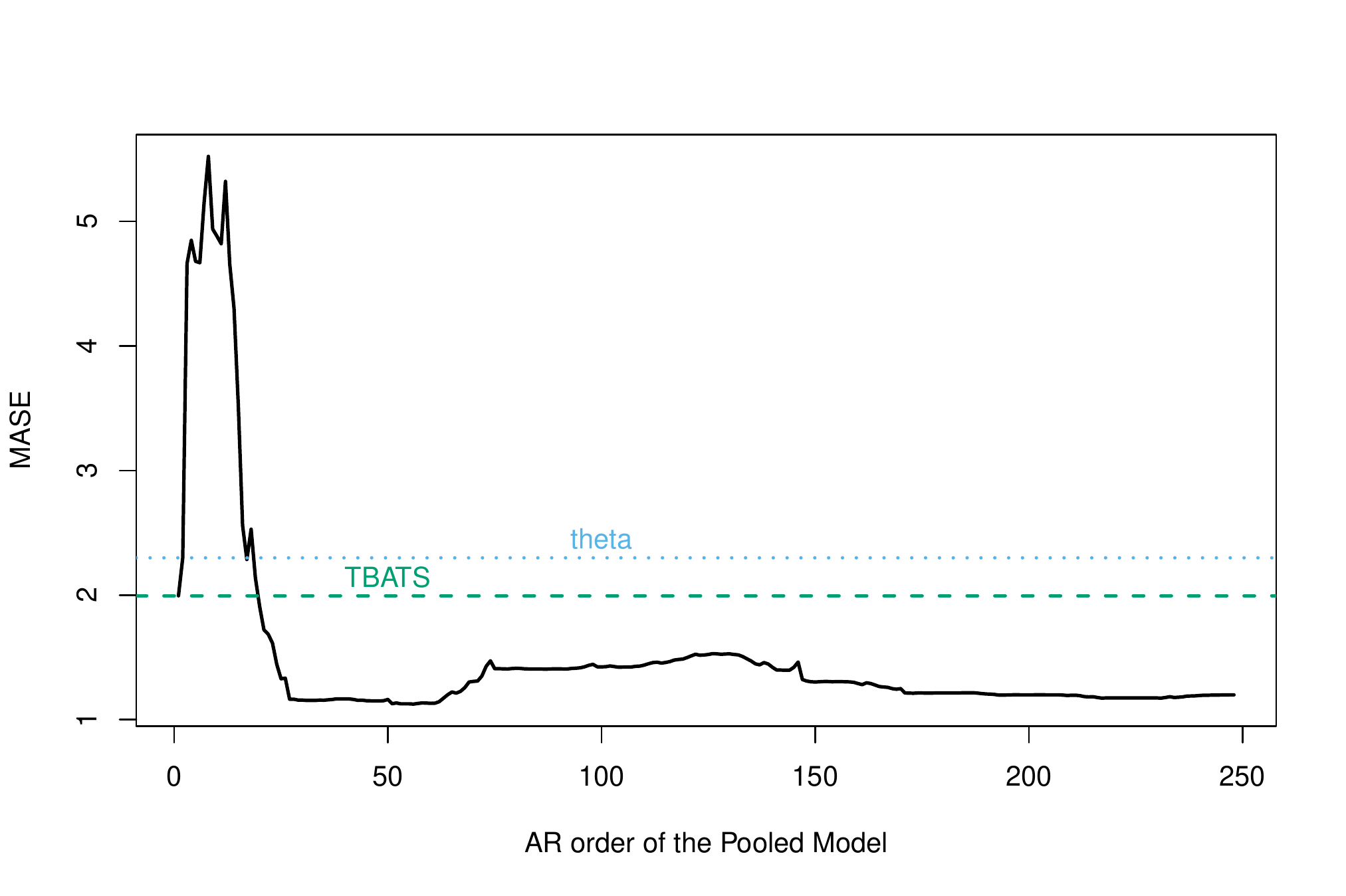}
  \caption{MASE for the pedestrian. Each time series represent hourly pedestrian count of a different point across Melbourne's CBD. Local models are TBATS and Theta in order of better performance. The same pattern of increasing accuracy with memory is experienced in this dataset, peaking at the daily cycle AR(24).}
  \label{fig:pedestrian}
\end{figure}

\begin{figure}
  \centering
  \includegraphics[width=0.75\textwidth]{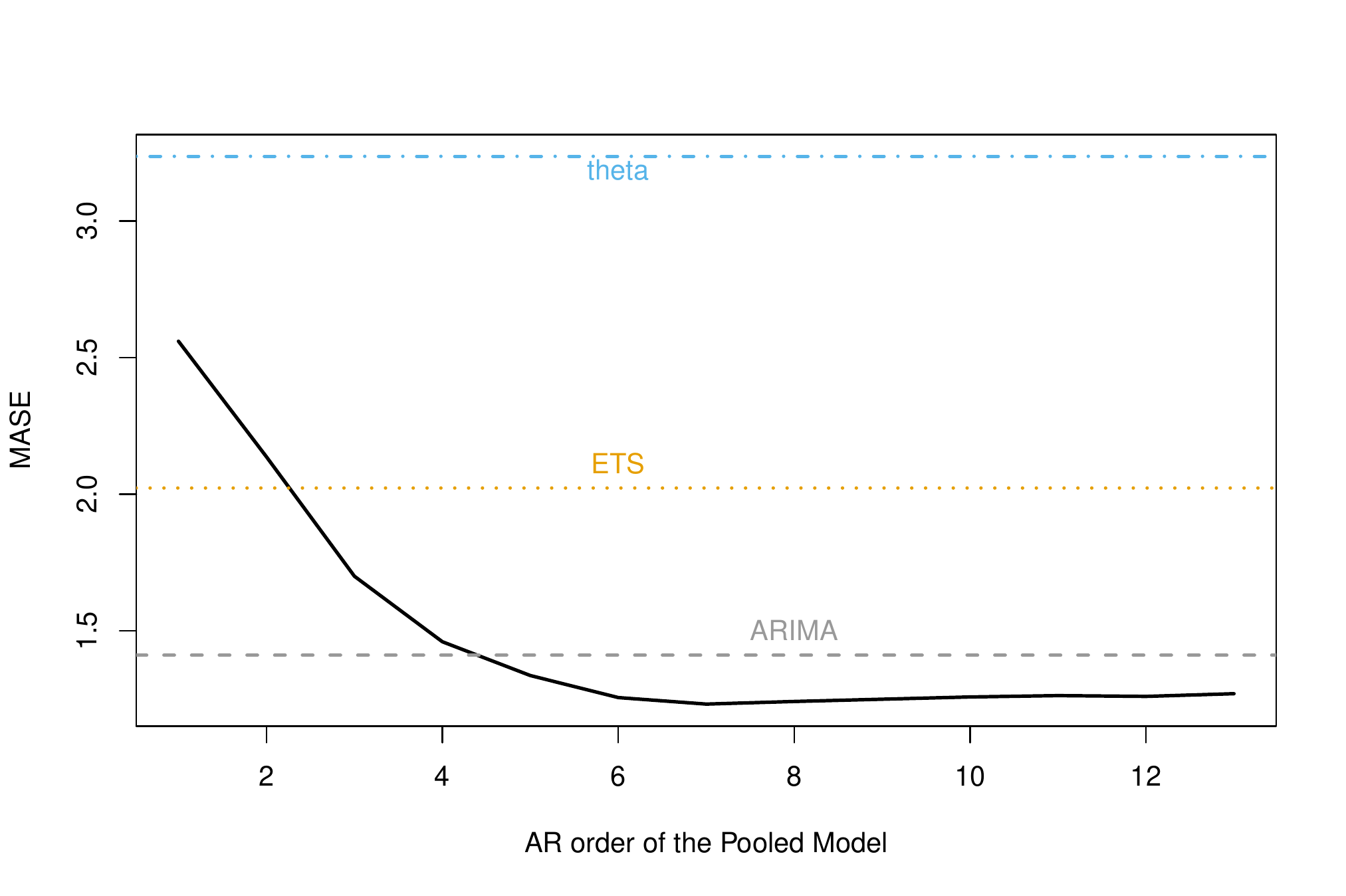}
  \caption{MASE for the COVID-19 dataset, daily cumulative deaths in 14 day in 56 regions across the globe. The performance of the global linear model increases with the memory, outperforming local state-of-the-art and saturating at a memory level of 7 days. }
  \label{fig:COVID19}
\end{figure}

%
%

\end{document}